\documentclass[sn-mathphys-num, Numbered]{sn-jnl}

\usepackage{graphicx}%
\usepackage{multirow}%
\usepackage{amsmath,amssymb,amsfonts}%
\usepackage{amsthm}%
\usepackage{mathrsfs}%
\usepackage[title]{appendix}%
\usepackage{xcolor}%
\usepackage{textcomp}%
\usepackage{manyfoot}%
\usepackage{booktabs}%
\usepackage{algorithm}%
\usepackage{algorithmicx}%
\usepackage{algpseudocode}%
\usepackage{listings}%
\usepackage{float}
\usepackage{tabularx}
\usepackage{caption}
\usepackage{subcaption} 
\usepackage{overpic}
\usepackage{tikz}
\usepackage{relsize} 
\usepackage{siunitx} 
\usepackage{colortbl}
\usepackage{multirow}
\theoremstyle{definition}\newtheorem{assumption}{Assumption}
\usepackage[capitalise]{cleveref} 
\usepackage[export]{adjustbox}
\newcommand{\reynolds}{R \hspace{-0.25mm} e}
\newcommand{\mach}{M \hspace{-0.5mm} a}

\usepackage{eso-pic}
\usepackage{graphicx} 

\AddToShipoutPictureBG*{%
  \AtPageUpperLeft{%
    \setlength{\unitlength}{1mm}%
    \put(0,-12){\makebox(\paperwidth,0)[c]{\parbox{0.8\textwidth}{\centering\textcolor{gray}{\large This paper has been accepted for publication at Nature Communications Engineering journal}}}}
  }
}

\begin{document}

\title[Learning Aerodynamics for the Control of Flying Humanoid Robots]{Learning Aerodynamics for the Control of Flying Humanoid Robots}

\author*[1,2,]{\fnm{Antonello} \sur{Paolino}}\email{antonello.paolino@iit.it}
\author[1]{\fnm{Gabriele} \sur{Nava}}
\author[1]{\fnm{Fabio} \sur{Di Natale}}
\author[1,3]{\fnm{Fabio} \sur{Bergonti}}
\author[1,3]{\fnm{Punith} \sur{Reddy Vanteddu}}
\author[4]{\fnm{Donato} \sur{Grassi}}
\author[4]{\fnm{Luca} \sur{Riccobene}}
\author[4]{\fnm{Alex} \sur{Zanotti}}
\author[2]{\fnm{Renato} \sur{Tognaccini}}
\author[5]{\fnm{Gianluca} \sur{Iaccarino}}
\author*[1,3,]{\fnm{Daniele} \sur{Pucci}}\email{daniele.pucci@iit.it}

\affil[1]{\orgdiv{Artificial and Mechanical Intelligence Laboratory}, \orgname{Istituto Italiano di Tecnologia}, \orgaddress{\street{Via San Quirico 19d}, \city{Genova}, \postcode{16163}, \country{Italy}}}

\affil[2]{\orgdiv{Dipartimento di Ingegneria Industriale}, \orgname{Universit\`a di Napoli Federico II}, \orgaddress{\street{Piazzale Vincenzo Tecchio 80}, \city{Napoli}, \postcode{80125}, \country{Italy}}}

\affil[3]{\orgdiv{School of Computer Science}, \orgname{University of Manchester}, \orgaddress{\street{Kilburn Building Oxford Road}, \city{Manchester}, \postcode{M13 9PL}, \country{UK}}}

\affil[4]{\orgdiv{Dipartimento di Scienze e Tecnologie Aerospaziali}, \orgname{Politecnico di Milano}, \orgaddress{\street{Via La Masa 34}, \city{Milano}, \postcode{20156}, \country{Italy}}}

\affil[5]{\orgdiv{Department of Mechanical Engineering}, \orgname{Stanford University}, \orgaddress{\street{440 Escondido Mall}, \city{Stanford}, \state{CA}, \postcode{94305}, \country{United States}}}

\abstract{Robots with multi-modal locomotion are an active research field due to their versatility in diverse environments. In this context, additional actuation can provide humanoid robots with aerial capabilities. Flying humanoid robots face challenges in modeling and control, particularly with aerodynamic forces. This paper addresses these challenges from a technological and scientific standpoint. The technological contribution includes the mechanical design of iRonCub-Mk1, a jet-powered humanoid robot, optimized for jet engine integration, and hardware modifications for wind tunnel experiments on humanoid robots for precise aerodynamic forces and surface pressure measurements. The scientific contribution offers a comprehensive approach to model and control aerodynamic forces using classical and learning techniques. Computational Fluid Dynamics (CFD) simulations calculate aerodynamic forces, validated through wind tunnel experiments on iRonCub-Mk1. An automated CFD framework expands the aerodynamic dataset, enabling the training of a Deep Neural Network and a linear regression model. These models are integrated into a simulator for designing aerodynamic-aware controllers, validated through flight simulations and balancing experiments on the iRonCub-Mk1 physical prototype.}


\maketitle

\section{Introduction}\label{intro} 

Aerial-terrestrial multi-modal robots are a significant area of research due to their ability to overcome the limitations of single-mode locomotion and adapt to various environments. This makes them ideal platforms for transportation, exploration, and search and rescue missions.

This paper proposes a technological approach to equip a humanoid robot with jet engines providing it with flight capabilities; consequently, it presents a comprehensive scientific methodology to model, validate, simulate, and control the aerodynamic forces acting on a flying humanoid robot during flight allowing it to operate in external environments with non-zero relative wind intensity.  

The literature on modeling and control of multi-modal robotic platforms can be roughly divided into distinct fields: task-driven mechanical design of multi-modal flying platforms, aerial systems control with and without aerodynamic modeling, and Computational Fluid Dynamics (CFD) simulations and wind tunnel tests of the flow around bluff bodies.

Concerning task-driven mechanical design of multi-modal flying platforms, the increasing demand for multi-purpose robots to replace human intervention in both tedious and hazardous environments has boosted interest in enhancing robots' adaptability to diverse scenarios, particularly focusing on their locomotion capabilities. In this context, there is a significant amount of research aiming to integrate both terrestrial and aerial locomotion into a single robot \cite{ramirez2023multimodal}. State-of-the-art platforms exhibiting multi-modal locomotion comprise various prototypes and design strategies, each tailored to the specific tasks they are intended to perform.
Existing solutions have been recently developed: i) a tailsitter drone capable of crawling reconfiguring the morphing wings as legs, ii) a wheg-based glider capable of walking off the launch point, power dive, and walk again, iii) a bio-inspired robot, mimicking a flying squirrel, combining walking and gliding locomotion modes, iv) a quadrotor with a cylindrical cage connected to a bearing that enables single-axis rolling forward and steering via differential propeller forces, v) a wheeled quadrotor to address ground obstacle avoidance by flight, vi) a multi-modal morphing robot capable to fly, roll, crawl, crouch, balance, tumble, scout, and loco-manipulate by employing its components in different ways as wheels, thrusters, and legs, vii) a multi-modal bipedal robot employing ducted fans to step over a wide ditch, and viii) a bipedal robot demonstrating remarkable versatility, being it capable of walking, flying, slacklining, and skateboarding through the seamless integration of propellers and legs \cite{daler2015bioinspired,boria2005sensor,Shin2019development,kalantari2013design,premachandra2019study,sihite2023multi,huang2017jet,kim2021leonardo}.

Usually, these platforms are designed to be lightweight to optimize flight endurance according to the low thrust-to-weight ratio propulsion mechanism used for flying to the detriment of a larger mission payload. To overcome this limitation and potentially increase the task versatility of these platforms, a viable solution is to equip a humanoid robot with flight capabilities. This approach would combine the terrestrial bipedal locomotion and manipulation abilities of humanoid robots with the aerial mobility typical of flying robots \cite{lerario2024learning}. Our goal is to enhance a humanoid robot \cite{natale2017icub} with flight capabilities, by equipping it with compact, onboard jet engines (\cref{subfig:ironcub-robot-1,subfig:ironcub-robot-2}) for high thrust-to-weight ratio, crucial for its \qty{40}{\kilo\gram} weight and potential high payload capacity \cite{mohamed2021momentum,lerario2022whole}.

The second area of literature revolves around the flight control of aerial systems, covering different approaches: some not considering aerodynamic modeling and others investigating simpler or data-driven aerodynamic models. Many of the aforementioned multi-modal robots fall under the category of Vertical Take Off and Landing (VTOL) systems \cite{premachandra2019study,huang2017jet,kim2021leonardo}. When analyzing such systems, it is common to disregard aerodynamic forces affecting the robot. This omission is justified by the typically low operational speeds of these robots, particularly when operating in indoor spaces where they are shielded from significant external wind forces \cite{ollero2021past,orsag2017dexterous}. Nevertheless, as operational velocities increase or when environmental winds generate substantial aerodynamic forces, it becomes necessary to consider these effects in control design. One approach to account for aerodynamic forces involves treating them as external disturbances. This can be achieved through the design of a momentum-based observer or similar disturbance estimation strategies \cite{cabecinhas2014nonlinear,ruggiero2014impedance}, followed by compensation of the estimated aerodynamic forces using adaptive control strategies \cite{roberts2010adaptive,antonelli2018adaptive}. However, a drawback of these methodologies is the lack of an explicit model that could be utilized, for instance, to leverage aerodynamic forces for executing aggressive maneuvers.

The complexity of the model describing aerodynamic forces depends on factors such as the robot's shape, propulsion mechanism, and operational conditions. For small VTOL systems like quadrotors, modeling the aerodynamic forces exerted on the propellers alone is often sufficient. Blade Element Momentum (BEM) theory is commonly employed to characterize the aerodynamics of the propeller under normal operating conditions \cite{theodorsen1954theory}. To mitigate the limitations of the BEM model during aggressive maneuvers, machine learning algorithms can be employed \cite{Bauersfeld2021neurobem}.
For fixed-wing aerial vehicles, incorporating a model of the aerodynamic forces generated by the lifting surfaces can improve trajectory control and agile maneuvering \cite{kai2019unified,bergonti2023codesign}. However, these models for such aerodynamic bodies are often simplified, typically overlooking aerodynamic interference between different robot links. In this context, modeling the aerodynamic effects on a flying humanoid robot presents challenges due to its non-aerodynamic shape and inherent complexity due to the high number of links. Additionally, non-negligible aerodynamic interference occurs between the robot's links, which act as bluff bodies, resulting in a large separated flow region.

The aforementioned considerations lead to the last issue examined in the literature, which involves CFD simulations and wind tunnel tests of the flow around bluff bodies aiming to evaluate the aerodynamic forces acting on them.

The bluff bodies aerodynamics studies mainly focus on spherical and cylindrical objects \cite{brown2003sphere, achenbach1972experiments, wieselsberger1922further, kritzinger2004drag, hoerner1965fluid}. However, as the complexity of body shapes increases, there is a need for designing dedicated models tailored to the specific object shape. This is due to the unsteady and multiscale characteristics of the separated turbulent flow surrounding the robot, which is represented by the non-linear Navier-Stokes partial differential equations \cite{rieutord2014fluid}. Such models may be developed based on data obtained from wind tunnel experiments and CFD simulations.

CFD solvers designed to model turbulent external flows typically utilize the Finite Volume Method \cite{eymard2000finite}. An important consideration in the computational analysis of realistic turbulent flows is the choice of the suitable modeling approach to solve the Navier-Stokes equations, balancing computational efficiency with accuracy. While highly accurate, simulations employing turbulence models such as Large Eddy Simulations (LES) tend to incur significant computational costs, especially for complex geometries \cite{erlebacher1992les}. Therefore, for engineering applications, turbulence models based on the Reynolds Averaged Navier-Stokes (RANS) equations are commonly preferred. Examples include the \textit{Realizable $k-\varepsilon$} and \textit{Shear-Stress Transport (SST) $k-\omega$} models \cite{Reynolds1895RANS, shih1995realke, menter1994sstkw}.
Due to the assumptions invoked in the RANS closures, the corresponding simulations must subsequently undergo empirical validation through wind tunnel experiments. This type of validation process has already been undertaken for different models, including comparisons between results obtained from wind tunnel experiments and CFD simulations on cyclists, which share similarities with humanoid robots. These comparisons have demonstrated that RANS models can represent aerodynamic forces with a limited margin of error when compared to LES \cite{defraeye2010aerodynamic, blocken2018aerodynamic, blocken2021impact,  defraeye2010computational}.

Here, we aim to provide flight control capabilities to flying humanoid robots in the presence of aerodynamic disturbances. Additionally, the contribution at the modeling and control level can be generalized and applied to any shape-shifting drone under aerodynamic disturbances. 

Initially, we introduce the mechanical design of iRonCub-Mk1, a pioneering jet-powered humanoid robot developed at the Istituto Italiano di Tecnologia (IIT), emphasizing the aspects of hardware optimization essential for the integration of jet engines into its design. Additionally, we outline the hardware modifications required to allow wind tunnel experiments on humanoid robots. Specifically, our focus extends to enabling precise measurements of the overall aerodynamic forces and moments exerted on the robot, and providing the capability to capture real-time averaged distributed pressure data.

Subsequently, we propose a comprehensive approach for modeling and control of the aerodynamic forces influencing a flying humanoid robot. Our methodology starts from wind tunnel experiments performed on the iRonCub robot at the wind tunnel of Politecnico di Milano (GVPM) to measure the aerodynamic forces. Then, we validate the CFD simulation results with the data acquired during experiments at GVPM and collect an extended aerodynamic dataset via a framework for automated CFD simulations. Furthermore, this dataset is leveraged to develop models to establish the relationship between distributed aerodynamic forces and the robot's attitude and joint configurations: a Deep Neural Network, and a linear regression model. Finally, the integration of these models into a dynamics simulator facilitates the design of aerodynamic-aware controllers. The efficacy of our methodology is verified through validation with iRonCub-Mk1 via flight simulations and balancing experiments conducted on the physical prototype. Although the proposed methodology does not introduce new methods for computing the aerodynamics simulations and for performing the wind tunnel experiments, its novelty lies in the complete approach that leverages CFD simulations and wind tunnel experiments for designing a whole-body aerodynamic controller for a flying humanoid robot.

\section{Results}\label{results}

This section presents the outcomes of our proposed methodology for evaluating, validating, and designing controls for aerodynamic forces on the flying humanoid robot iRonCub-Mk1, after introducing the mechanical design of the robot itself.

\subsection{iRonCub-Mk1 Mechanical Design}

\begin{figure}[htpb]
    \centering
    \begin{subfigure}[]{0.4\textwidth}
        \centering
        \includegraphics[width=\textwidth]{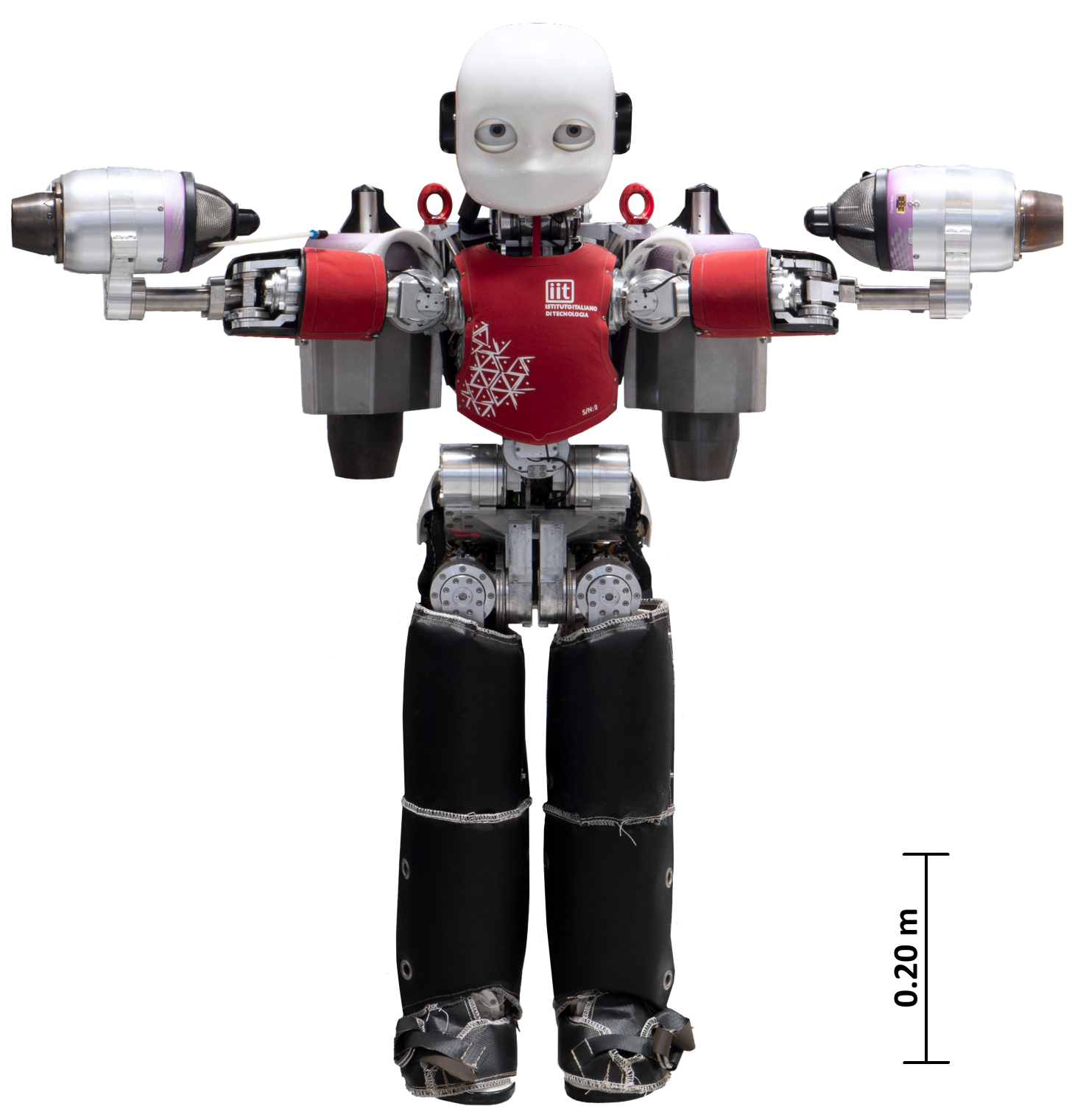}
        \caption{}
        \label{subfig:ironcub-robot-1}
    \end{subfigure}
    \hspace{2cm}
    \begin{subfigure}[]{0.35\textwidth}
        \centering
        \includegraphics[width=\textwidth]{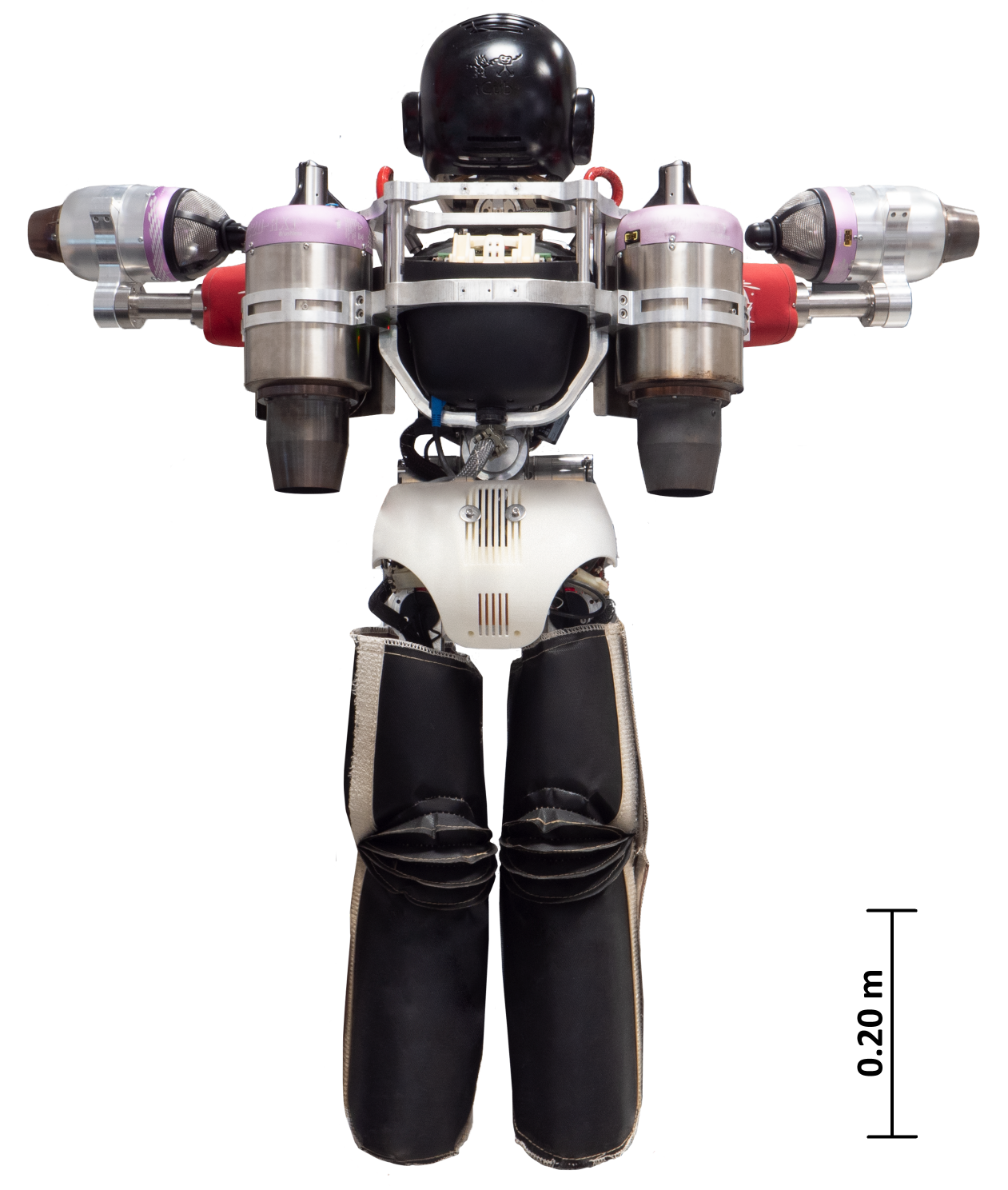}
        \caption{}
        \label{subfig:ironcub-robot-2}
    \end{subfigure}
    \\
    \begin{subfigure}[]{0.49\textwidth}
        \centering
        \includegraphics[width=\textwidth]{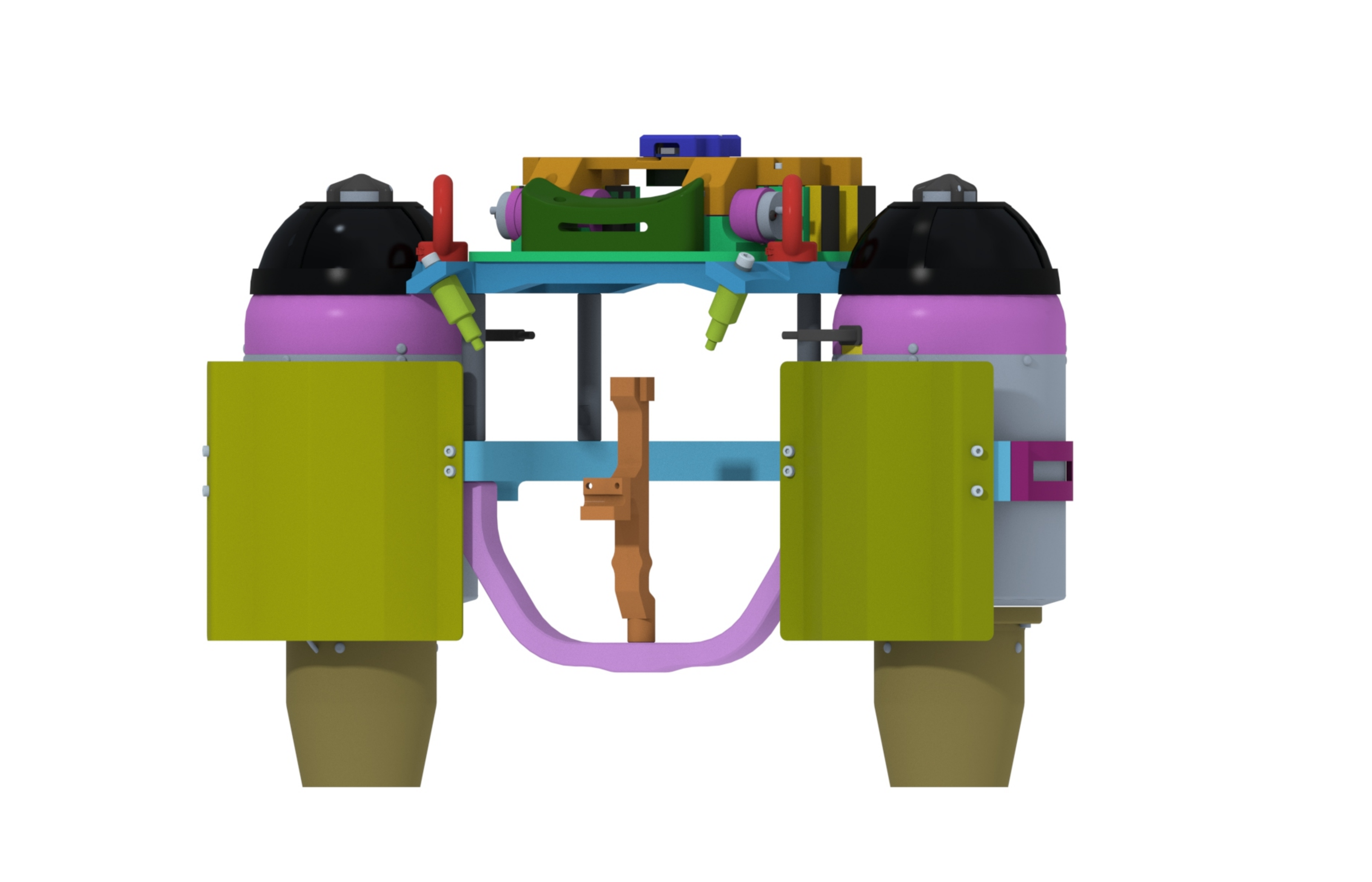}
        \caption{}
        \label{subfig:jetpack_connection_sectional}
    \end{subfigure}
    \hfill
    \begin{subfigure}[]{0.45\textwidth}
        \centering
        \begin{tikzpicture}
            \draw (0, 0) node[inner sep=0] {\includegraphics[width=\textwidth]{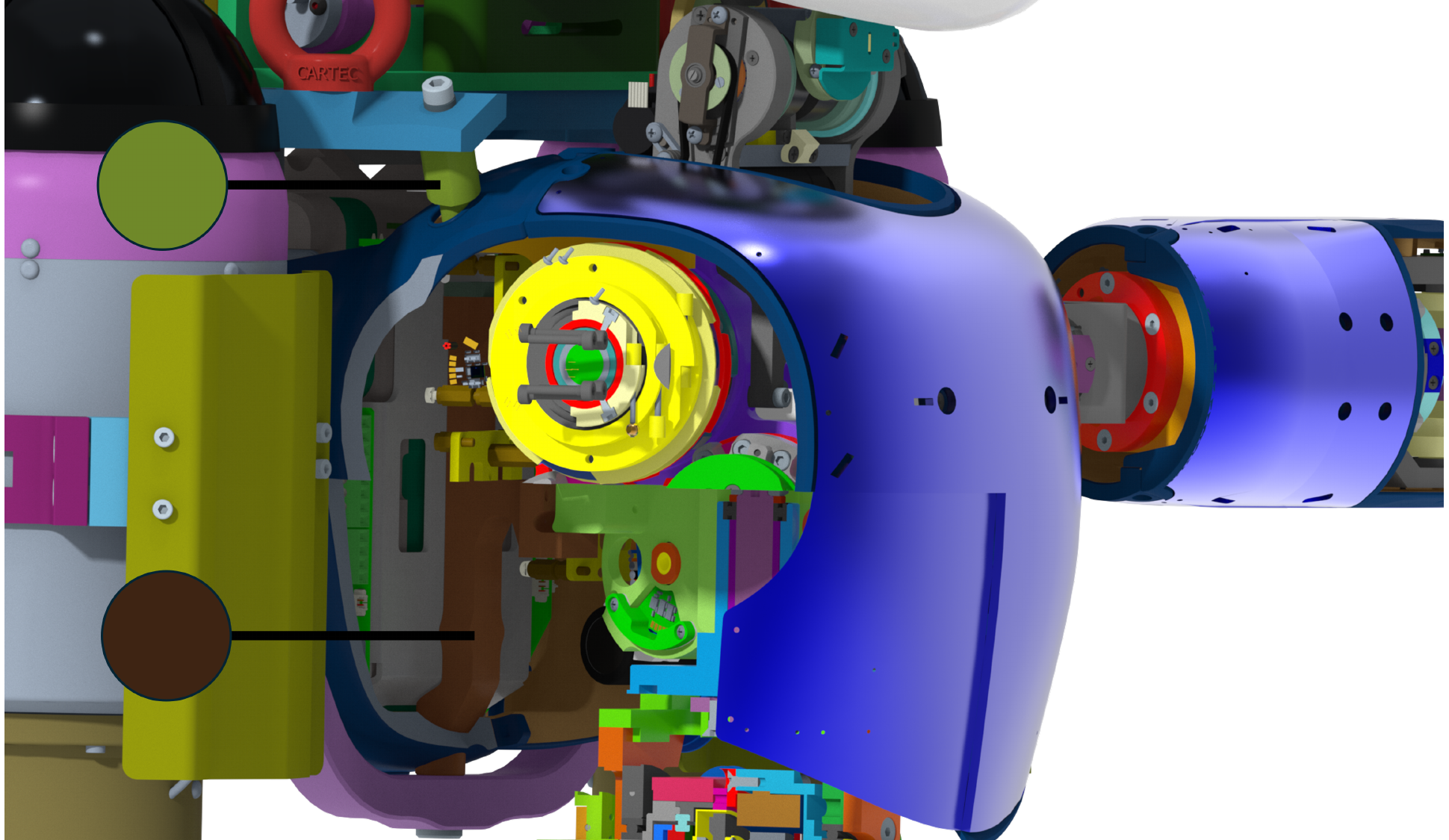}};
            \draw (-2.68, -1.03) node[text=white] {1};
            \draw (-2.69, 1.12) node[text=white] {2};
        \end{tikzpicture}
        \caption{}
        \label{subfig:jetpack_connection_sectional_super_zoom}
    \end{subfigure}
    \\
    \begin{subfigure}[]{0.49\textwidth}
        \centering
        \includegraphics[width=\textwidth]{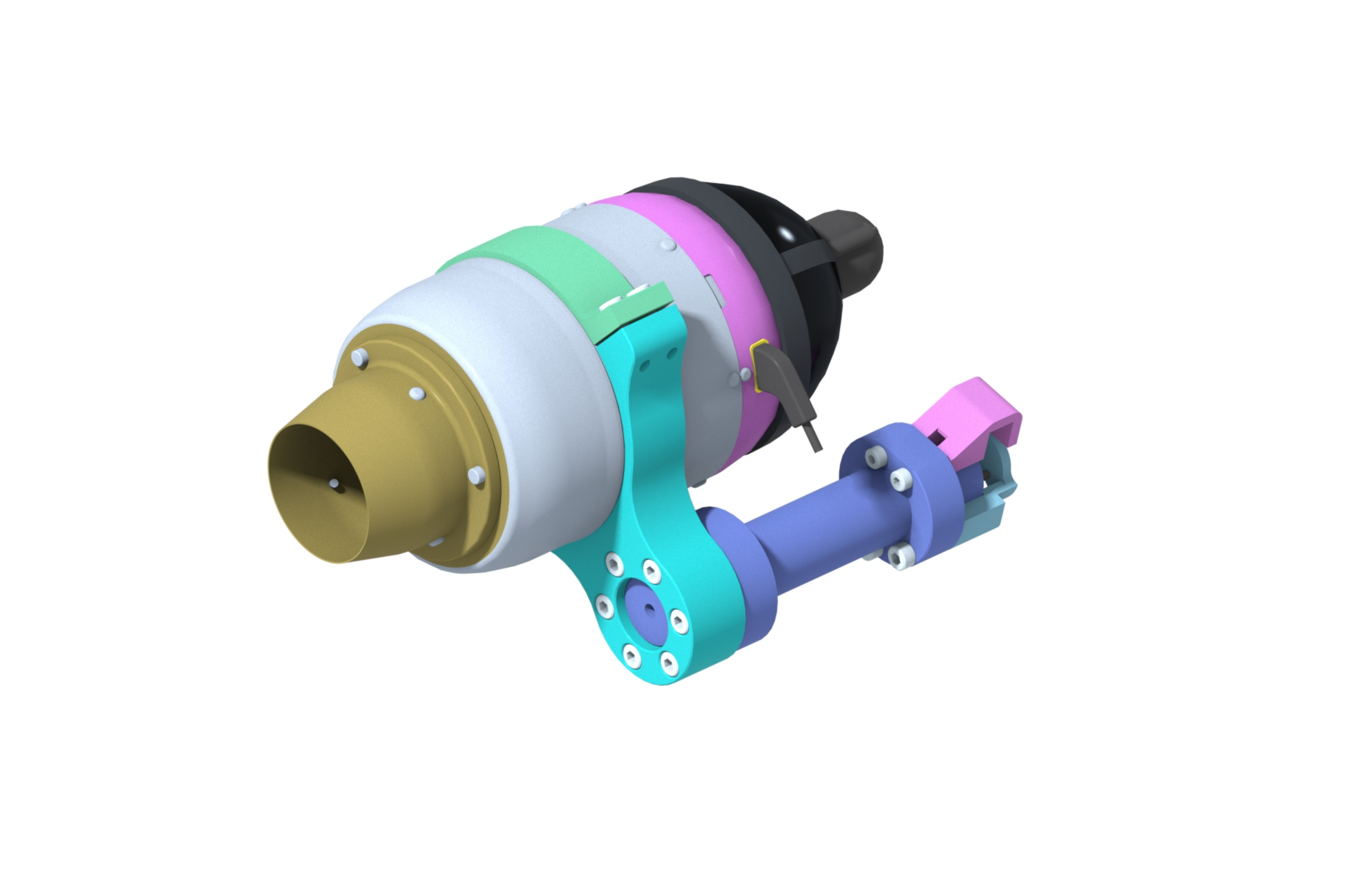}
        \caption{}
        \label{subfig:forearm_assembled}
    \end{subfigure}
    \hfill
    \begin{subfigure}[]{0.49\textwidth}
        \centering
            \begin{tikzpicture}
                \draw (0, 0) node[inner sep=0] {\includegraphics[width=\textwidth]{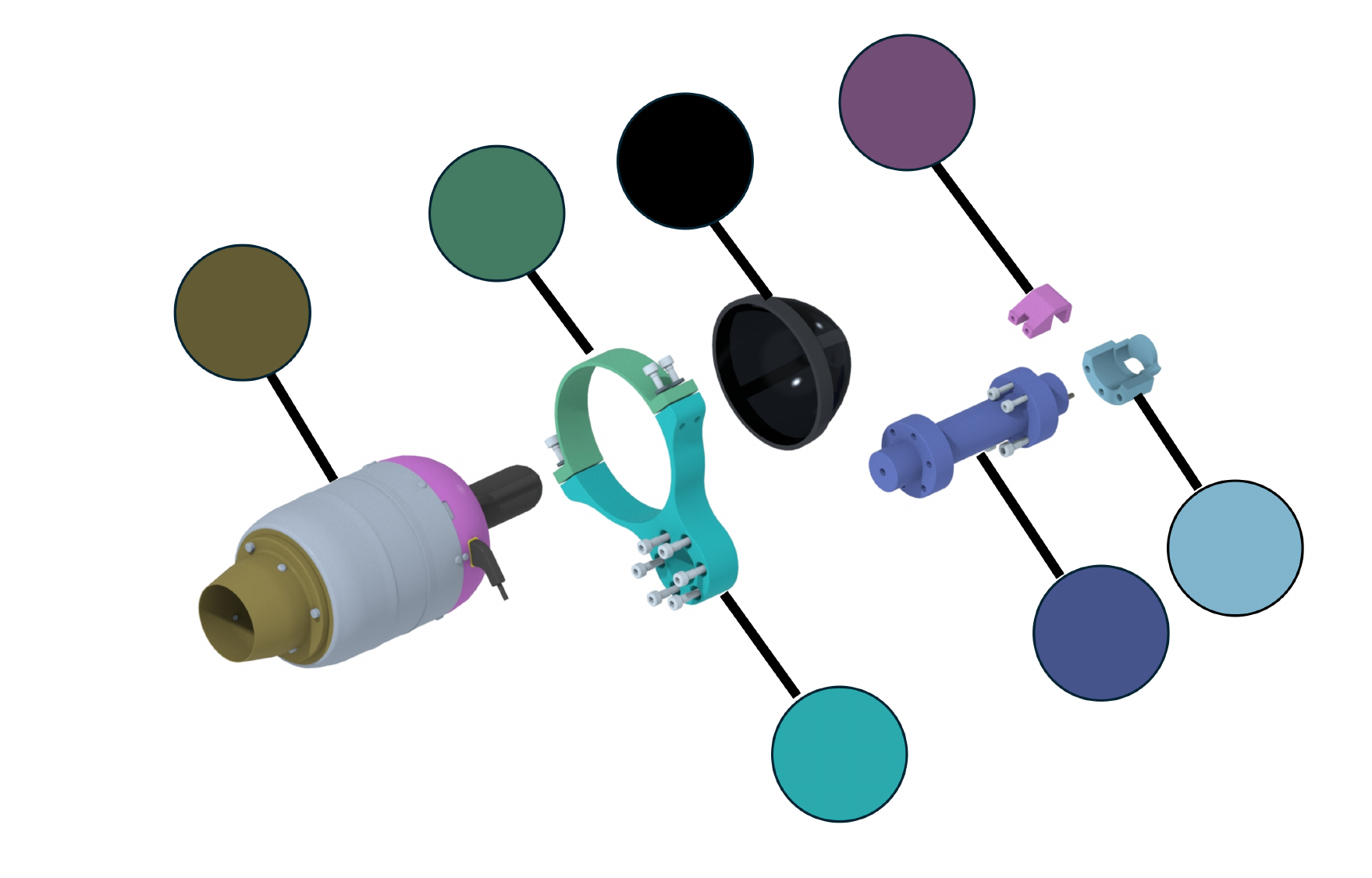}};
                \draw (3.03, -0.61) node[text=white] {3};
                \draw (1.23, 1.85) node[text=white] {4};
                \draw (2.28, -1.09) node[text=white] {5};
                \draw (-0.02, 1.53) node[text=white] {6};
                \draw (0.83, -1.75) node[text=white] {7};
                \draw (-1.05, 1.23) node[text=white] {8};
                \draw (-2.44, 0.68) node[text=white] {9};
            \end{tikzpicture}
        \caption{}
        \label{subfig:forearm_exploded}
    \end{subfigure}
    \caption{\textbf{Design of the iRonCub-Mk1 physical prototype.} Front (\subref{subfig:ironcub-robot-1}) and rear (\subref{subfig:ironcub-robot-2}) pictures of the iRonCub-Mk1 robot physical prototype.
    (\subref{subfig:jetpack_connection_sectional}) Jetpack assembled with jet engines and electronics; (\subref{subfig:jetpack_connection_sectional_super_zoom}) jetpack connections: (1) steel spine assembled on the robot, (2) spacers on both sides of the robot as additional jetpack connections; (\subref{subfig:forearm_assembled}) forearm assembled with the jet engine; (\subref{subfig:forearm_exploded}) exploded view of the forearm assembly: (3 and 4) reinforcements for the elbow joint of the robot, (5) titanium forearm, (6) jetnet, (7 and 8) bracket support for the jet engine, (9) JetCat P100-RX jet engine.
    }
    \label{fig:1}
\end{figure}

iRonCub-Mk1, an iteration of iCub v2.7 \cite{natale2017icub}, utilizes jet propulsion with four jet engines. Positioned on the robot's back, two engines generate lift to balance its weight, while the remaining two, situated on the forearms, manage attitude control. This setup allows the robot to regulate its attitude through thrust adjustment and joint movement, altering both thrust direction and force application points.

The connection of the back jet engines is achieved through a jetpack, which serves as an interface rigidly linking the two engines to the back of the robot's torso. These engines are configured to exert vertical thrust when the robot stands upright. The jetpack is connected to iCub v2.7 using two existing connection points near the shoulder, and a custom-designed stainless steel spine, as depicted in \cref{subfig:jetpack_connection_sectional_super_zoom}.
Mounting the jetpack did not necessitate the removal of any iCub v2.7 components, but required the permanent installation of the spine. The jet engines are secured with a friction-based mechanism regulated by screws. Nord-Lock washers are employed to prevent screw loosening due to jet engine vibrations.
Furthermore, the jetpack serves as a storage unit for all necessary jet engine components, including electronics and pumps.

The front jet engines are linked to the robot forearms through a custom-designed component, replacing the standard iCub forearms. This iRonCub forearm is engineered with a connection strategy mirroring that of iCub, facilitating the conversion process from iCub to iRonCub. The forearm is rigidly connected to the elbow via a threaded connection. As additional safety measure in the event of threaded connection failure, two supplementary components have been incorporated to secure the forearm to the elbow.

The jetpack and the forearms are designed to hold P100-RX JetCat jet engines for the forearms and P220-RXi JetCat jet engines for the jetpack.
The robot mass with engines assembled is \qty{43.3}{\kilo \gram}, \qty{10.3}{\kilo \gram} more than iCub v2.7, which distributes among jetpack, forearms, and engines. \cref{subfig:ironcub-robot-1,subfig:ironcub-robot-2} represents the final result of iRonCub Computer-aided Design (CAD).

\subsection{Wind Tunnel Experiments}\label{results/wind_tunnel}

\begin{figure}[htpb]
    \centering
    \begin{subfigure}[b]{0.44\textwidth}
        \centering
        \begin{subfigure}[b]{\textwidth}
            \centering
            \includegraphics[width=\textwidth]{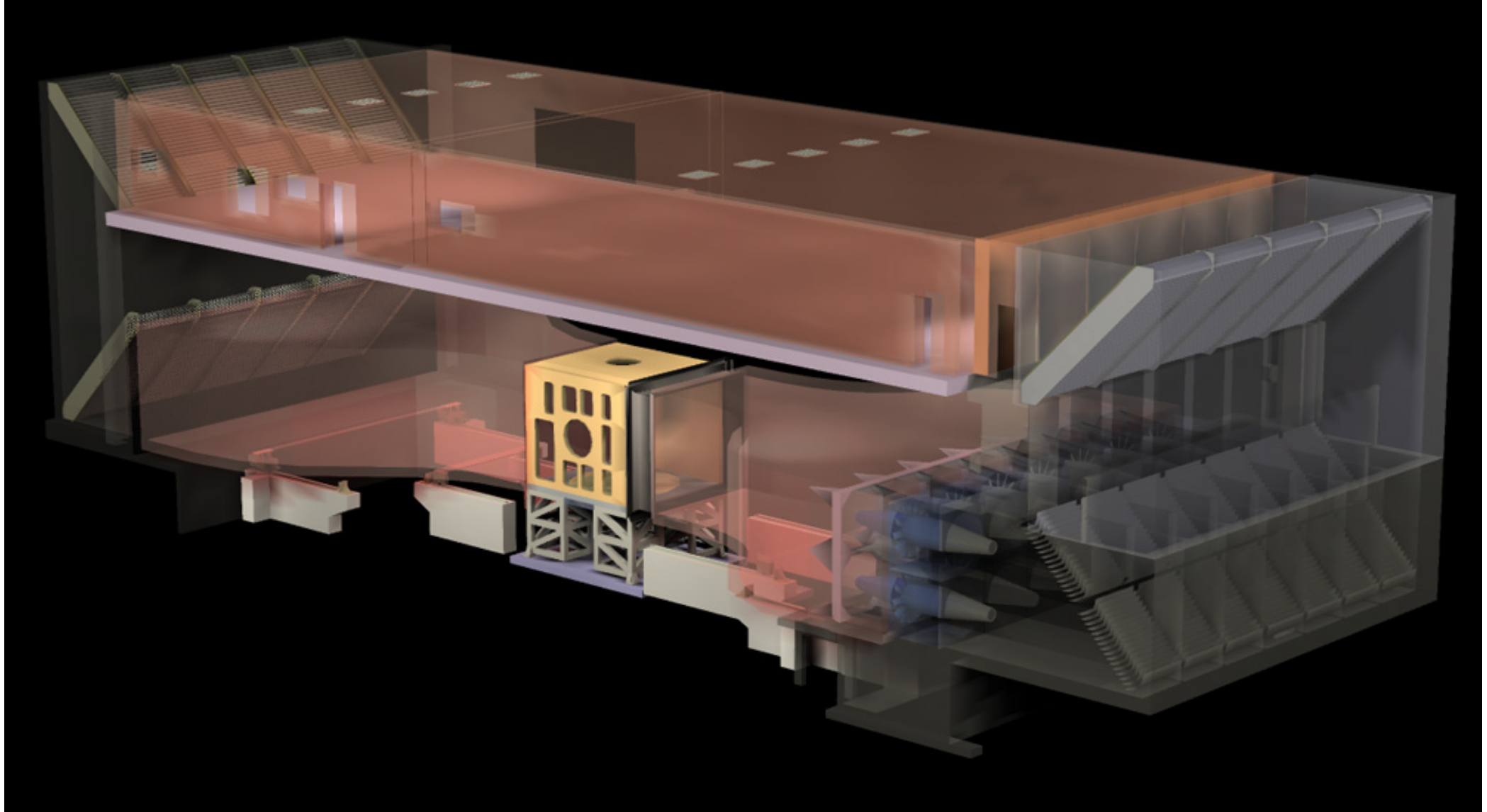}
            \caption{}
            \label{subfig:gvpm_layout}
        \end{subfigure}
        \\
        \begin{subfigure}[b]{\textwidth}
            \centering
            \includegraphics[width=\textwidth]{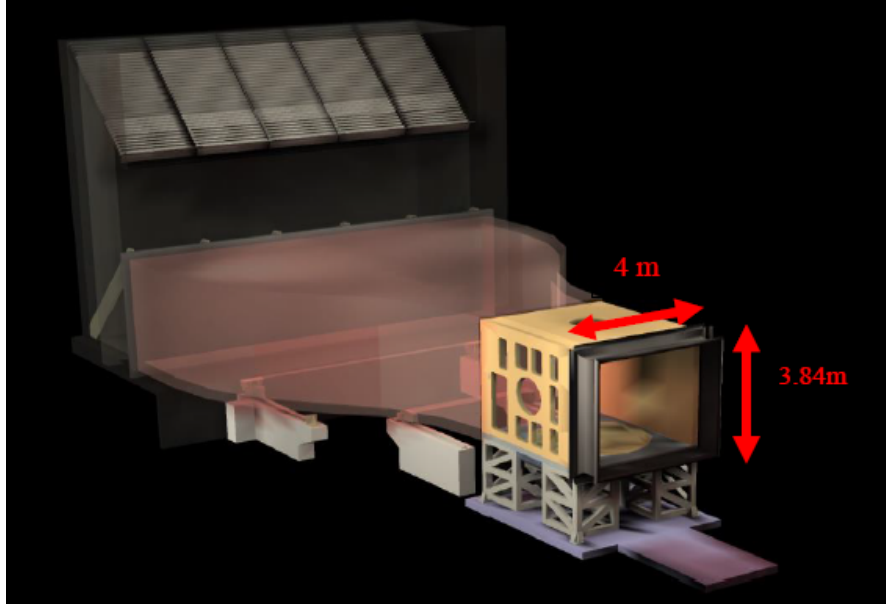}
            \caption{}
            \label{subfig:gvpm_test}
        \end{subfigure}
    \end{subfigure}
    \hspace{0.5cm}
    \begin{subfigure}[b]{0.45\textwidth}
         \centering
         \includegraphics[width=\textwidth]{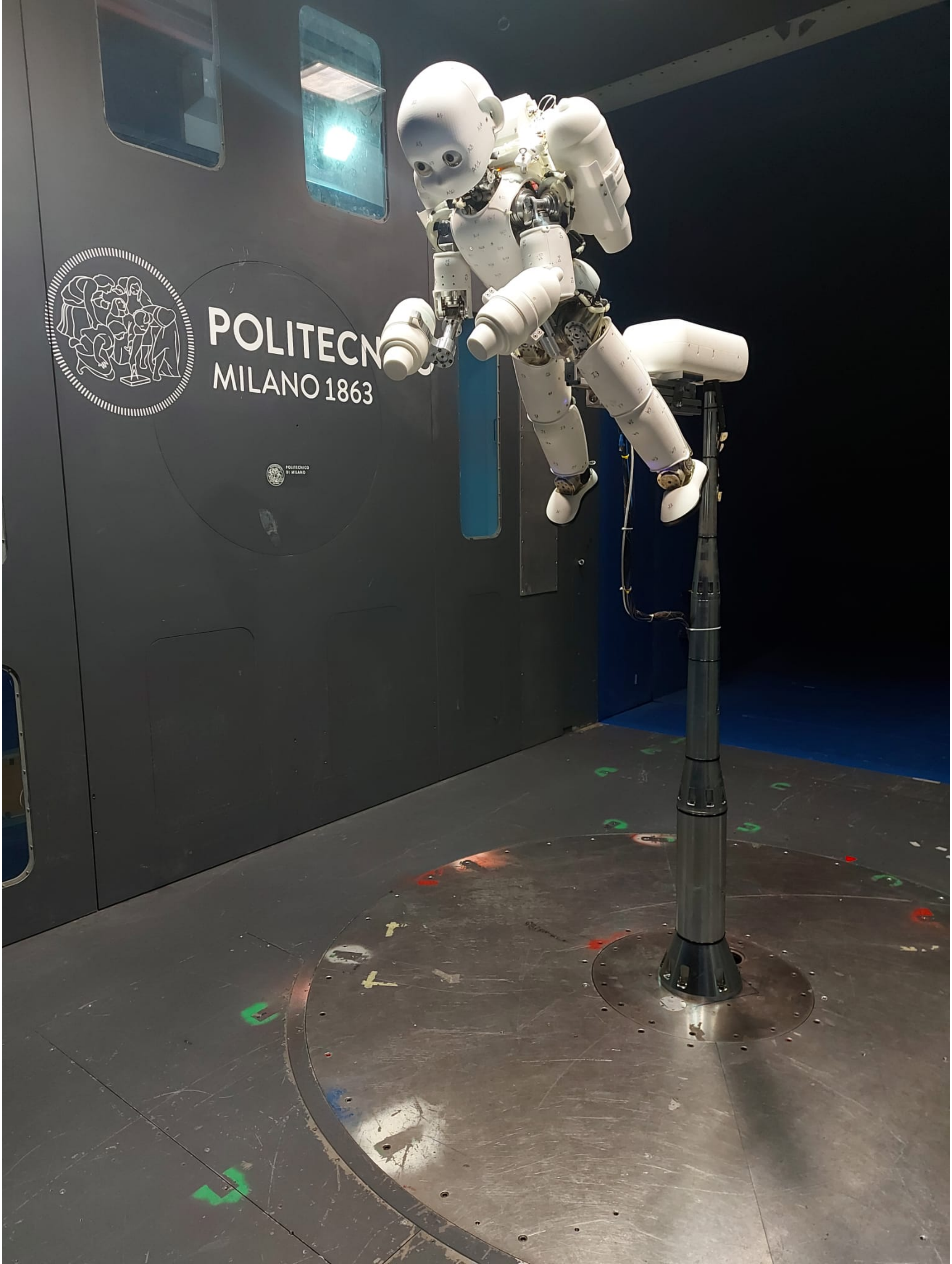}
         \caption{}
         \label{subfig:gvpm_robot}
    \end{subfigure}
    \\
    \begin{subfigure}[]{0.45\textwidth}
        \centering
        \adjincludegraphics[width=\textwidth,trim={{0.05\width} {0.05\height} {0.1\width} {0.12\height}},clip]{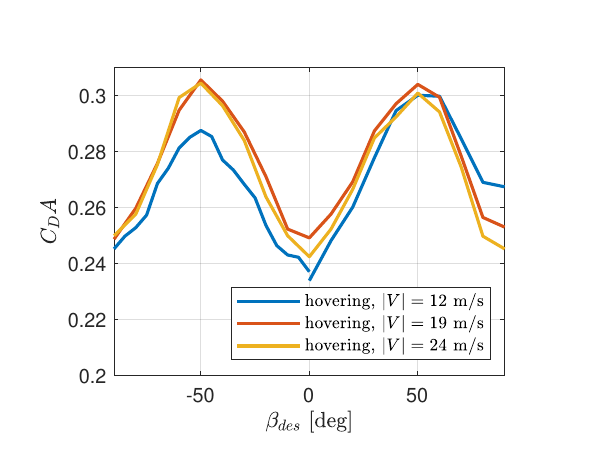}
        \caption{}
        \label{subfig:hov-WT-Rey-CdA}
    \end{subfigure}
    \hspace{0.5cm}
    \begin{subfigure}[]{0.45\textwidth}
        \centering
        \adjincludegraphics[width=\textwidth,trim={{0.05\width} {0.05\height} {0.1\width} {0.12\height}},clip]{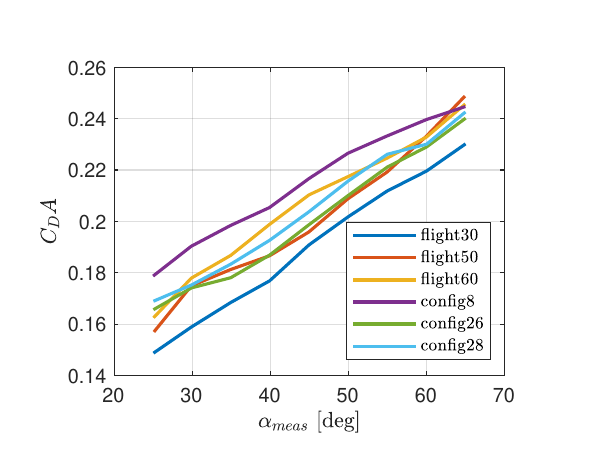}
        \caption{}
        \label{subfig:fl-wt-jointvar-CdA}
    \end{subfigure}
    \caption{\textbf{Wind tunnel experiments.} (\subref{subfig:gvpm_layout}) Layout of the large wind tunnel of Politecnico di Milano (GVPM), (\subref{subfig:gvpm_test}) detail of the low-turbulence test section, (\subref{subfig:gvpm_robot}) iRonCub robot mounted on the moving pole inside the low-turbulence test chamber. Experiment data on drag force area ($C_DA$, reference area $A=\qty{1}{m^2}$) acting on the robot: (\subref{subfig:hov-WT-Rey-CdA}) \texttt{hovering} configuration for different airspeeds, (\subref{subfig:fl-wt-jointvar-CdA}) airspeed $|v_a|=\qty{17}{\meter/\second}$ for different \texttt{flight} configurations.}
    \label{fig:2}
\end{figure}

Wind tunnel experiments were conducted to understand the global properties of the fluid flow surrounding the iRonCub robot. A total of 400 measurements were taken across 54 tests, examining 4 symmetric and 15 non-symmetric configurations of robot joints. These tests encompassed various airspeed values and attitude angles.
The outcome of each test consists of a set of measurements, including global aerodynamic force and torque, and pressure distribution data.

We first investigate the effects resulting from changes in airspeed and geometry. Specifically, we conducted 4 tests with the robot in \texttt{hovering} configuration. In each test, we varied the robot's side-slip angle and adjusted the airspeed, affecting the Reynolds number ($\reynolds$) accordingly (for more details on Reynolds number, see Supplementary Note~1). 
The results depicted in \cref{subfig:hov-WT-Rey-CdA} indicate a measurable local effect of $\reynolds$ only for high positive side-slip angles, corresponding to when the support bar is directly exposed to the wind rather than being in the wake of the robot. Additionally, an asymmetry is observed between the positive and negative side-slip angle ranges in the test with airspeed \qty{12}{\meter/\second}. This discrepancy could be attributed to a calibration error in the scale, which experiences greater loading than under standard conditions due to the robot's weight. 
Throughout the remaining range, the variation in force areas (see Supplementary Note~2) is below 5\%.

The experimental activities also included tests with different joint configurations (whose details are reported in Supplementary Figs.~8-9).
Results in \cref{subfig:fl-wt-jointvar-CdA} show that the global force areas change by almost 20\% between different robot configurations, highlighting the fact that effects produced by different joint angles are more relevant, in our use case, than differences produced by different freestream $\reynolds$. 

Additionally, the dependency of the flow behavior from the local Reynolds number cannot be decoupled from the joint configurations effects, and it has not been possible to have local measurements to correlate local laminar/turbulent separations with the local Reynolds number variation.

These considerations led us to assume $\reynolds$-independence for such high turbulence largely separated flow on the robot, at least in the range of the studied flow conditions. Therefore, we initially decided not to include wind speed as a variable of the aerodynamic CFD dataset collection, even if this assumption could be removed in the future by including a larger number of CFD simulations at different airspeeds.

A more complete overview of the results of the wind tunnel experimental campaign is reported in Supplementary Figs.~10-13, where also the experiments on the robot different flight configurations performed by changing the only pitch angle are reported.

From the analysis of the aerodynamic efficiency reported in Supplementary Fig~14 and obtained for different flight symmetric configurations, the robot exhibits a low maximum efficiency. This is an expected result since the robot has not yet been optimized for aerodynamics, but future developments regarding this aspect are detailed in the Discussion section.

\subsection{Aerodynamics CFD Simulations}\label{results/CFD}

We conduct Computational Fluid Dynamics (CFD) simulations using simplified geometry models of both the robot and the wind tunnel support (see Supplementary Fig.~3).
We encounter several challenges in our simulations. Firstly, the robot's geometry remains complex, and we face a high reference Reynolds number (see Supplementary Note~1). Consequently, we anticipate highly turbulent flow over the robot's surface and wake. Additionally, generating accurate meshes for such complex geometries poses another challenge. To address these issues, we have opted to utilize the RANS equations written for incompressible fluid along with two turbulence models: \textit{Realizable $k-\varepsilon$} and \textit{SST $k-\omega$}, following insights from the literature. We conduct two separate sets of simulations. The first set utilizes both the robot model and the wind tunnel support geometry to ensure consistency with wind tunnel tests. The second set focuses solely on the robot model to generate results for the flying robot in a free airstream, which is intended for aerodynamic modeling purposes. Results on this second set of simulations will be discussed in the \textit{Performances of the Aerodynamic Models} Section.

\begin{figure}[htpb]
    \centering
    \begin{subfigure}[]{0.35\textwidth}
         \centering
         \adjincludegraphics[width=\textwidth,trim={{0.05\width} {0.05\height} {0.1\width} {0.12\height}},clip]{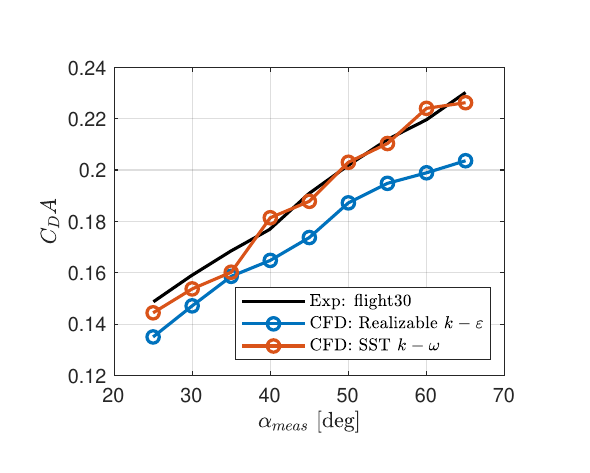}
         \caption{}
         \label{subfig:fl30-2-CFD-WT-CdA}
    \end{subfigure}
    \hspace{1cm}
    \begin{subfigure}[]{0.35\textwidth}
         \centering
         \adjincludegraphics[width=\textwidth,trim={{0.05\width} {0.05\height} {0.1\width} {0.12\height}},clip]{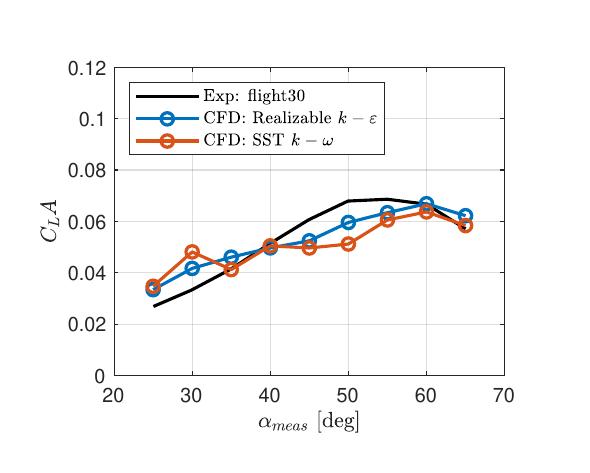}
         \caption{}
         \label{subfig:fl30-2-CFD-WT-ClA}
    \end{subfigure}
    \\
    \begin{subfigure}[]{0.45\textwidth}
        \centering
        \adjincludegraphics[width=\textwidth,trim={0 {0.22\height} 0 {0.15\height}},clip]{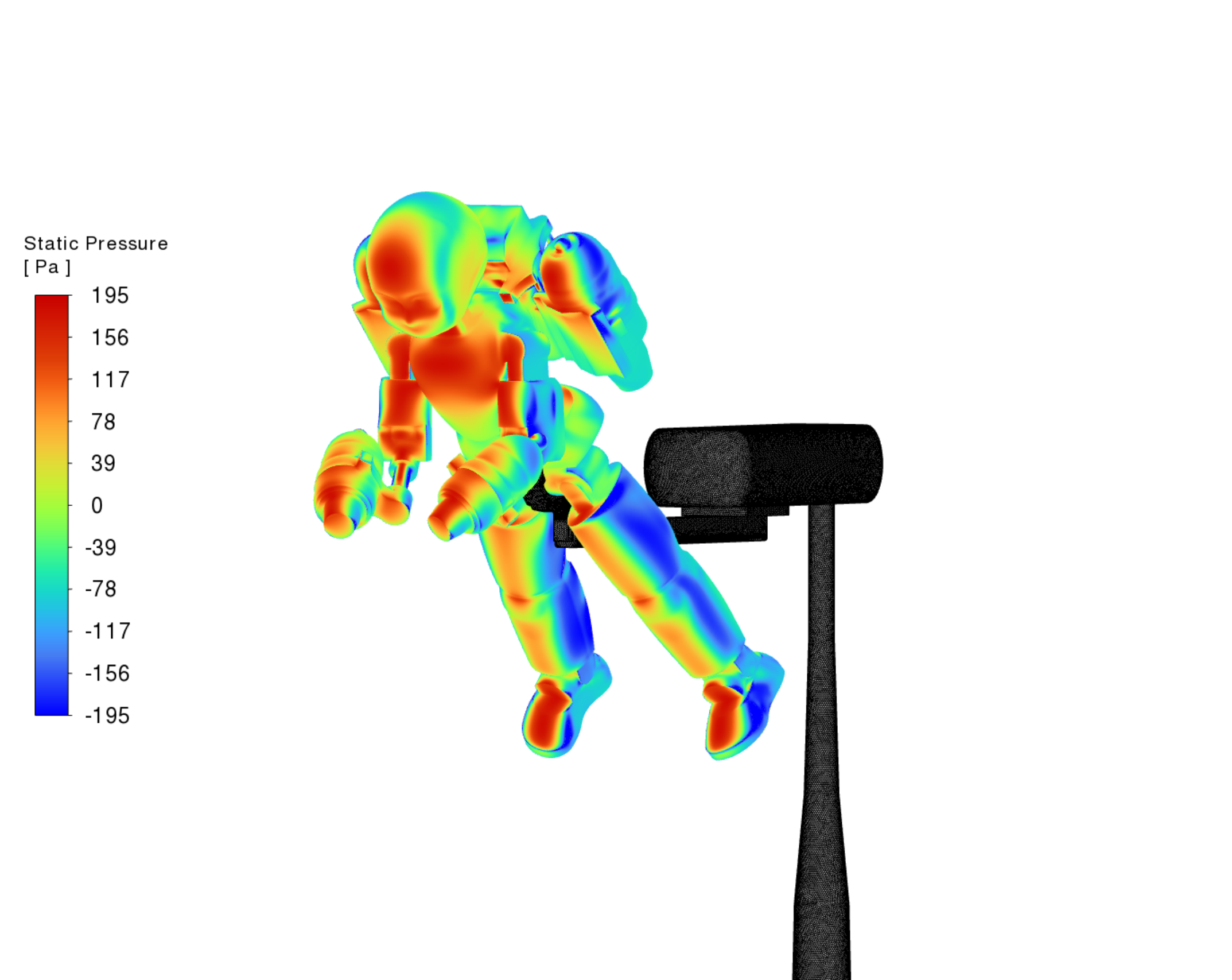}
        \caption{}
    \label{subfig:cfd-robot-press}
    \end{subfigure}
    \hspace{0.5cm}
    \begin{subfigure}[]{0.45\textwidth}
        \centering
        \adjincludegraphics[width=\textwidth,trim={0 {0.20\height} 0 {0.17\height}},clip]{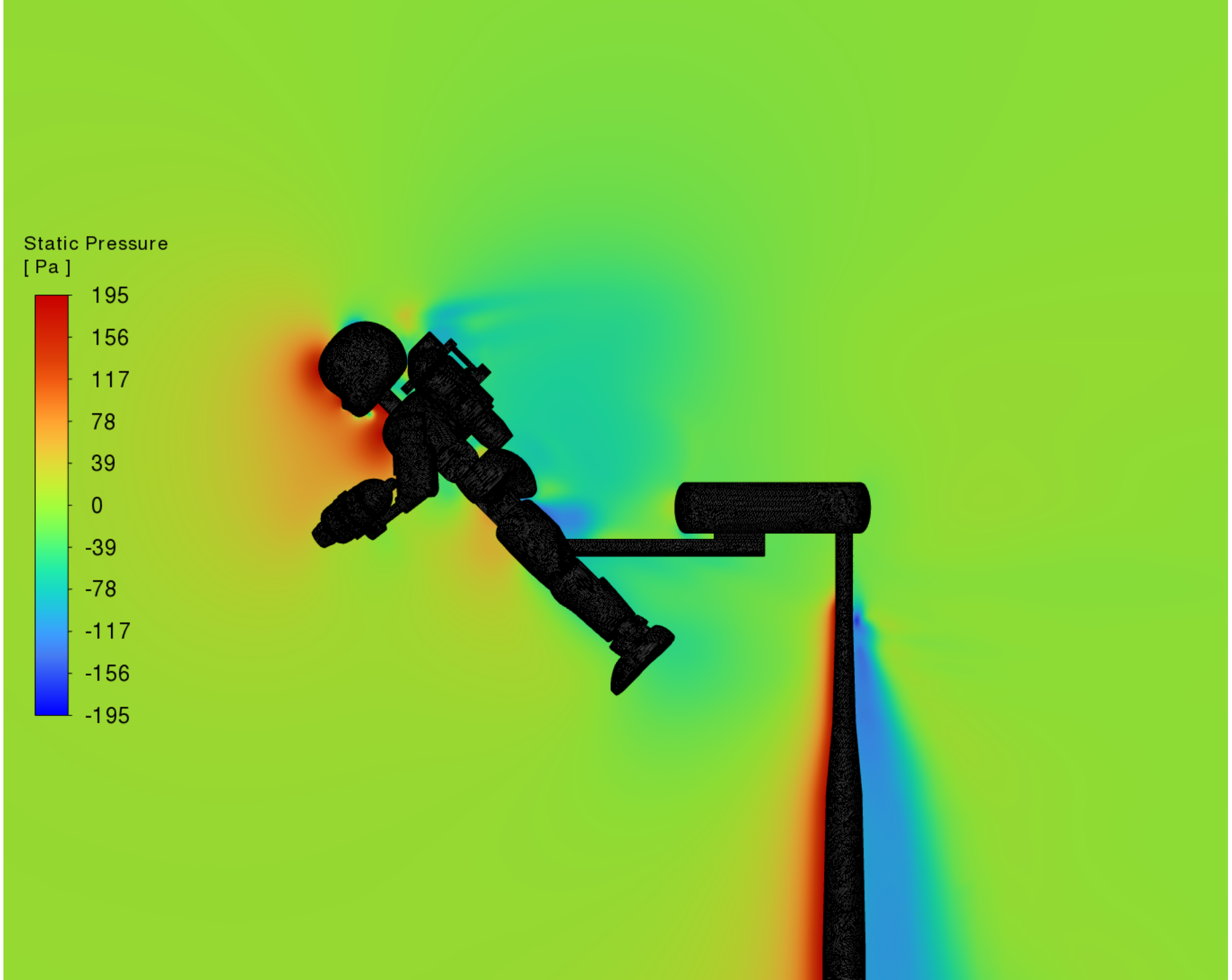}
        \caption{}
        \label{subfig:cfd-press-contour}
    \end{subfigure}
    \\
    \begin{subfigure}[]{0.45\textwidth}
        \centering
        \adjincludegraphics[width=\textwidth,trim={0 {0.21\height} 0 {0.20\height}},clip]{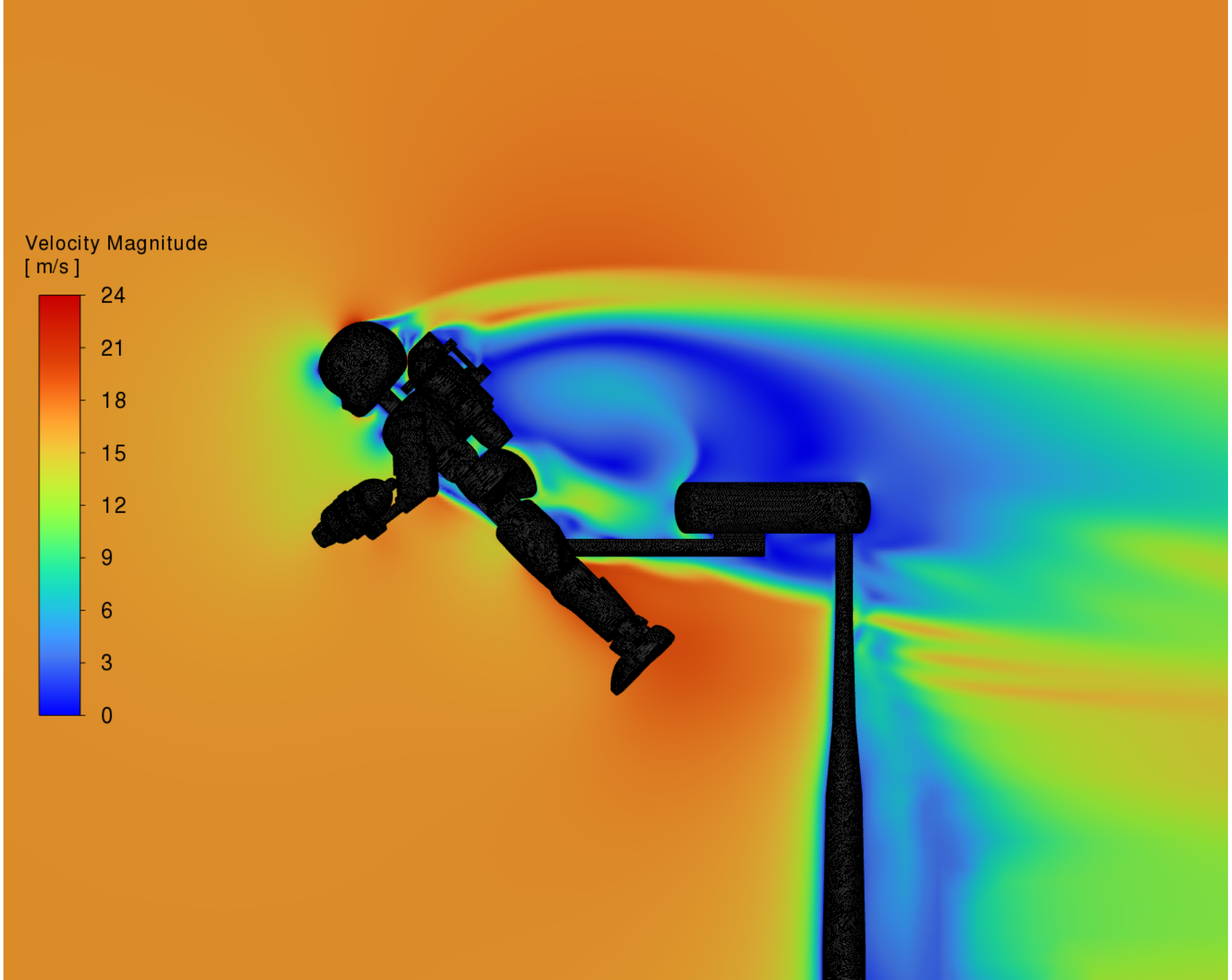}
        \caption{}
        \label{subfig:cfd-vel-contour}
    \end{subfigure}
    \hspace{0.5cm}
    \begin{subfigure}[]{0.45\textwidth}
        \centering
        \adjincludegraphics[width=\textwidth,trim={0 {0.21\height} 0 {0.20\height}},clip]{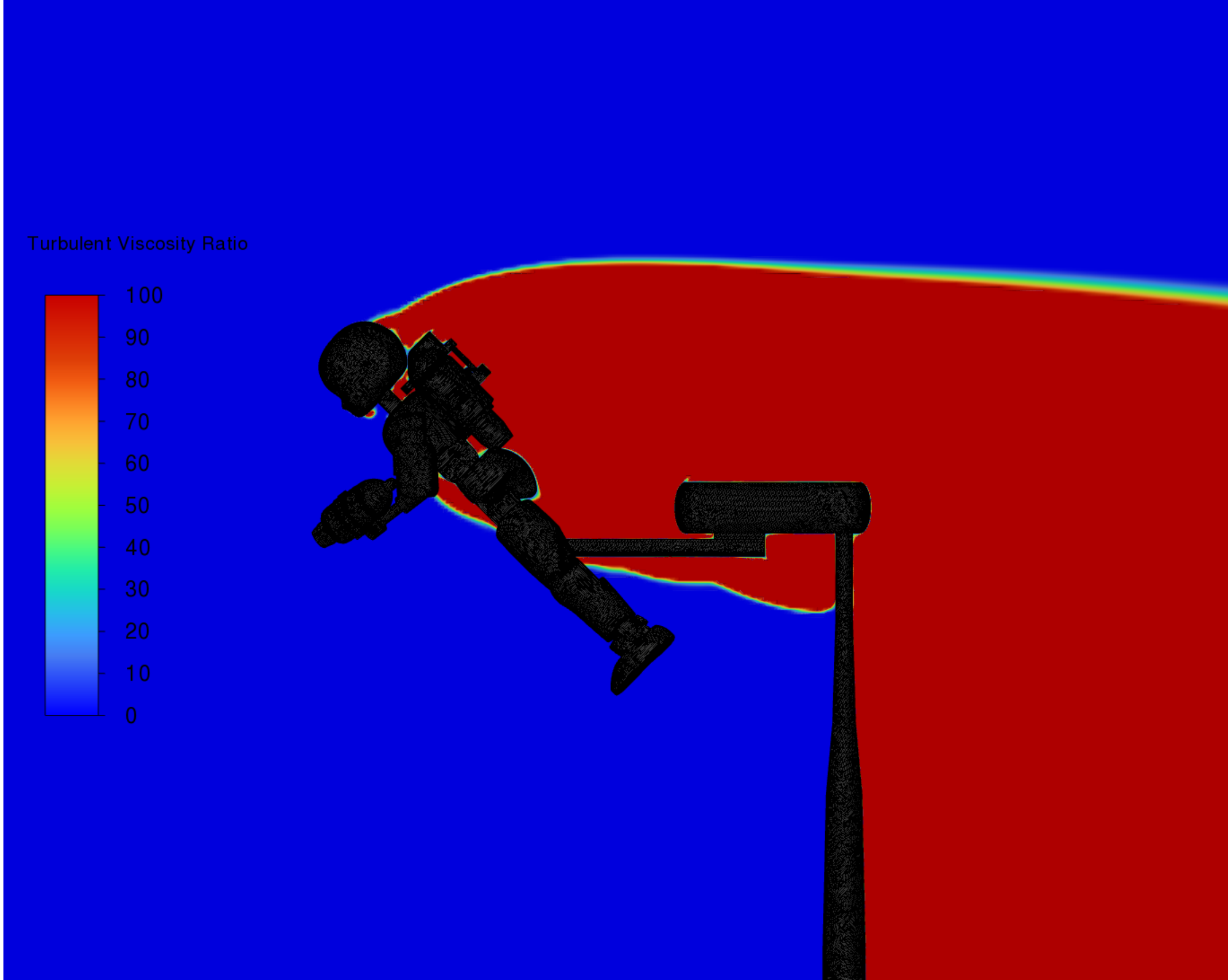}
        \caption{}
        \label{subfig:cfd-turb-contour}
    \end{subfigure}
    \\
    \begin{subfigure}[]{0.45\textwidth}
        \centering
        \adjincludegraphics[width=\textwidth,trim={{0.1\width} {0.12\height} {0.1\width} {0.2\height}},clip]{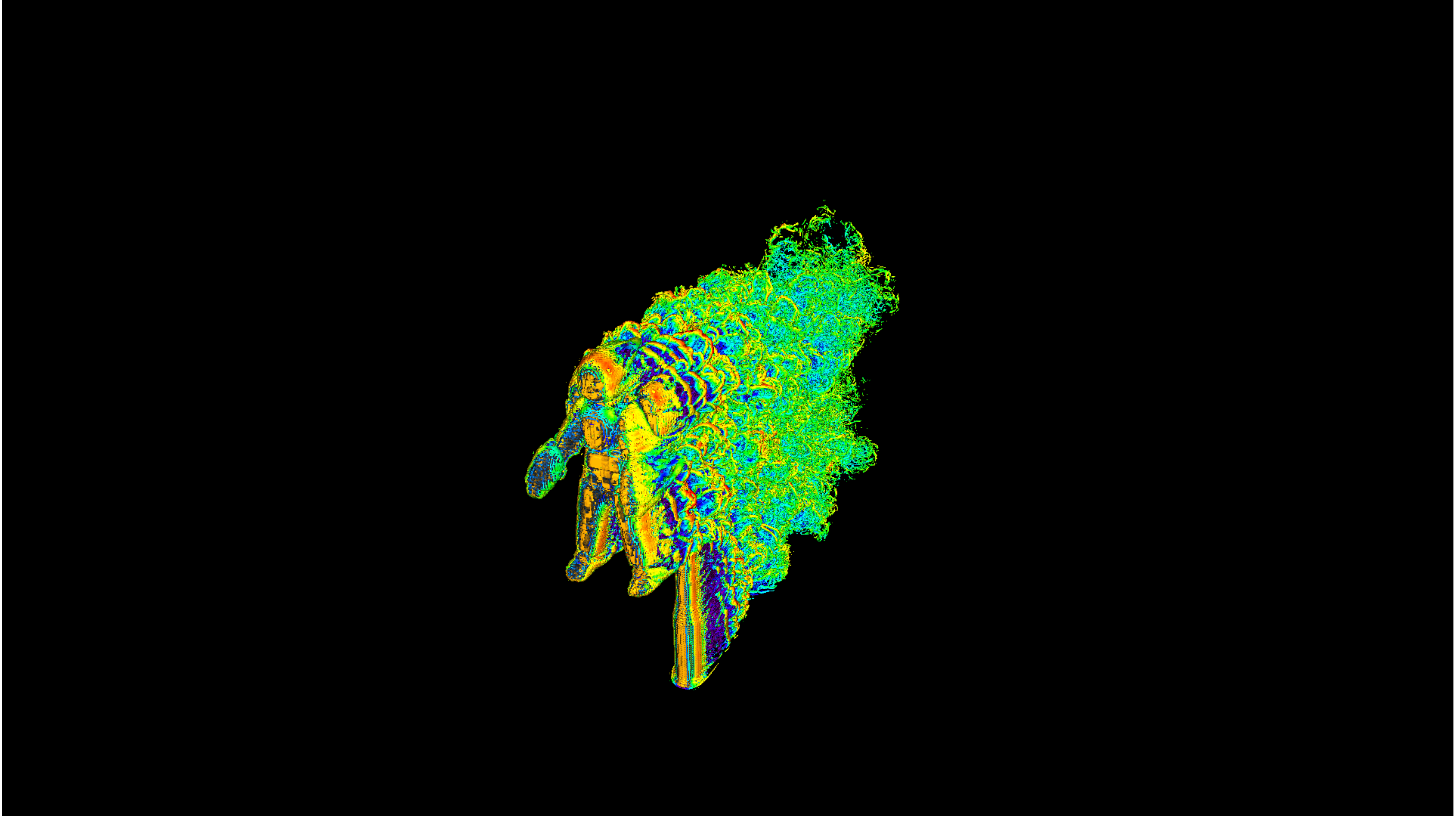}
        \caption{}
        \label{subfig:cfd-lbm-qcrit}
    \end{subfigure}
    \hspace{0.5cm}
    \begin{subfigure}[]{0.45\textwidth}
        \centering
        \adjincludegraphics[width=\textwidth,trim={{0.00\width} {0.00\height} {0.00\width} {0.00\height}},clip]{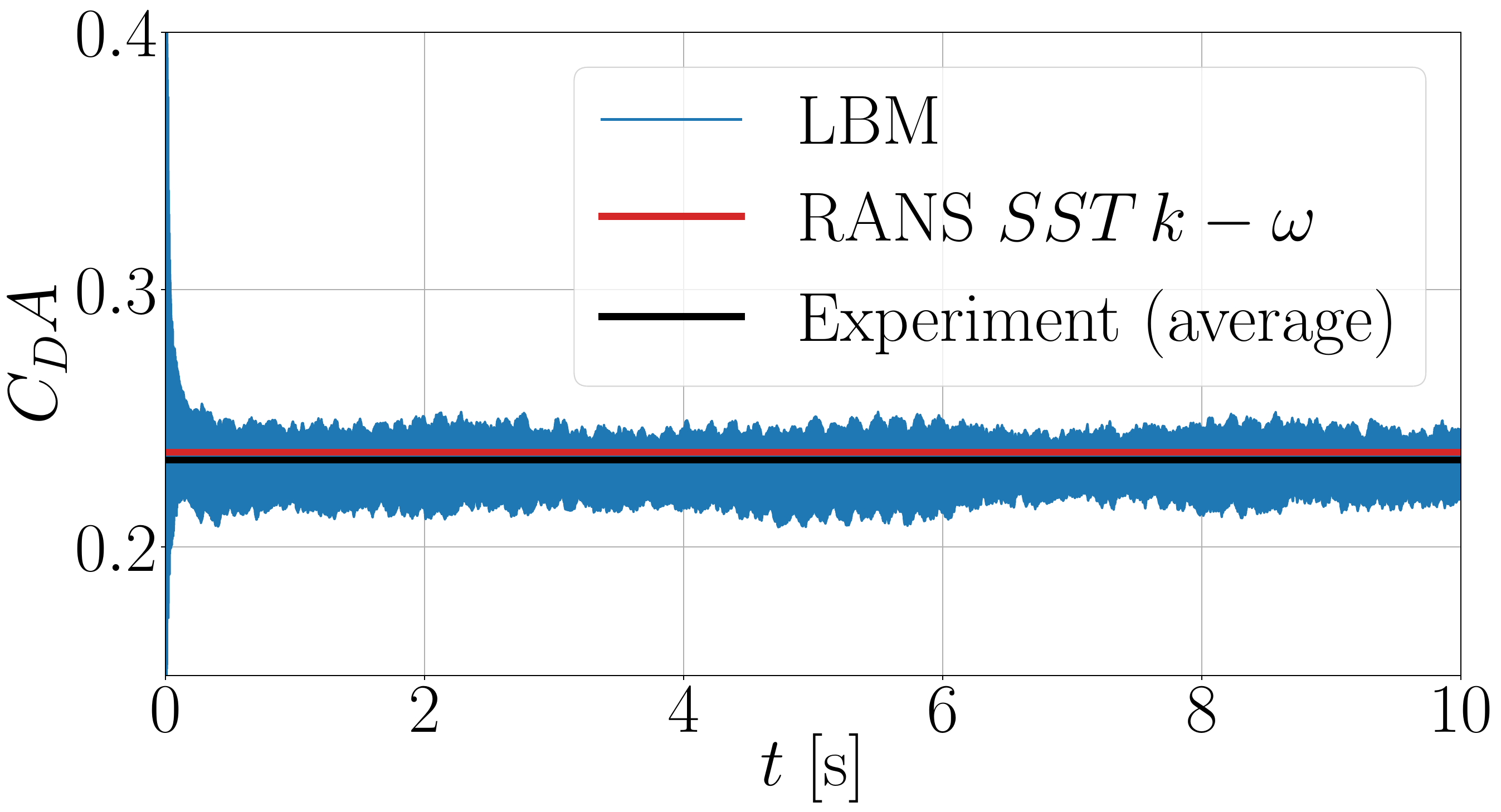}
        \caption{}
        \label{subfig:cfd-lbm-plot}
    \end{subfigure}
    \caption{\textbf{iRonCub aerodynamics CFD simulations.} Validation of RANS simulations on iRonCub in wind tunnel in \texttt{flight30} configuration: (\subref{subfig:fl30-2-CFD-WT-CdA}) $C_DA$, (\subref{subfig:fl30-2-CFD-WT-ClA}) $C_LA$. RANS simulations with \textit{SST $k-\omega$} turbulence model results for iRonCub in \texttt{flight30} configuration at $\alpha=\qty{45}{\degree}$: (\subref{subfig:cfd-robot-press}) pressure on robot surface, (\subref{subfig:cfd-press-contour},\subref{subfig:cfd-vel-contour},\subref{subfig:cfd-turb-contour}) longitudinal plane contours of pressure variation, velocity magnitude, and turbulent/laminar viscosity ratio. Validation of LBM simulation of iRonCub in wind tunnel setup in \texttt{hovering} configuration at $\beta=\qty{0}{\degree}$: (\subref{subfig:cfd-lbm-qcrit}) Q-criterion iso-surface, (\subref{subfig:cfd-lbm-plot}) comparison of $C_DA$ with RANS simulations and wind tunnel experiments.}
    \label{fig:3}
\end{figure}

In \crefrange{subfig:cfd-robot-press}{subfig:cfd-turb-contour}, we present the outcomes of a simulation from this first set. The results indicate a large separated flow region behind the robot, a characteristic feature of bluff bodies. Specifically, the velocity magnitude contour \cref{subfig:cfd-vel-contour} highlights a significant low-velocity region situated at the robot's rear. This observation is further supported by the corresponding low-pressure region depicted in the pressure variation contour \cref{subfig:cfd-press-contour}. Of particular interest is  \cref{subfig:cfd-turb-contour}, depicting a contour of the turbulent/laminar viscosity ratio ($\nu_t/\nu$), which is an indicator of the importance of turbulence activity in the RANS equations. Values exceeding 100 throughout the wake of the robot and on its surface indicate fully turbulent conditions, corroborating the choice of the selected turbulence models. Examining \cref{subfig:cfd-robot-press,subfig:cfd-press-contour} we observe how the significant separation leads to an imbalance in the pressure distributions between the front and rear surfaces of the robot, resulting in a higher aerodynamic drag force for bluff bodies compared to streamlined bodies. In this simulation the pressure drag accounts for approximately $95\%$ of the total aerodynamic drag, with the remaining portion generated by the wall shear stress.

The data gathered from wind tunnel experiments has been utilized to evaluate the accuracy and reliability of the CFD simulations. \cref{subfig:fl30-2-CFD-WT-CdA,subfig:fl30-2-CFD-WT-ClA} shows a comparison between the collected wind tunnel data and the corresponding CFD simulations with the \texttt{flight30} robot configuration at an airspeed of $|v_a| = \qty{17}{\meter/\second}$. It was observed that the average error in the magnitude of aerodynamic force area remains below 10\%, with higher fidelity when using \textit{SST $k-\omega$} model for CFD simulations, especially concerning the drag force. Consequently, we decided to employ this model for the second set of simulations carried on to augment our aerodynamics dataset.

As a further assessment, we chose to compare our adopted CFD simulation approach (Finite Volume Method) with a turbulent resolving simulation method, the Lattice-Boltzmann Method (LBM). Therefore, we conducted an LBM simulation of iRonCub within the wind tunnel setup using FluidX3D software \cite{lehmann2023computational}. This simulation utilized 2$\times$NVIDIA A100 PCIe 80GB GPUs (totaling 160GB of VRAM) and ran for 2 days, yielding the results detailed in \cref{subfig:cfd-lbm-qcrit,subfig:cfd-lbm-plot}. The LBM simulation demonstrated high accuracy in computing the total drag force on the robot (error $\leq$ 1\%) and offered the capability to modify the geometry in real-time.
However, these simulations have two main limitations that make them unusable for collecting a large aerodynamic dataset: the first is the high computational expense associated with a single LBM simulation, and the second is the equally-spaced grid, which implies that it's not possible to reduce the number of mesh elements without a rapid degradation of the quality of the robot geometry.

\subsection{Performances of the Aerodynamic Models}\label{results/aerodynamic_models}

\begin{figure}[htpb]
    \centering
    \begin{subfigure}[]{0.49\textwidth}
        \centering
        \includegraphics[width=\textwidth]{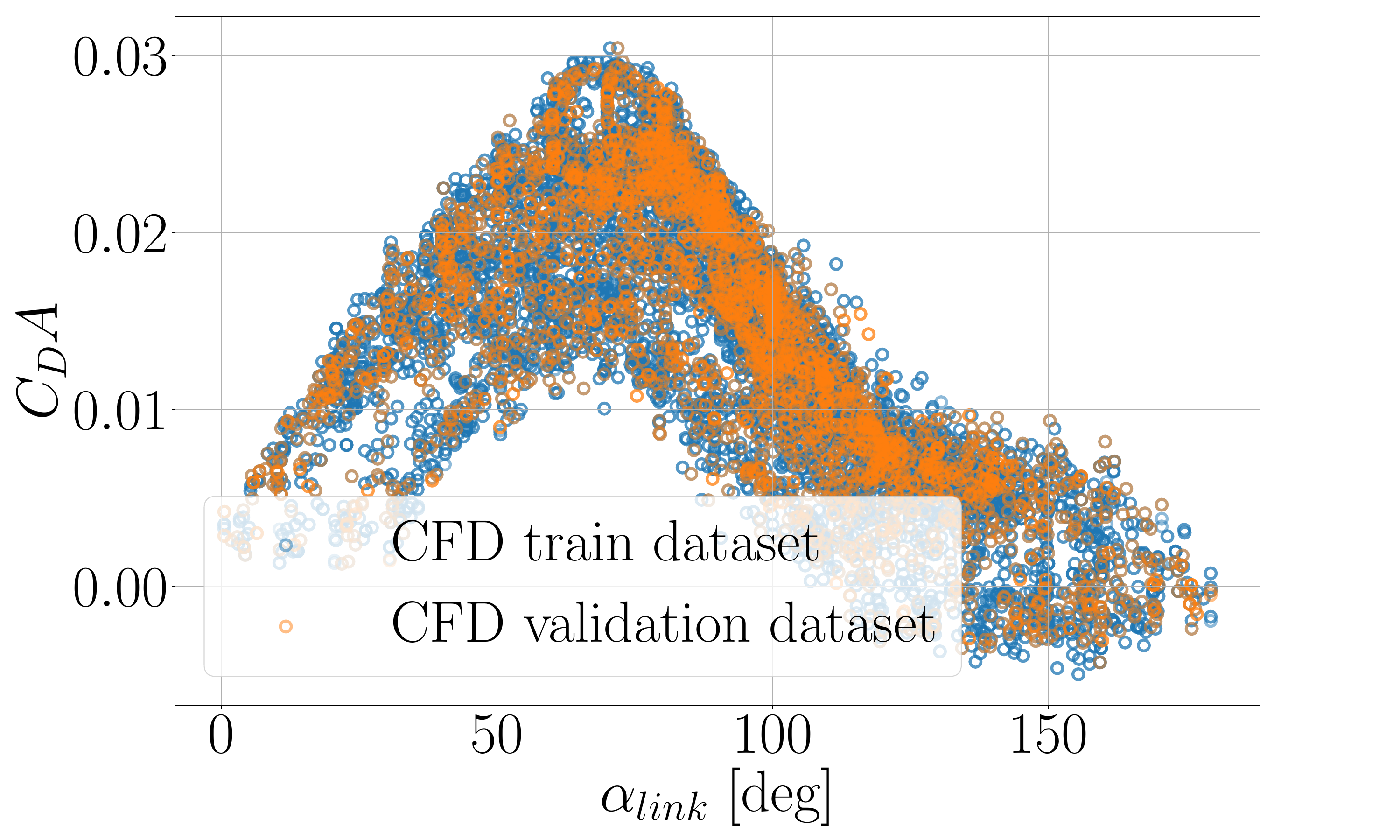}
        \caption{}
        \label{subfig:head-nn-dataset}
    \end{subfigure}
    \hfill
    \begin{subfigure}[]{0.49\textwidth}
        \centering
        \includegraphics[width=\textwidth]{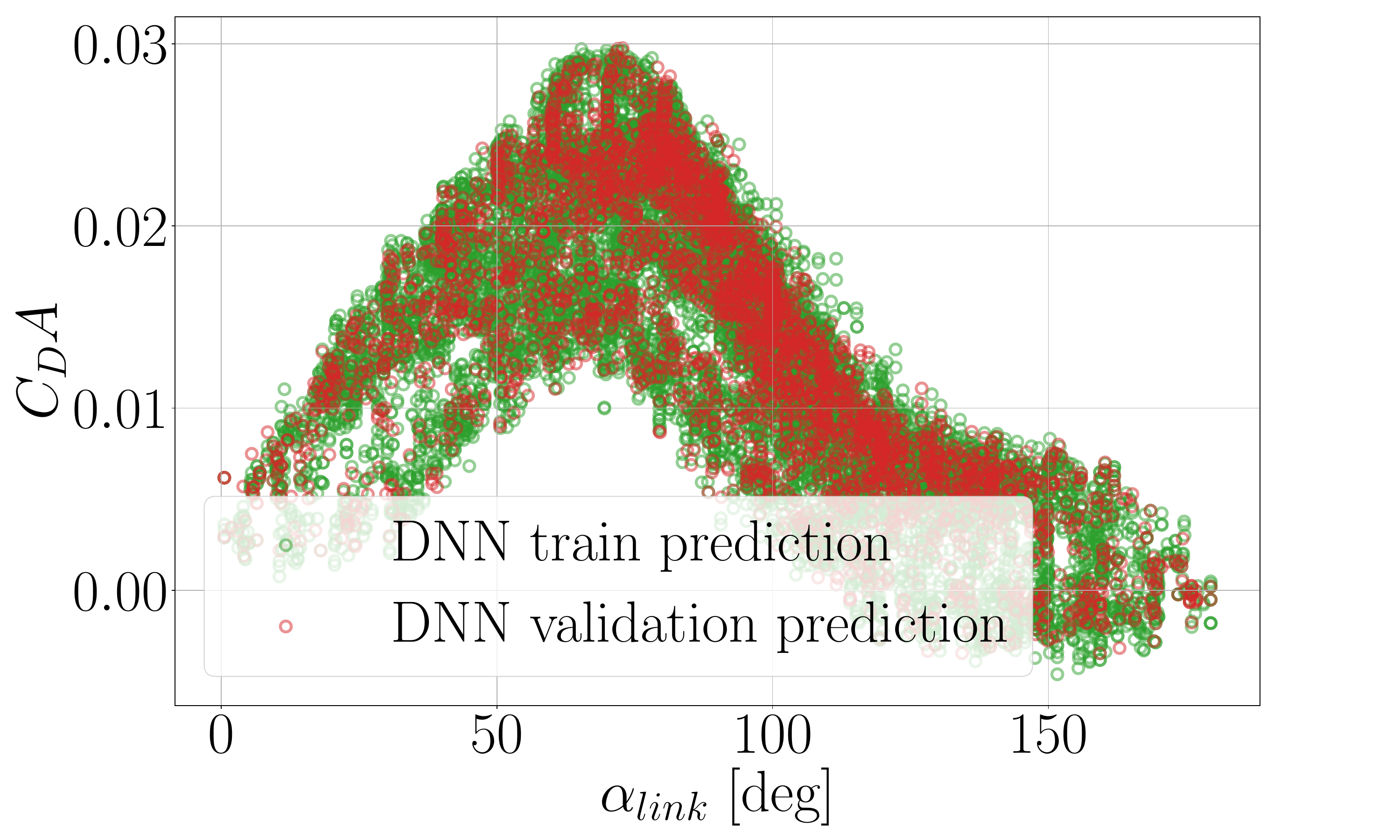}
        \caption{}
        \label{subfig:head-nn-predict}
    \end{subfigure}
    \\
    \begin{subfigure}[]{0.49\textwidth}
        \centering
        \includegraphics[width=\textwidth]{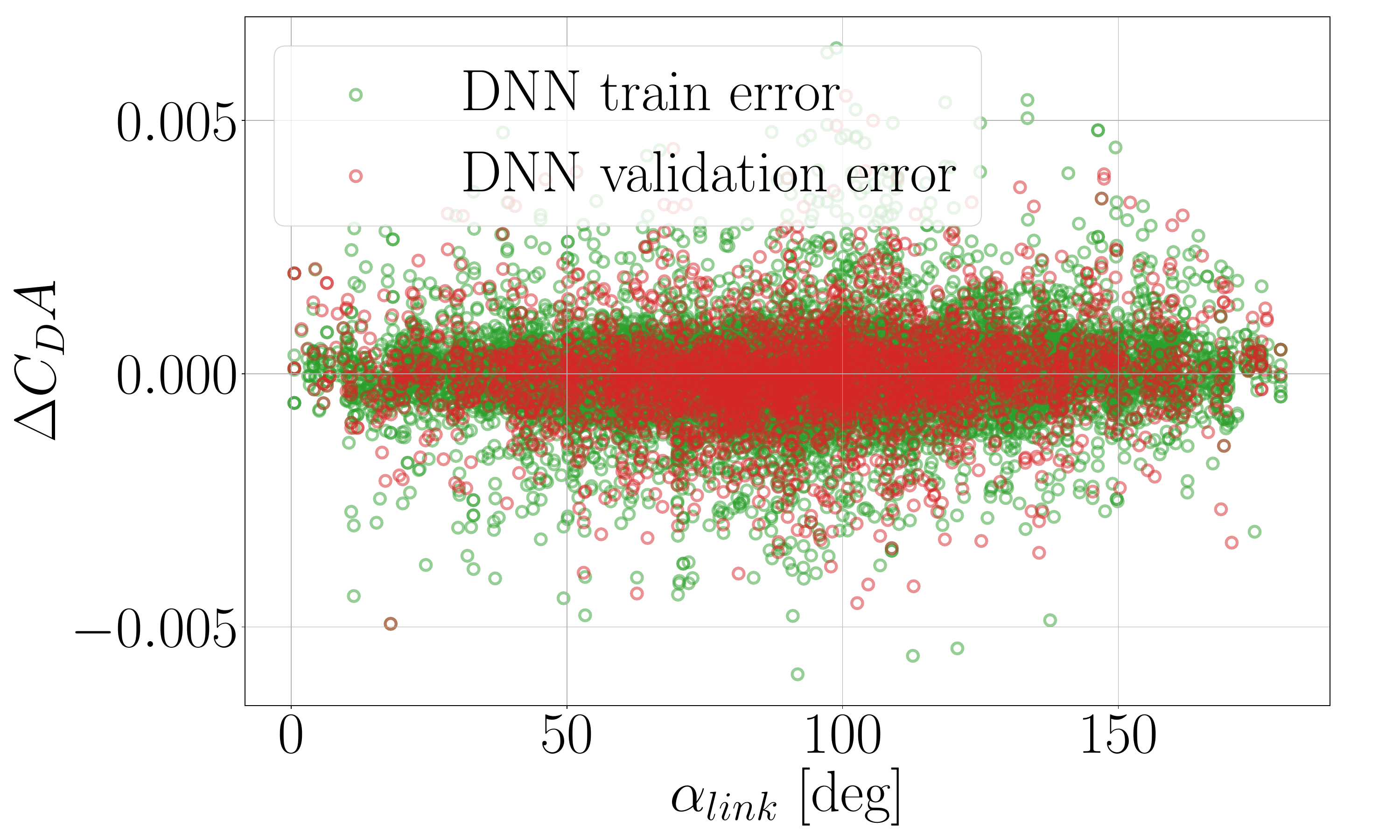}
        \caption{}
        \label{subfig:head-nn-error}
    \end{subfigure}
    \\
    \begin{subfigure}[]{0.49\textwidth}
        \centering
        \includegraphics[width=\textwidth]{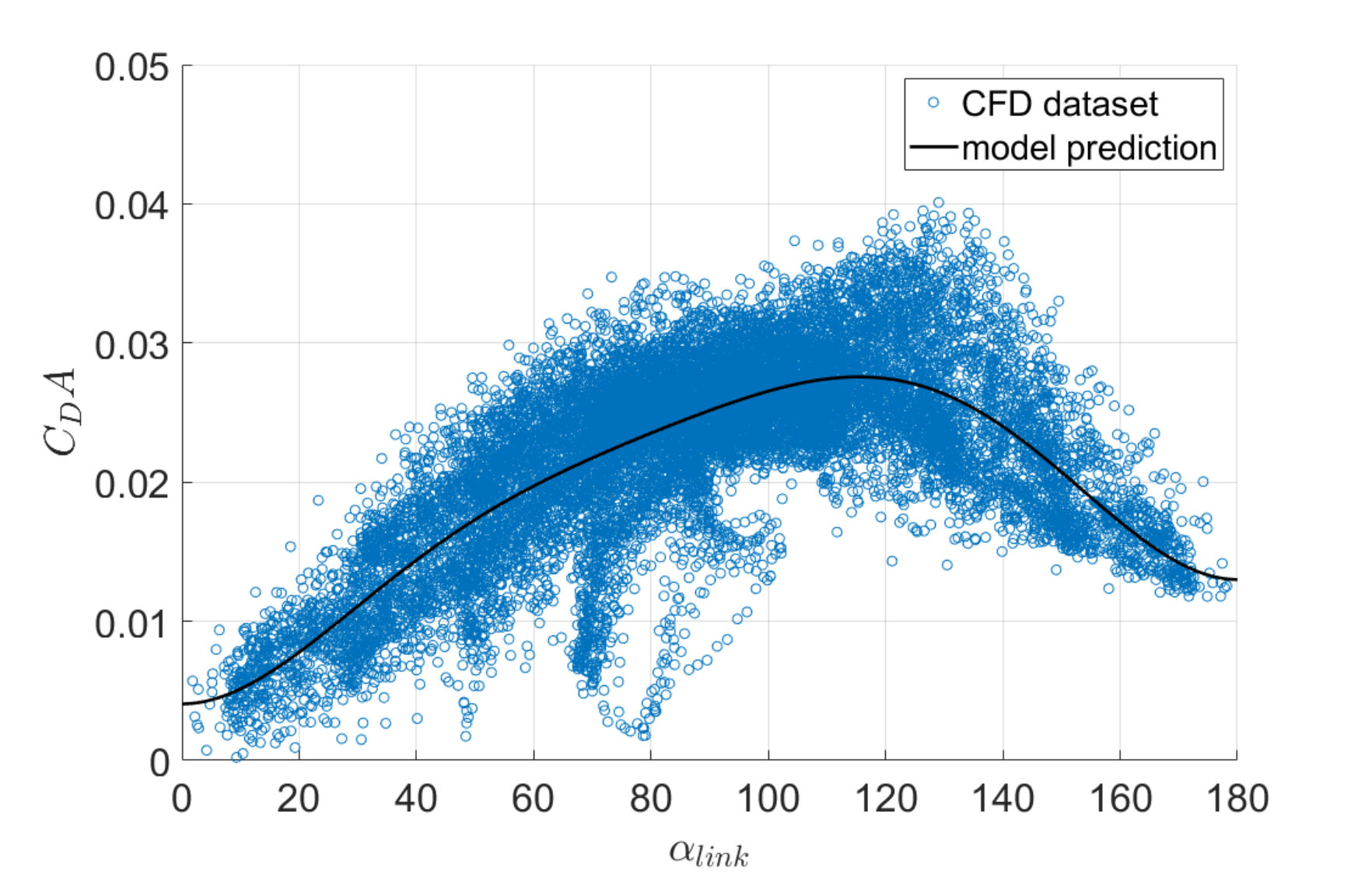}
        \caption{}
        \label{subfig:lower_leg-poly-CdA}
    \end{subfigure}
    \hfill
    \begin{subfigure}[]{0.49\textwidth}
         \centering
         \includegraphics[width=\textwidth]{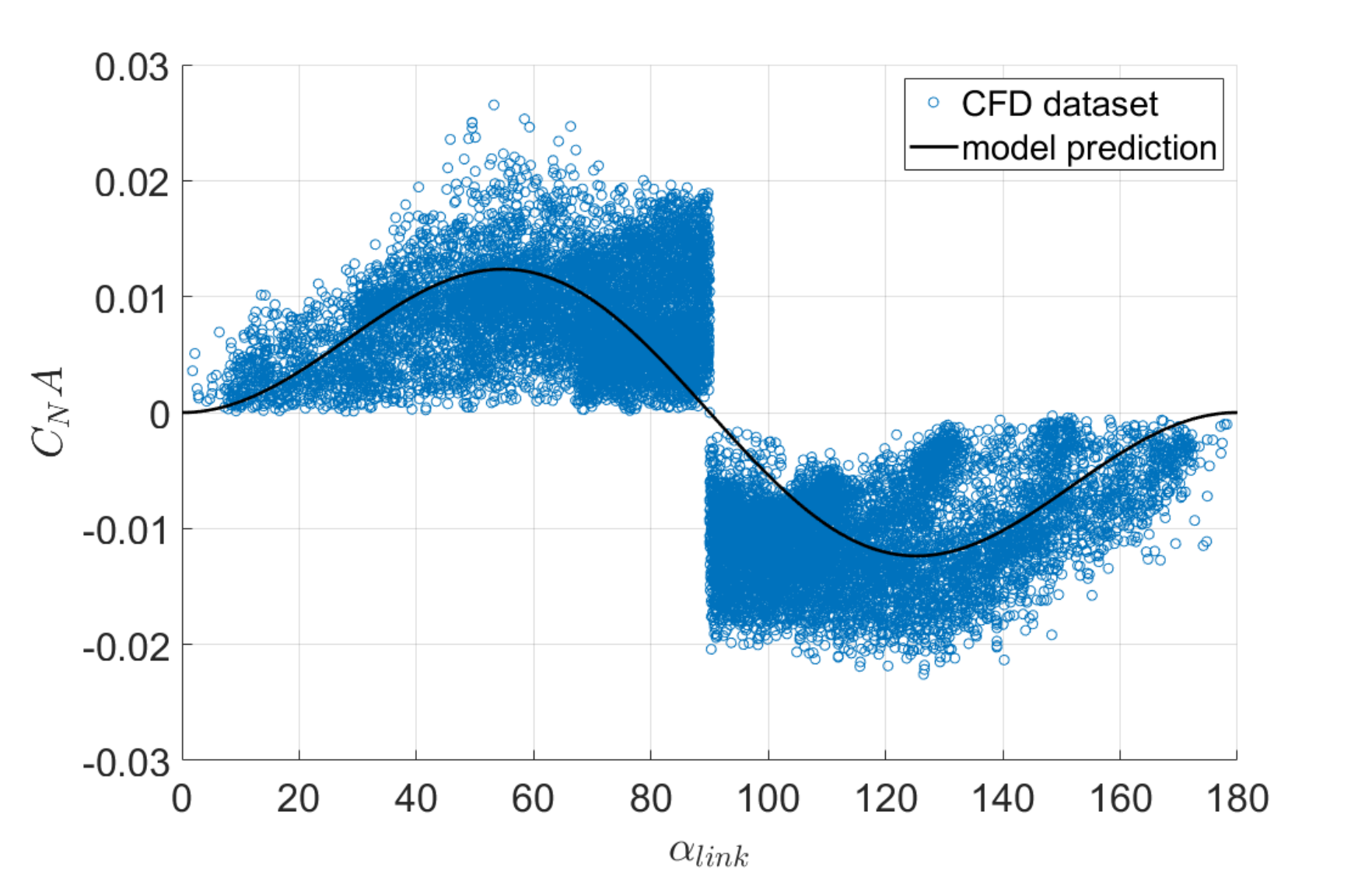}
         \caption{}
         \label{subfig:lower_leg-poly-CnA}
    \end{subfigure}
    \caption{\textbf{Aerodynamic models for simulation and control.} Deep Neural Network (DNN): (\subref{subfig:head-nn-dataset}) datasets, (\subref{subfig:head-nn-predict}) predictions, and (\subref{subfig:head-nn-error}) errors for the drag force area ($C_D A$) on the head link. Axisymmetric Model: datasets and predictions for the drag (\subref{subfig:lower_leg-poly-CdA}) and lift (\subref{subfig:lower_leg-poly-CnA}) force areas on the lower leg link. The DNN plots show the results as functions of the local angle of attack for better visualization.}
    \label{fig:4}
\end{figure}

The data from the second set of CFD simulations is used for creating aerodynamic models integrated into both the simulator and the controller, allowing real-time estimation of aerodynamic forces. First, we employed Deep Neural Network to construct the simulator model. Then, we developed a simpler model using linear regression, to be integrated into the robot controller.

\subsubsection*{Deep Neural Network for Aerodynamic Forces Prediction} \label{results/aerodynamic_models/DNN}

We developed a Deep Neural Network (DNN) to correlate inputs from the dataset, comprising the robot's joint positions and the wind-robot relative velocity direction, with the three aerodynamic force components acting on each robot link's Center of Mass (CoM). 

After training, the DNN show a maximum RME of $3\times10^{-6}$ when predicting aerodynamic forces on the validation set.
In \crefrange{subfig:head-nn-dataset}{subfig:head-nn-error}, we present a comparison between dataset values and predictions, along with the absolute error, for both the validation and training set data concerning the drag area of the robot's head. The validation error remains limited below 20\% of the maximum value, indicating not only a very low average error, but also a limited local error on the single samples.

\subsubsection*{Axisymmetric Links Model for Aerodynamic Forces Prediction}\label{results/aerodynamic_models/axisym_model}

To address the DNN model's reliance on extensive data and the difficulty of interpreting the model itself, which basically acts as a black box, we also derived a simpler model. This model serves a dual purpose: to enhance interpretability, and to facilitate integration into the controller for balancing and flight experiments, enabling performance comparison with the DNN. The model is constructed from the same CFD dataset of the DNN, but we introduced the following assumptions: 
i) The aerodynamic behavior of the robot links is steady and axisymmetric.
ii) The aerodynamic interference effects resulting from close links or separated flow are negligible.
This Axisymmetric Model is based on simple mathematical expressions retaining physical properties such as continuity and zero conditions. To fit the model coefficients, we employed linear regression with Lasso regularization to resolve redundancy. Results depicted in \cref{subfig:lower_leg-poly-CdA,subfig:lower_leg-poly-CnA} on the lower leg link demonstrate that this model follows, approximately, the average value of the actual data, resulting in a one order of magnitude higher RME than the DNN. Despite the lower prediction accuracy, the model maintains a physically coherent representation of aerodynamic forces.

\subsection{Validation on iRonCub robot}\label{results/sim_exp}

The procedure for modeling and estimating aerodynamic forces is validated through balancing experiments on the real iRonCub and flight simulations, both performed using the aerodynamic-aware momentum-based controller presented in the Methods section. The balancing experiments validate the methodology's reliability and robustness when implemented on an actual humanoid robot.

\subsubsection*{Flight Simulations}\label{results/sim_exp/simulations}

\begin{figure}[htpb]
    \hspace{0.65cm}
    \begin{subfigure}[b]{0.57\textwidth}
        \centering
        \adjincludegraphics[width=\textwidth,trim={{0.05\width} {0.00\height} {0.10\width} {0.00\height}},clip]{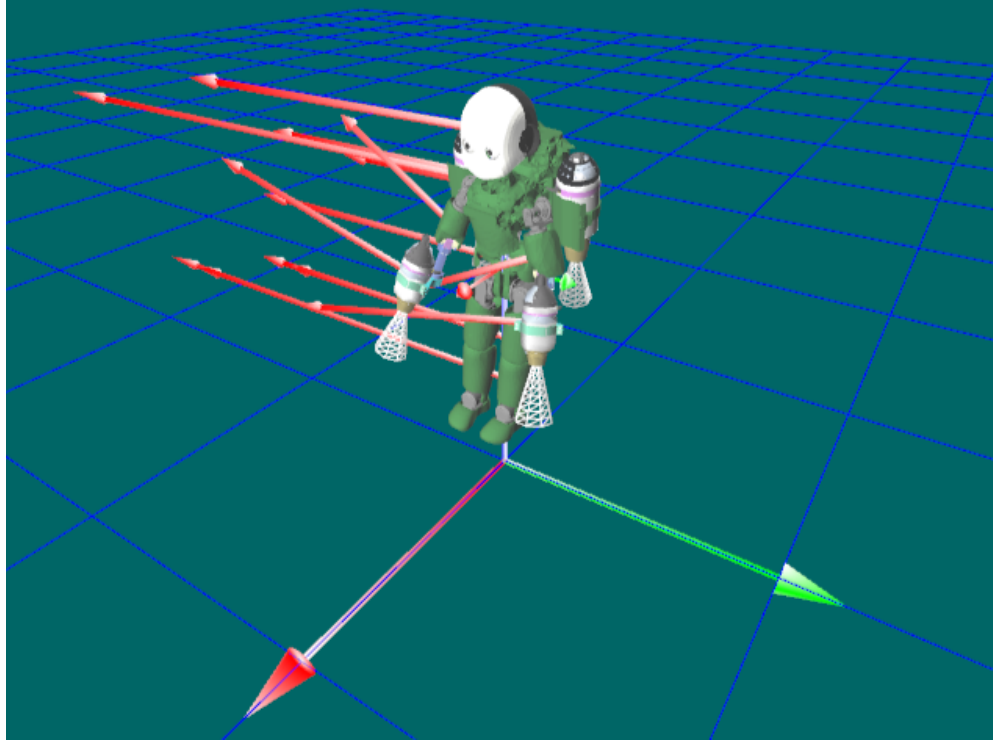}
        \caption{}
        \label{subfig:iRonCub-sim-iso-view}
    \end{subfigure}
    \hfill
    \begin{subfigure}[b]{0.32\textwidth}
         \begin{subfigure}[b]{\textwidth}
         \adjincludegraphics[width=\textwidth,trim={{0.00\width} {0.05\height} {0.0\width} {0.05\height}},clip]{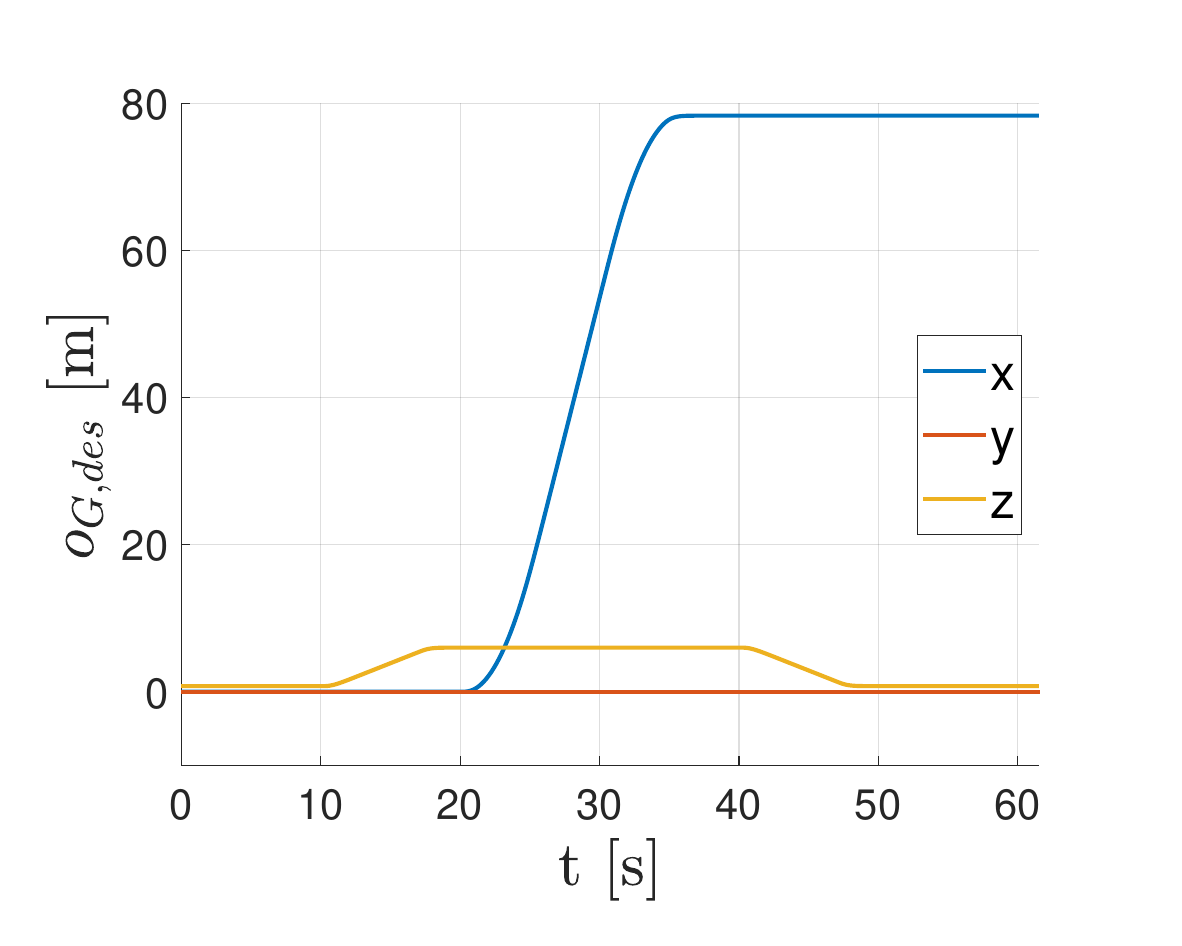}
         \caption{}
         \label{subfig:sim-CoM-pos-des}
         \end{subfigure}
        \\
        \begin{subfigure}[b]{\textwidth}
         \adjincludegraphics[width=\textwidth,trim={{0.00\width} {0.05\height} {0.0\width} {0.05\height}},clip]{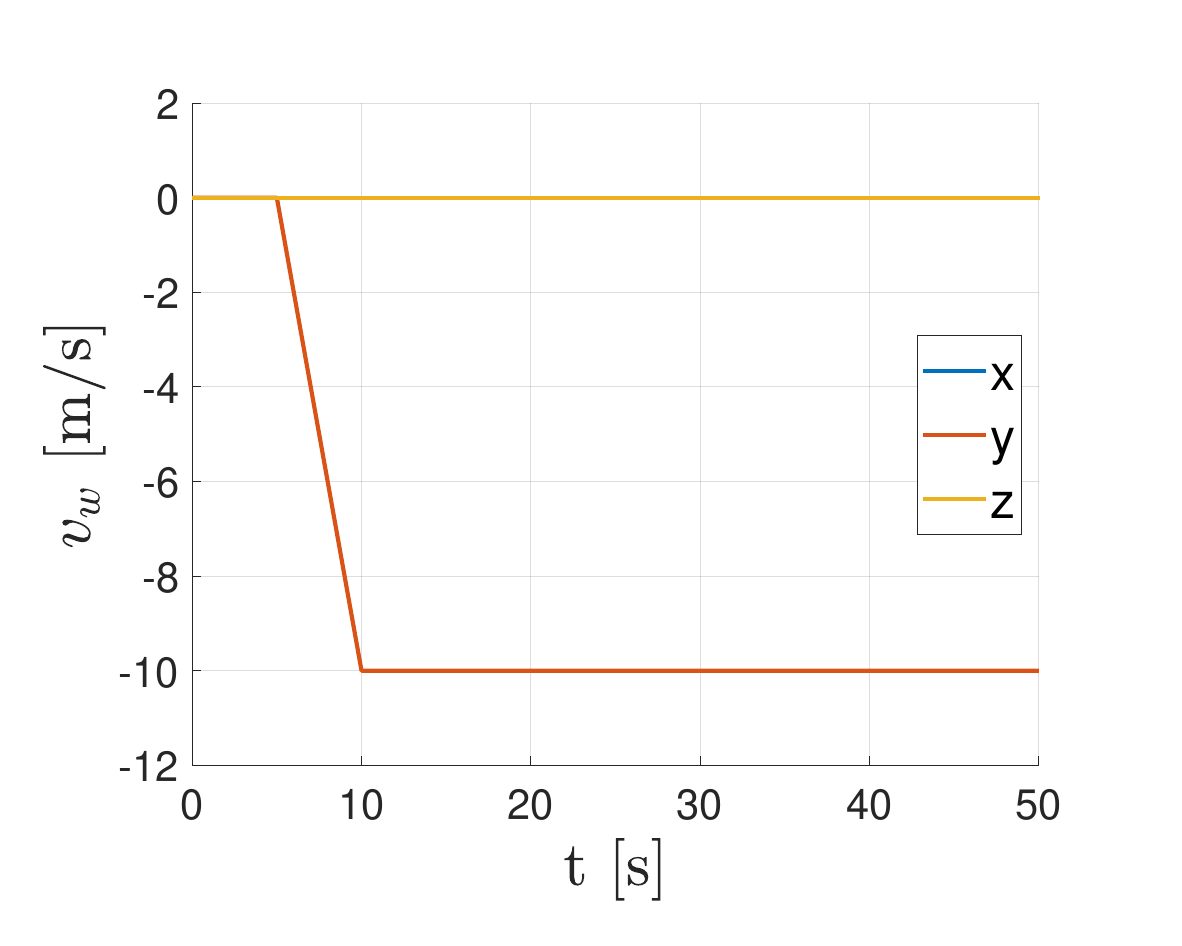}
         \caption{}
         \label{subfig:sim-wind-velocity}
         \end{subfigure}
    \end{subfigure}
    \\
    \begin{subfigure}[b]{0.32\textwidth}
         \centering
         \adjincludegraphics[width=\textwidth,trim={{0.00\width} {0.05\height} {0.0\width} {0.05\height}},clip]{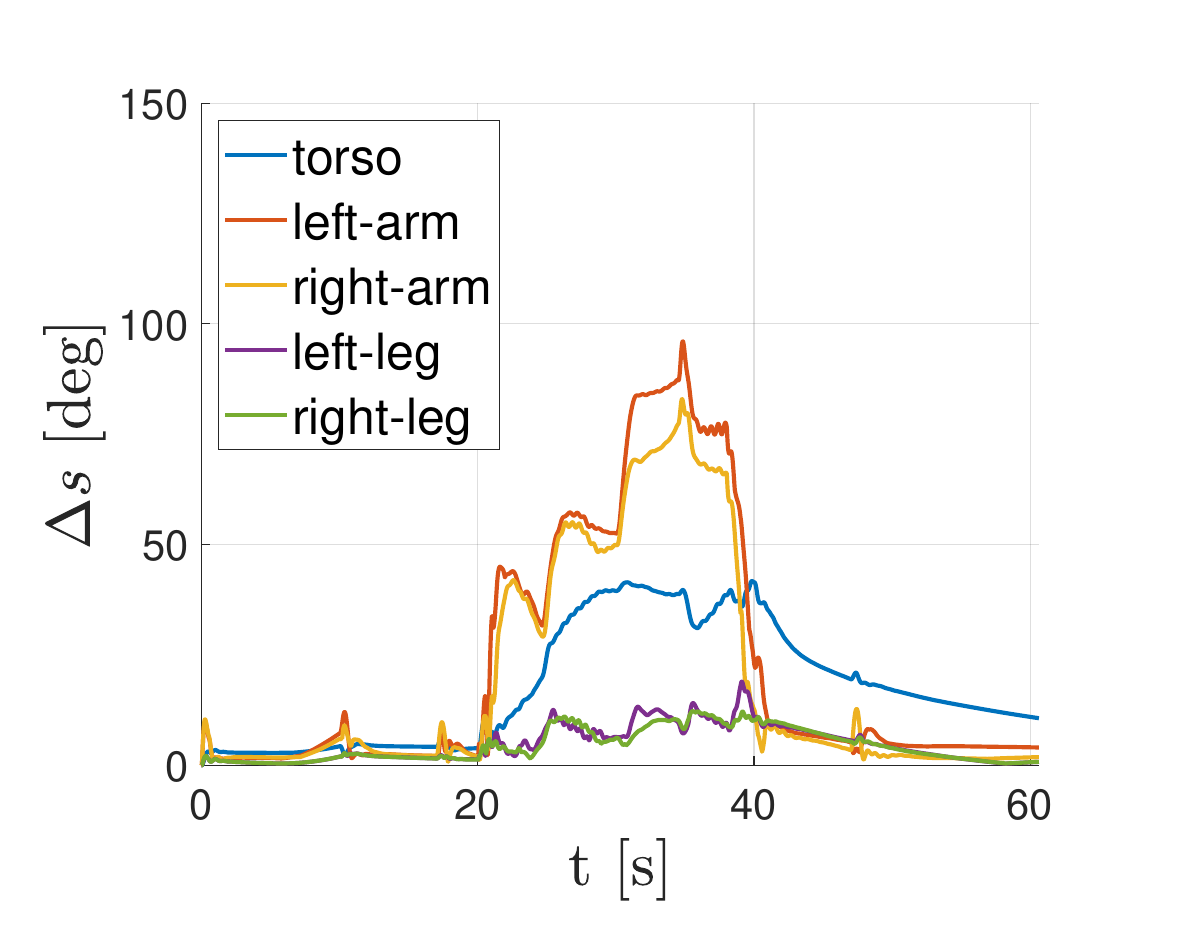}
         \caption{}
         \label{subfig:sim4-joint-pos-err}
    \end{subfigure}
    \hfill
    \begin{subfigure}[b]{0.32\textwidth}
         \centering
         \adjincludegraphics[width=\textwidth,trim={{0.00\width} {0.05\height} {0.0\width} {0.05\height}},clip]{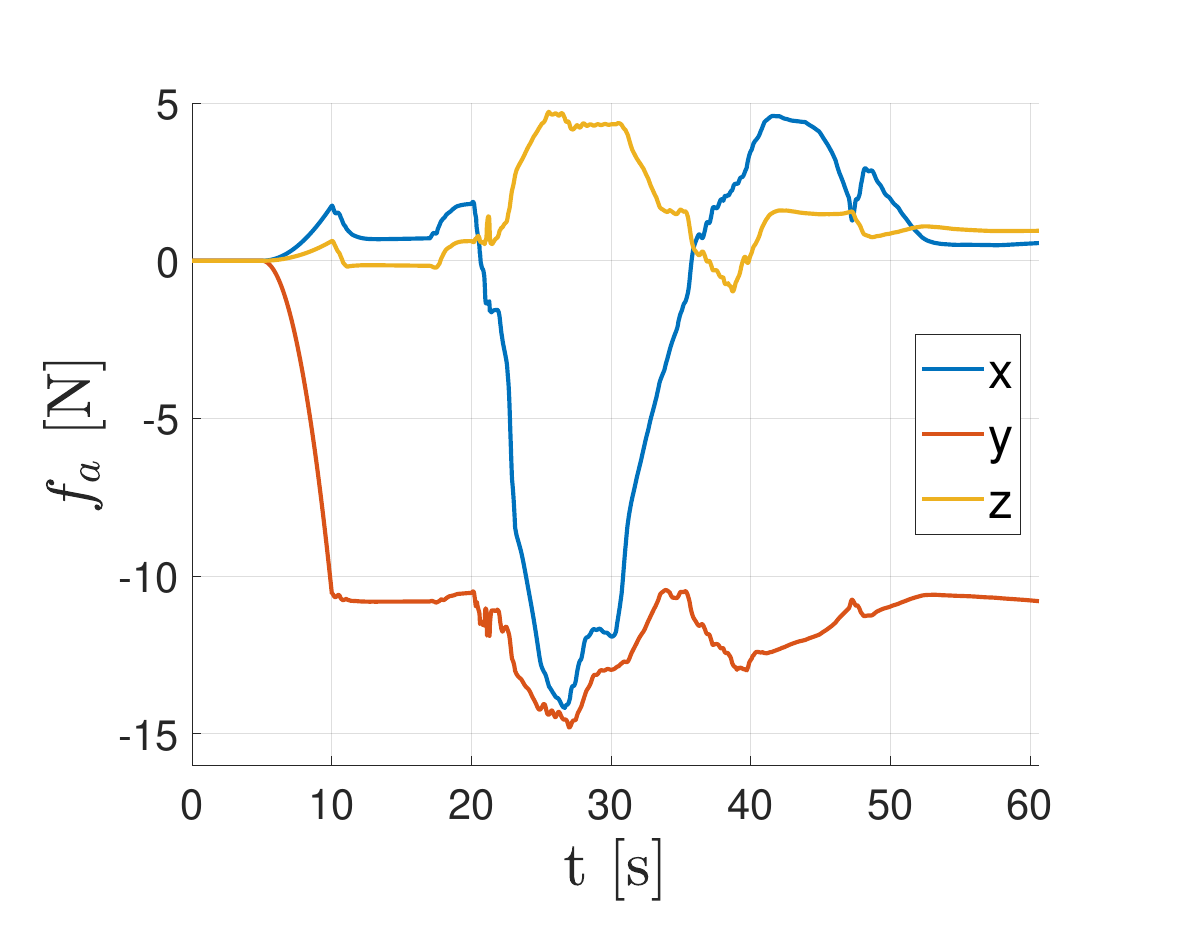}
         \caption{}
         \label{subfig:sim4-aero-forces-sim}
    \end{subfigure}
    \hfill
    \begin{subfigure}[b]{0.32\textwidth}
         \centering
         \adjincludegraphics[width=\textwidth,trim={{0.00\width} {0.05\height} {0.0\width} {0.05\height}},clip]{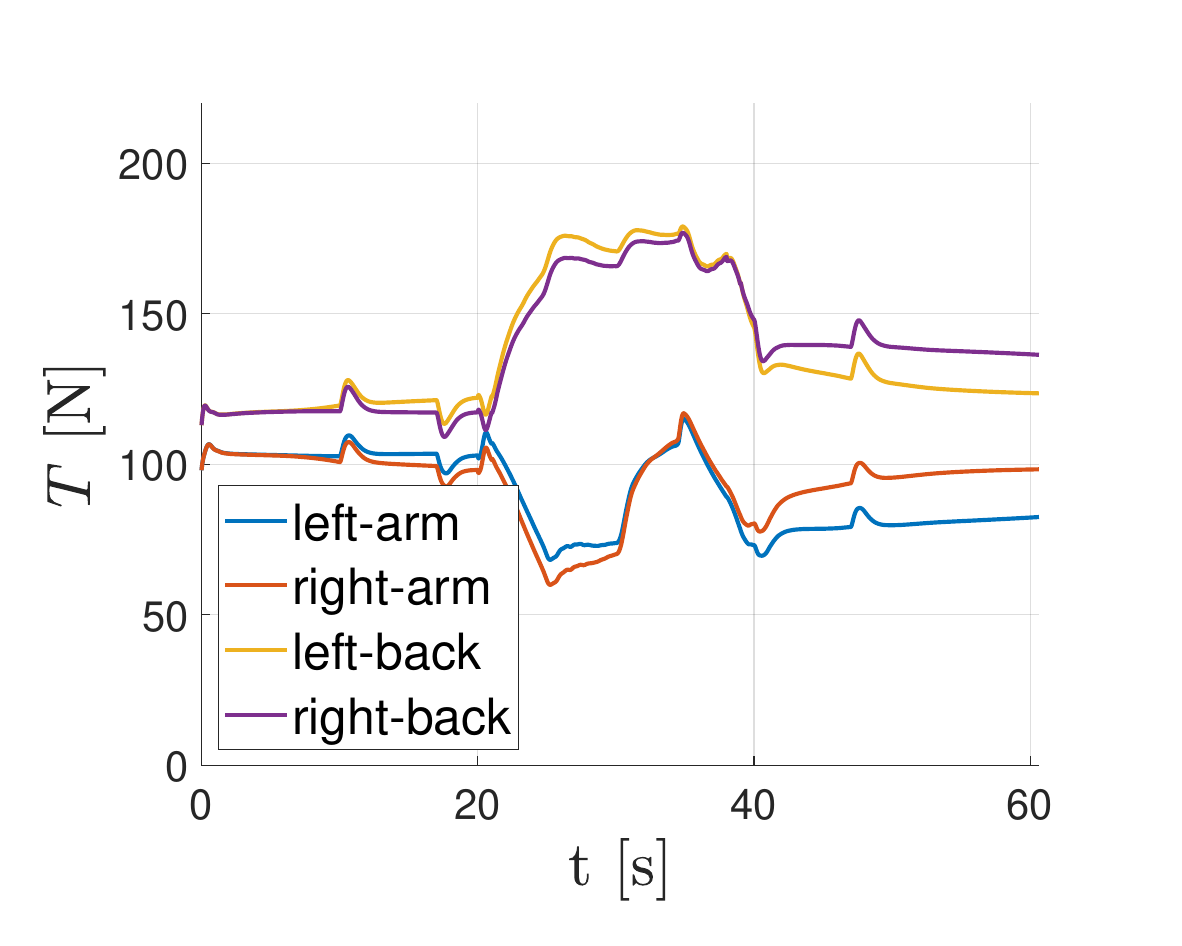}
         \caption{}
         \label{subfig:sim4-thrusts}
    \end{subfigure}
    \\
    \begin{subfigure}[b]{0.32\textwidth}
         \centering
         \adjincludegraphics[width=\textwidth,trim={{0.00\width} {0.05\height} {0.0\width} {0.05\height}},clip]{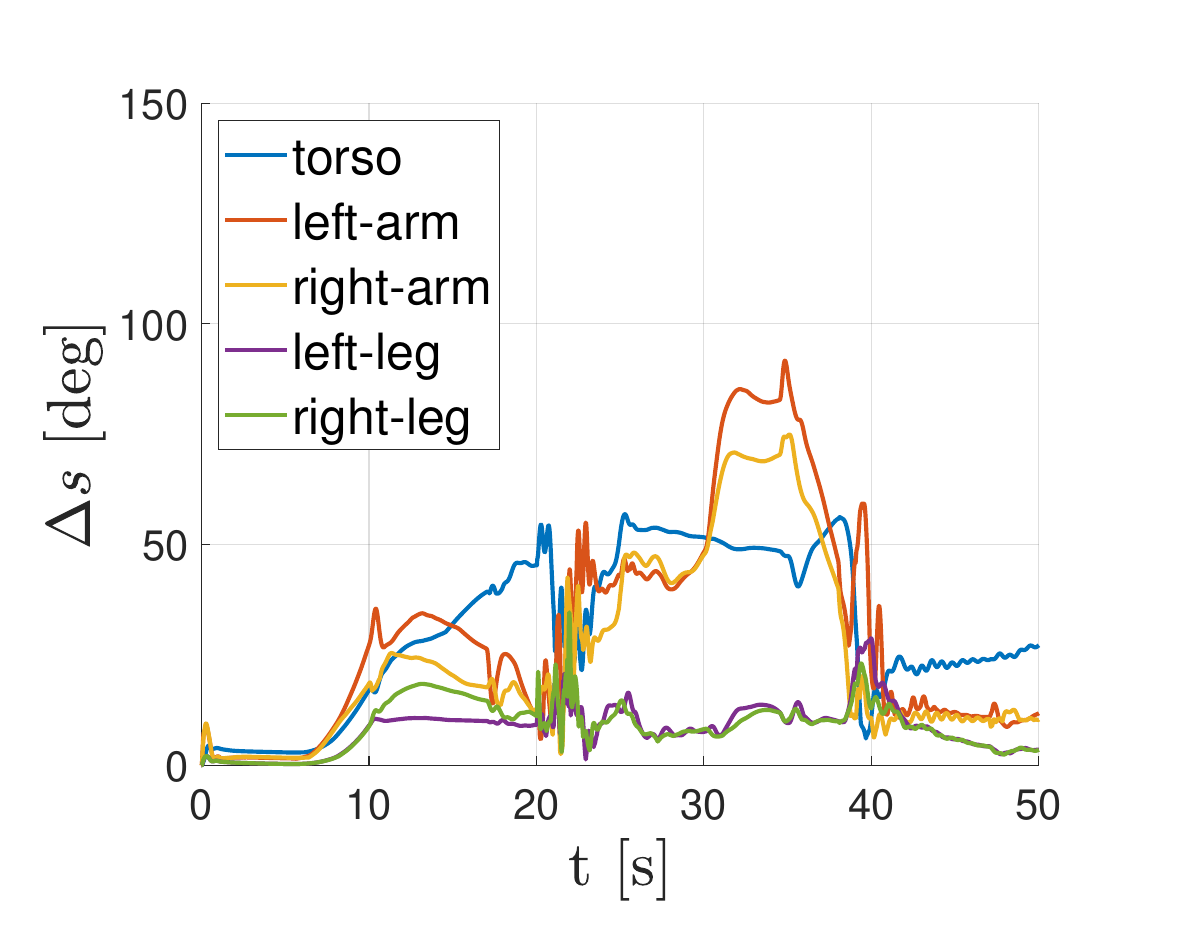}
         \caption{}
         \label{subfig:sim-rob-jointPos-err}
    \end{subfigure}
    \hfill
    \begin{subfigure}[b]{0.32\textwidth}
         \centering
         \adjincludegraphics[width=\textwidth,trim={{0.00\width} {0.05\height} {0.0\width} {0.05\height}},clip]{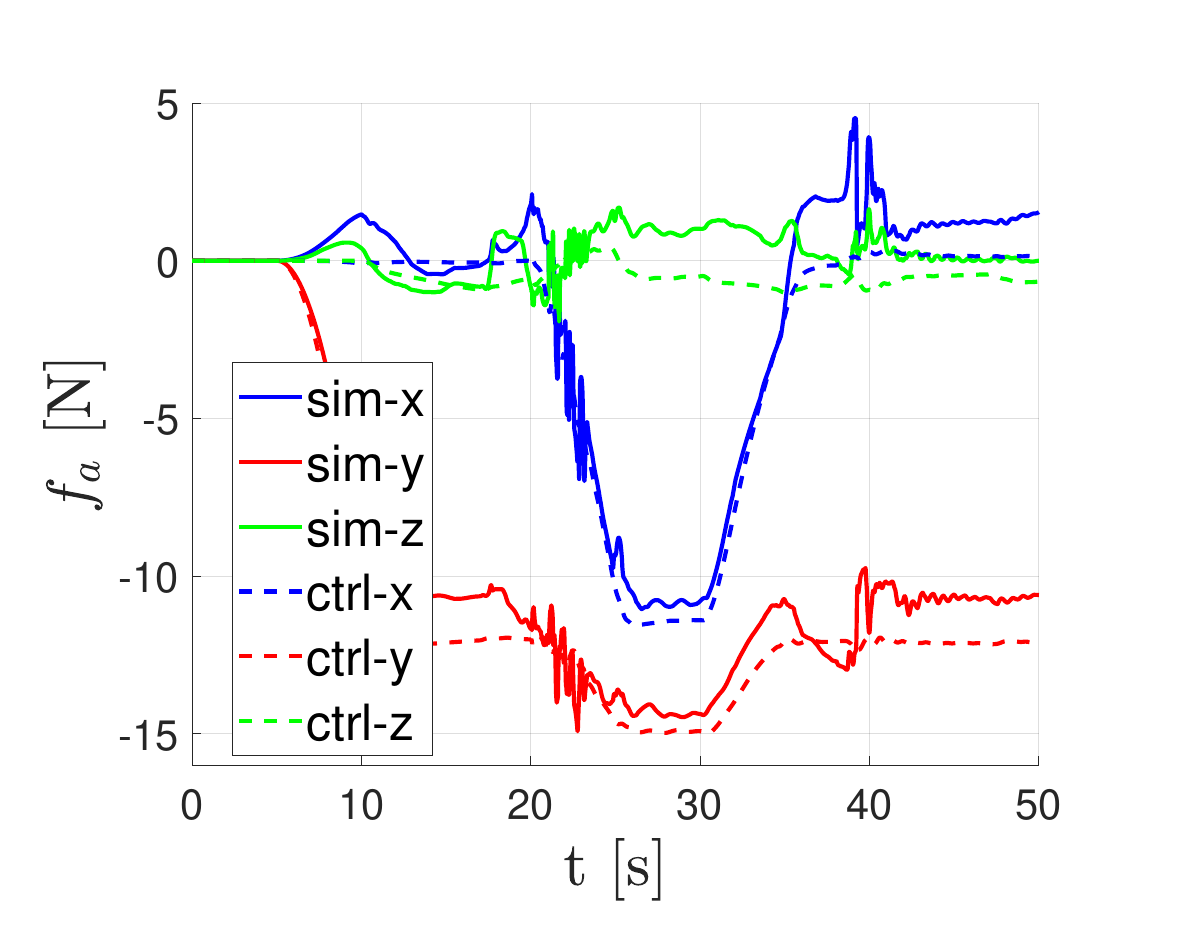}
         \caption{}
         \label{subfig:sim-rob-aero-forces}
    \end{subfigure}
    \hfill
    \begin{subfigure}[b]{0.32\textwidth}
         \centering
         \adjincludegraphics[width=\textwidth,trim={{0.00\width} {0.05\height} {0.0\width} {0.05\height}},clip]{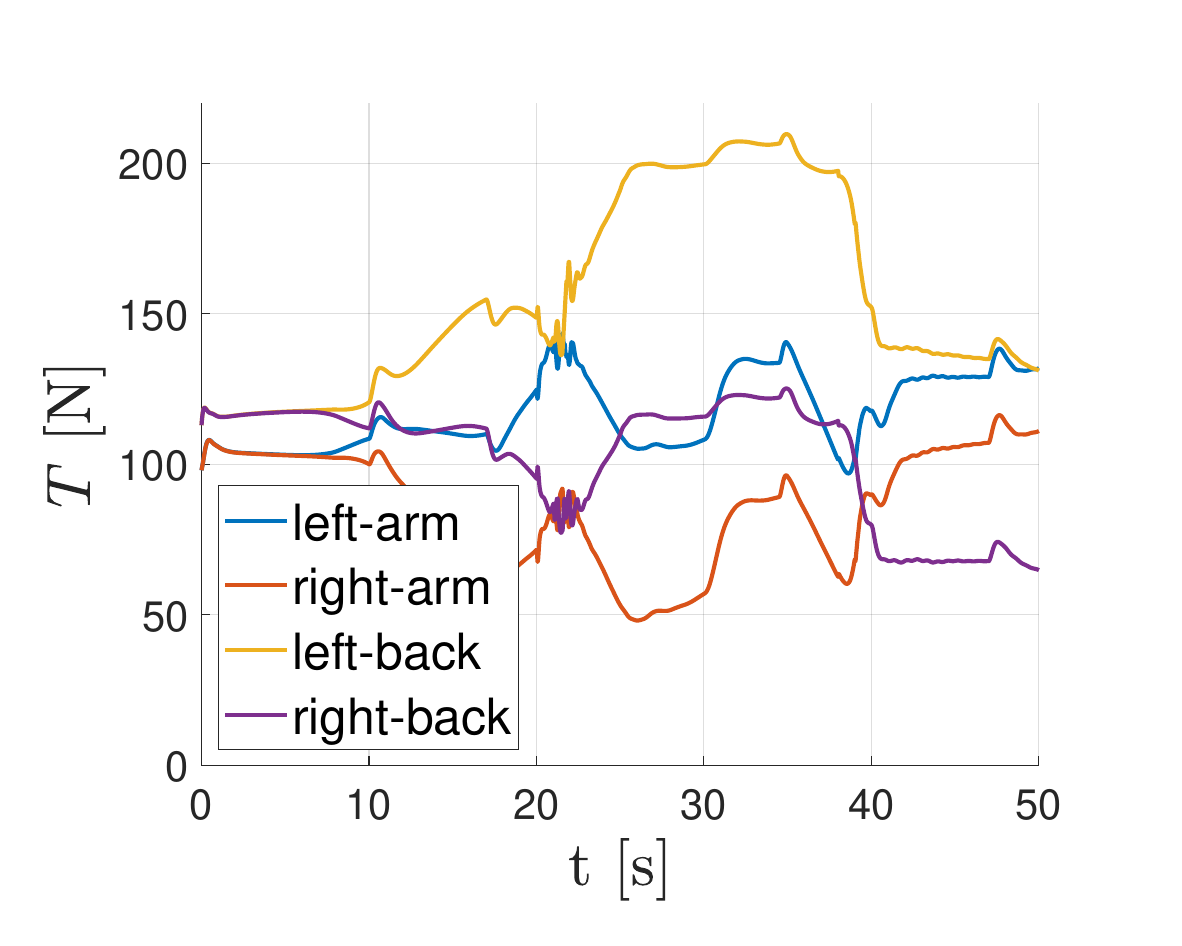}
         \caption{}
         \label{subfig:sim-rob-thrusts}
    \end{subfigure}
    \caption{\textbf{iRonCub flight simulations.} (\subref{subfig:iRonCub-sim-iso-view}) Snapshot from the \texttt{iDynTree} visualizer \cite{idyntree} of a flight simulation in the Whole-body Aerodynamic Simulator, implemented in Simulink, based on \cite{matlabwholebodysimulator}: the robot is balancing through thrust vectoring (white cones) while hovering when subject to the distributed aerodynamic forces acting on its links; the red vectors representing the distributed aerodynamic forces have been upscaled to improve visibility. (\subref{subfig:sim-CoM-pos-des}) Desired robot center of mass (CoM) position $^{\mathcal{I}} o_{\mathcal{G}}^{des}$. (\subref{subfig:sim-wind-velocity}) Wind velocity profile $^{\mathcal{I}} v_w$. Absolute error sum of joint positions $\Delta s$: (\subref{subfig:sim4-joint-pos-err}) \textsc{Test4}, (\subref{subfig:sim-rob-jointPos-err}) robustness test. Total aerodynamic force $f_a$ acting on the robot: (\subref{subfig:sim4-aero-forces-sim}) control and simulator using DNN model, (\subref{subfig:sim-rob-aero-forces}) control using Axisymmetric Model and simulator using DNN model. Jet engine thrusts commanded by the robot: (\subref{subfig:sim4-thrusts}) \textsc{Test4}, (\subref{subfig:sim-rob-thrusts}) robustness test.}
    \label{fig:5}
\end{figure}

Flight simulations confirm the necessity of compensating for aerodynamic effects when there is a non-zero relative wind velocity. Consequently, we designed the flight envelope presented in \cref{subfig:sim-CoM-pos-des}. Throughout the entire flight simulation, the robot is exposed to an external non-zero wind velocity (\cref{subfig:sim-wind-velocity}).

We conducted simulations for distinct scenarios utilizing the same flight envelope: \textsc{Test1} and \textsc{Test2} utilized the Axisymmetric Model to simulate aerodynamic forces, while \textsc{Test3} and \textsc{Test4} employed the DNN model. The tests employed different flight controllers: the Baseline Controller (\textsc{Test1} and \textsc{Test3}), which neglects aerodynamic effects, and the Aerodynamic-aware Controller (\textsc{Test2} and \textsc{Test4}), which uses feedback from the same aerodynamic forces model used by the corresponding simulator.

Results in Supplementary Figs.~7a to 7c and Supplementary Figs.~7g to 7i show that the Baseline Controller fails to complete the flight envelope, leading to the robot falling due to imbalances. Instead, tests with the Aerodynamic-aware Controller (\crefrange{subfig:sim4-joint-pos-err}{subfig:sim4-thrusts} and Supplementary Figs.~7d to 7f) allow the robot to achieve balance throughout all flight maneuvers without failures. 
The absolute joint position errors are computed as the absolute value of the difference between the measured and the desired joint values, with the latter defined as the starting robot hovering configuration, which has been chosen to enforce the distance between jets and the robot surfaces and to stabilize the QP solution. The absolute errors are then summed for each sub-assembly for representation reasons, showing that while errors continue to increase until failure in \textsc{Test1} and \textsc{Test3}, they remain bounded during successful trials (\cref{subfig:sim4-joint-pos-err}).
The observed increase in error in the latter case is due to postural adjustments required to counteract aerodynamic effects, yet decreasing during static phases.

\cref{subfig:sim4-aero-forces-sim} and Supplementary Fig.~7e depict the total aerodynamic force components for both successful tests. In Supplementary Fig.~7e, the forces appear smoother and largely aligned with the direction of relative wind velocity. This because the Axisymmetric Model does not consider the nonlinearities resulting from mutual aerodynamic interactions among robot links. Instead, it is formulated as a function of the local angle of attack $\alpha_{\text{link}}$ only. On the other hand, in \cref{subfig:sim4-aero-forces-sim}, the total aerodynamic force from DNN model exhibits a more nonlinear behavior. In Supplementary Fig.~7j, we reported the difference between the centroidal aerodynamic force components reported in \cref{subfig:sim-rob-aero-forces}. These results show how the difference is higher in the transition maneuvers, leading to oscillations in the joint positions in these flight phases.

The asymmetry of thrust forces from the jet engines in \cref{subfig:sim4-thrusts} is attributed to the aerodynamic forces, which generate a non-zero moment acting at the robot CoM, balanced by the thrust to maintain stability. Despite the different aerodynamic models used in simulation, the thrust profiles are comparable.

\subsubsection*{Robustness Analysis}\label{results/sim_exp/robustness}

We tested the robustness of the proposed flight control strategy when the estimated aerodynamic forces used for control feedback did not exactly match the simulated aerodynamic forces. Therefore, we performed a simulation where the system dynamics simulator uses the more realistic DNN model, while the controller aerodynamic feedback is provided by the simpler Axisymmetric Model. 
Results reported in \crefrange{subfig:sim-rob-jointPos-err}{subfig:sim-rob-aero-forces} show that the robot is able to complete the designed flight envelope.

The plot of the robot absolute joint position error $\Delta s$ (\cref{subfig:sim-rob-jointPos-err}) presents oscillations during the transition maneuvers, however, the CoM tracking error and the joint position error remain bounded.

In \cref{subfig:sim-rob-thrusts} we can observe that the thrusts actuated by the robot are asymmetric because of the resulting distributed aerodynamic forces acting on the robot links, coherently with the outcome of the previous simulations.
In \cref{subfig:sim-rob-aero-forces} we compare the global aerodynamic forces acting at the robot CoM computed by the simulator (DNN model) and the control (Axisymmetric model). Although they have similar global characteristics (e.g., they are consistent in the global force component signs), the difference between the two estimates is significant, especially for $t \geq \qty{35}{\second}$. An additional plot of the difference between the two models has been reported in Supplementary Fig.~7j.

This robustness analysis assesses the performances of the aerodynamic-aware controller in case of modeling errors. We believe this is an important preliminary step to understand the applicability of the proposed flight controller in real flight experiments.
Furthermore, in Supplementary Fig.~17 and Supplementary Note~7, a more detailed analysis has been performed via ablation tests on the aerodynamic feedback of the control architecture, showing the minimum required aerodynamic model information for a controlled flight envelope.

\subsubsection*{Ground Experiments}\label{results/sim_exp/experiments}

\begin{figure}[htpb]
    \centering
    \begin{subfigure}[]{0.32\textwidth}
         \centering
         \adjincludegraphics[width=\textwidth,trim={{0.00\width} {0.05\height} {0.0\width} {0.05\height}},clip]{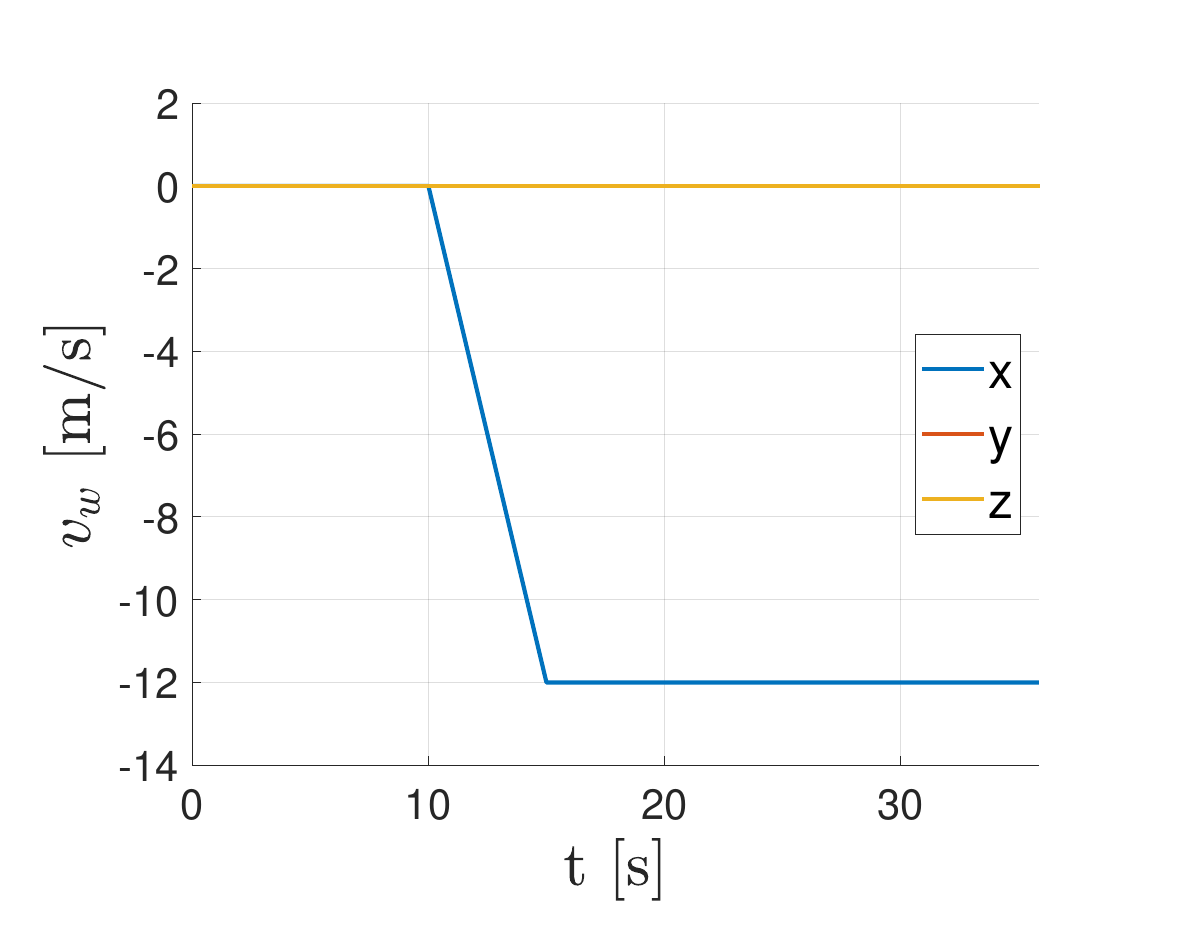}
         \caption{}
         \label{subfig:front-exp-aero-wind}
    \end{subfigure}
    \begin{subfigure}[]{0.32\textwidth}
         \centering
         \adjincludegraphics[width=\textwidth,trim={{0.00\width} {0.05\height} {0.0\width} {0.05\height}},clip]{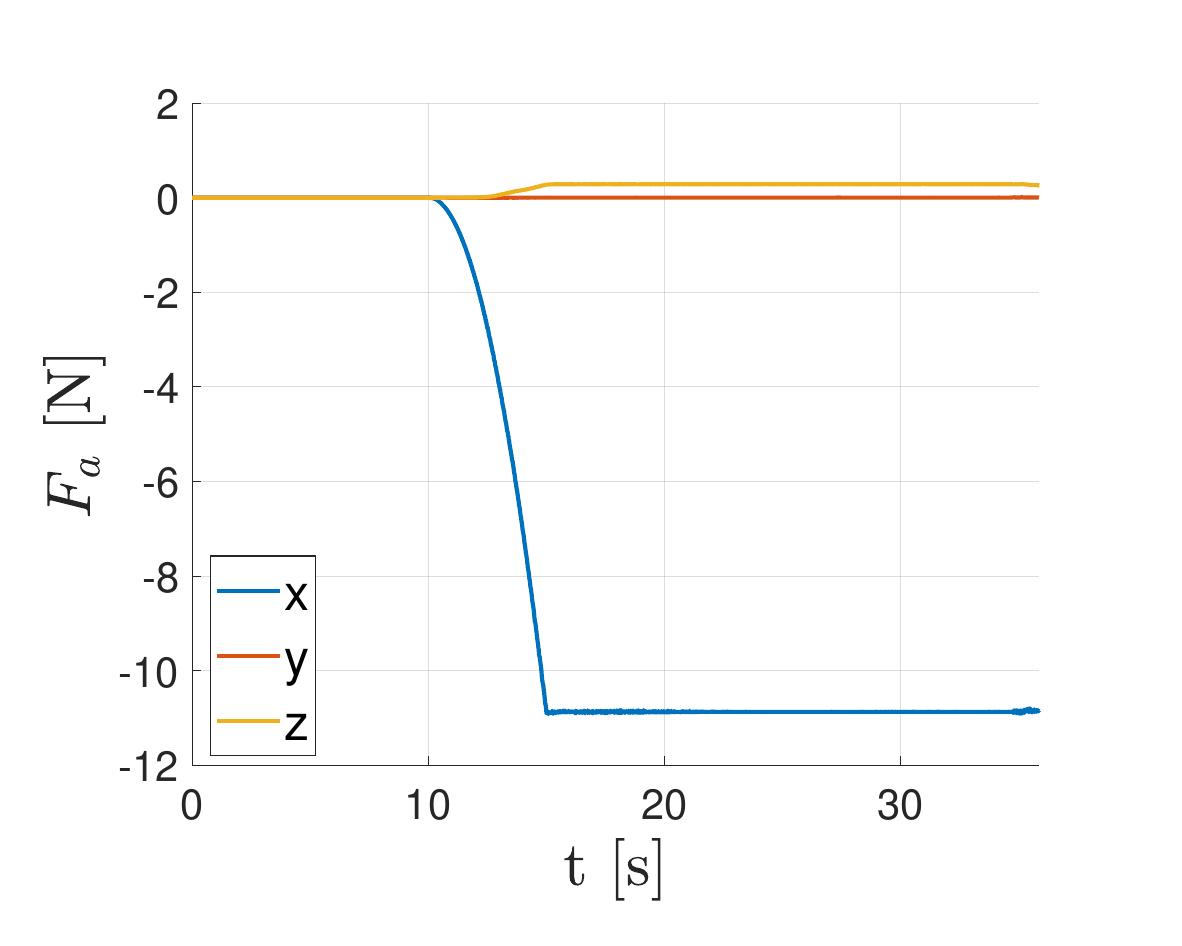}
         \caption{}
         \label{subfig:front-exp-aero-force}
    \end{subfigure}
    \hfill
    \begin{subfigure}[]{0.32\textwidth}
         \centering
         \adjincludegraphics[width=\textwidth,trim={{0.00\width} {0.05\height} {0.0\width} {0.05\height}},clip]{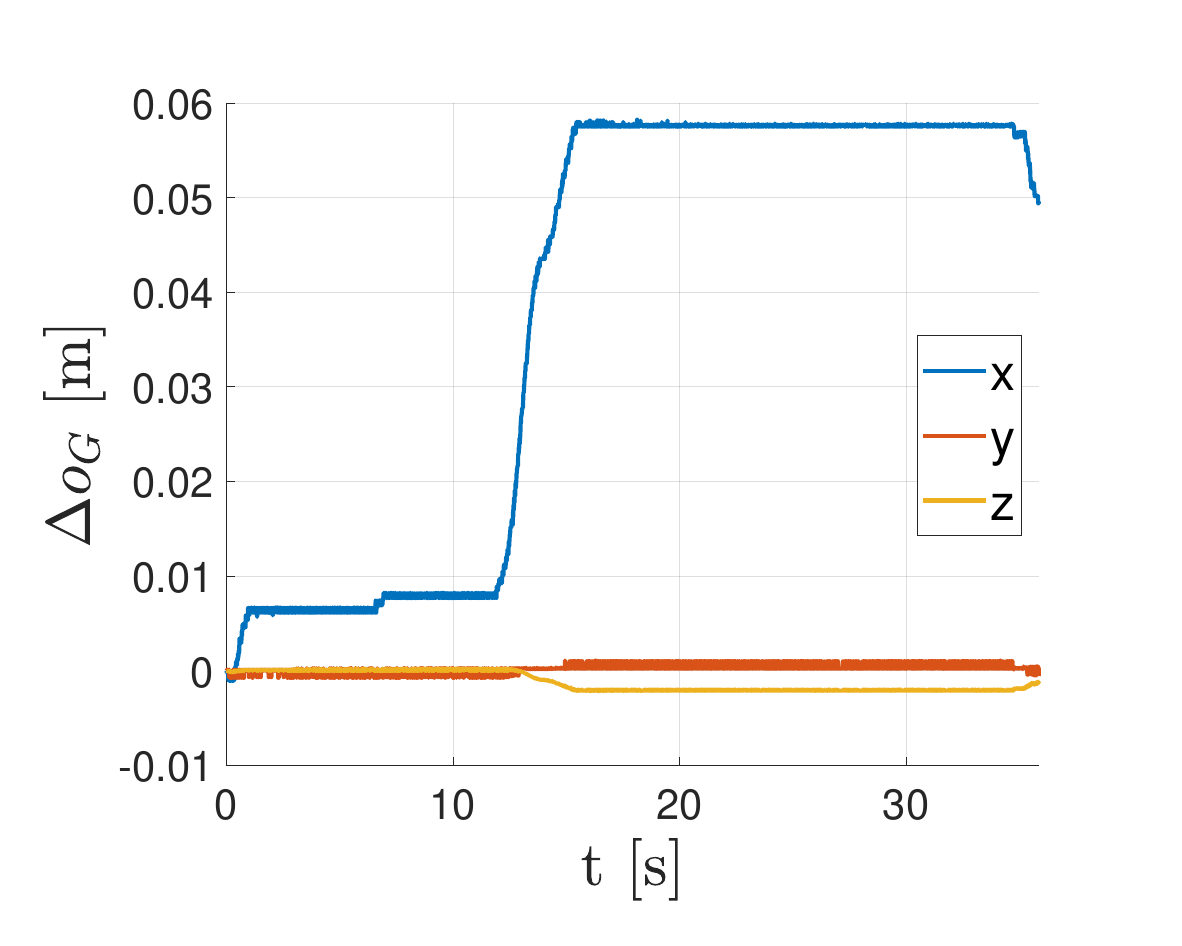}
         \caption{}
         \label{subfig:front-exp-plot-CoM}
    \end{subfigure}
    \\
    \begin{subfigure}[b]{0.32\textwidth}
         \centering
         \adjincludegraphics[width=\textwidth,trim={{0.00\width} {0.05\height} {0.0\width} {0.05\height}},clip]{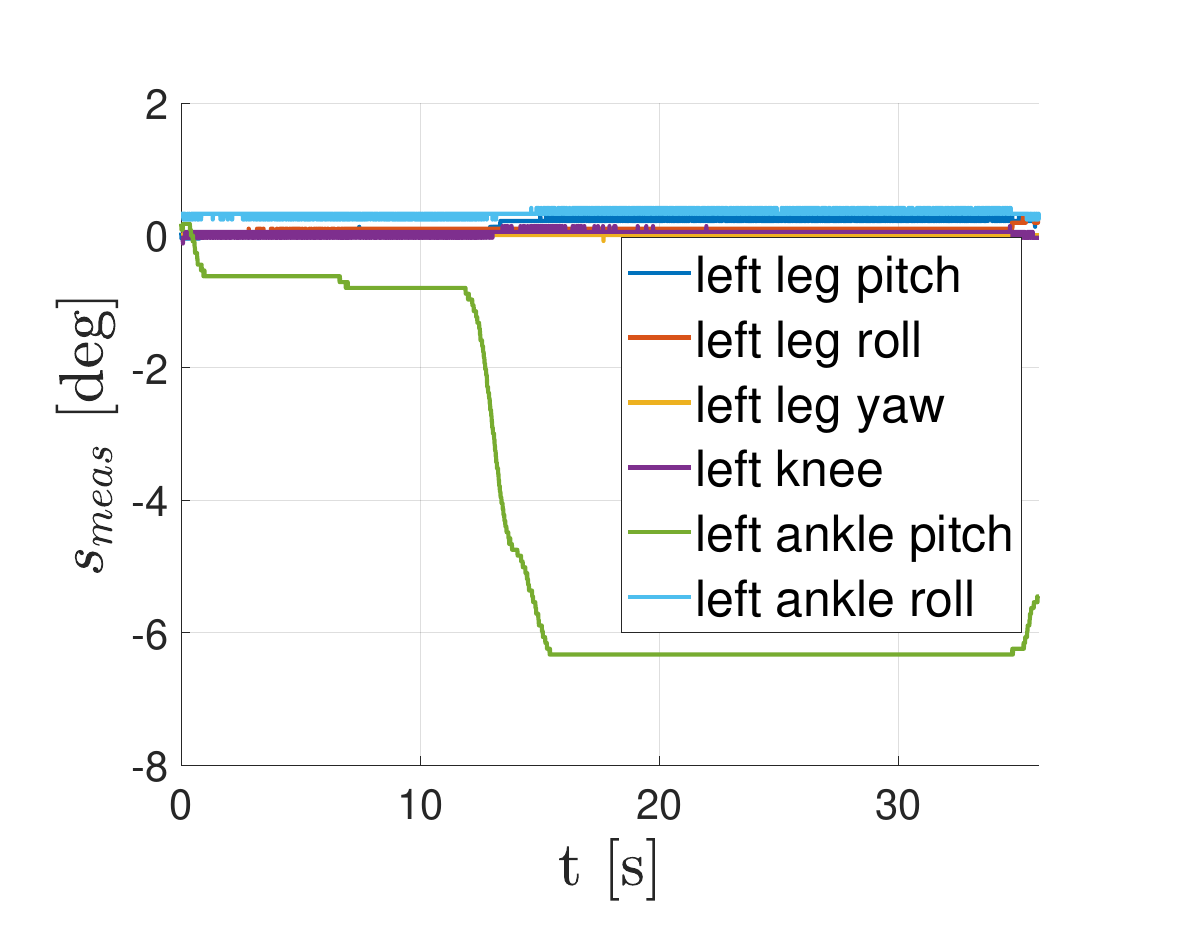}
         \caption{}
         \label{subfig:front-exp-plot-jointPosMeas-lLeg}
    \end{subfigure}
    \hfill
    \begin{subfigure}[b]{0.32\textwidth}
         \centering
         \adjincludegraphics[width=\textwidth,trim={{0.00\width} {0.05\height} {0.0\width} {0.05\height}},clip]{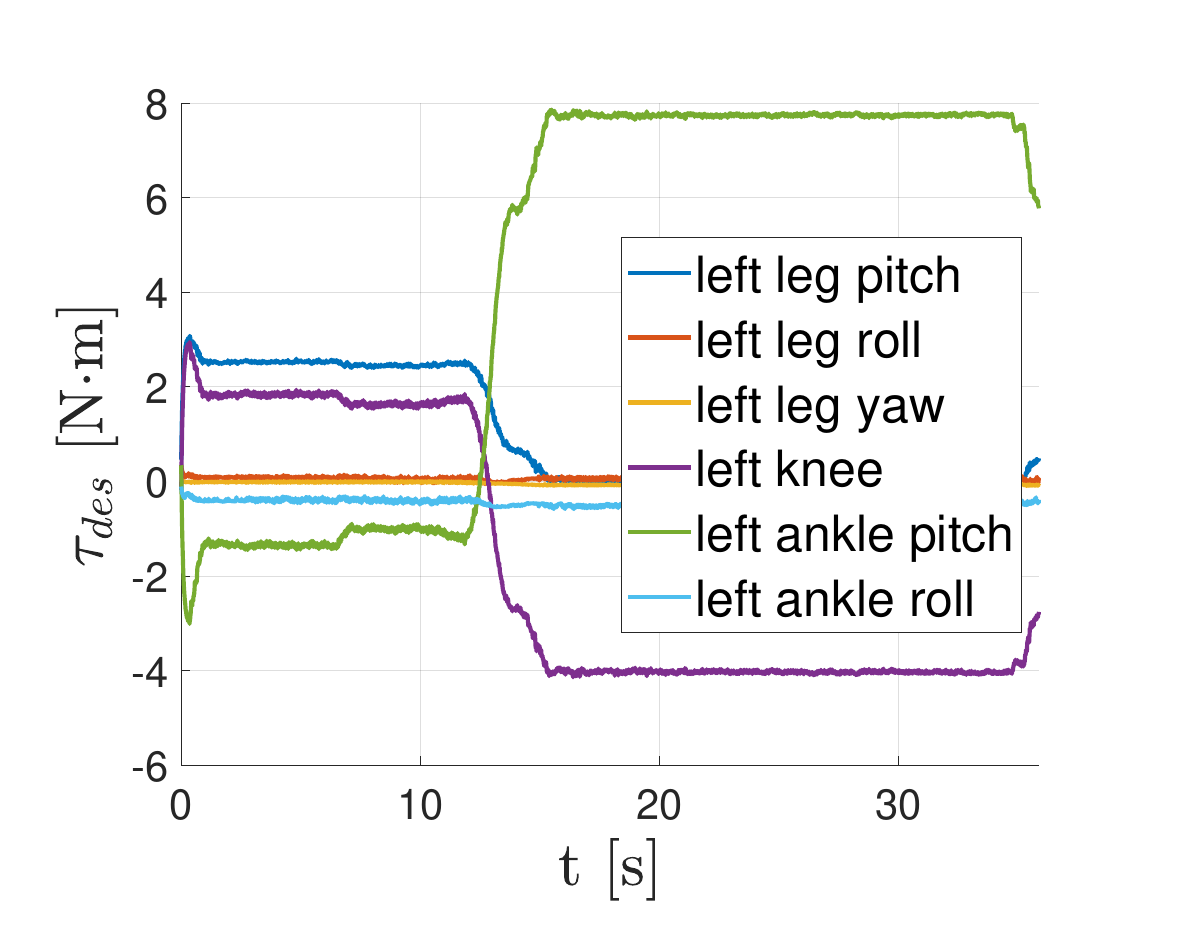}
         \caption{}
         \label{subfig:front-exp-plot-jointTorqueDes-lLeg}
    \end{subfigure}
    \hfill
    \begin{subfigure}[b]{0.32\textwidth}
         \centering
         \adjincludegraphics[width=\textwidth,trim={{0.00\width} {0.05\height} {0.0\width} {0.05\height}},clip]{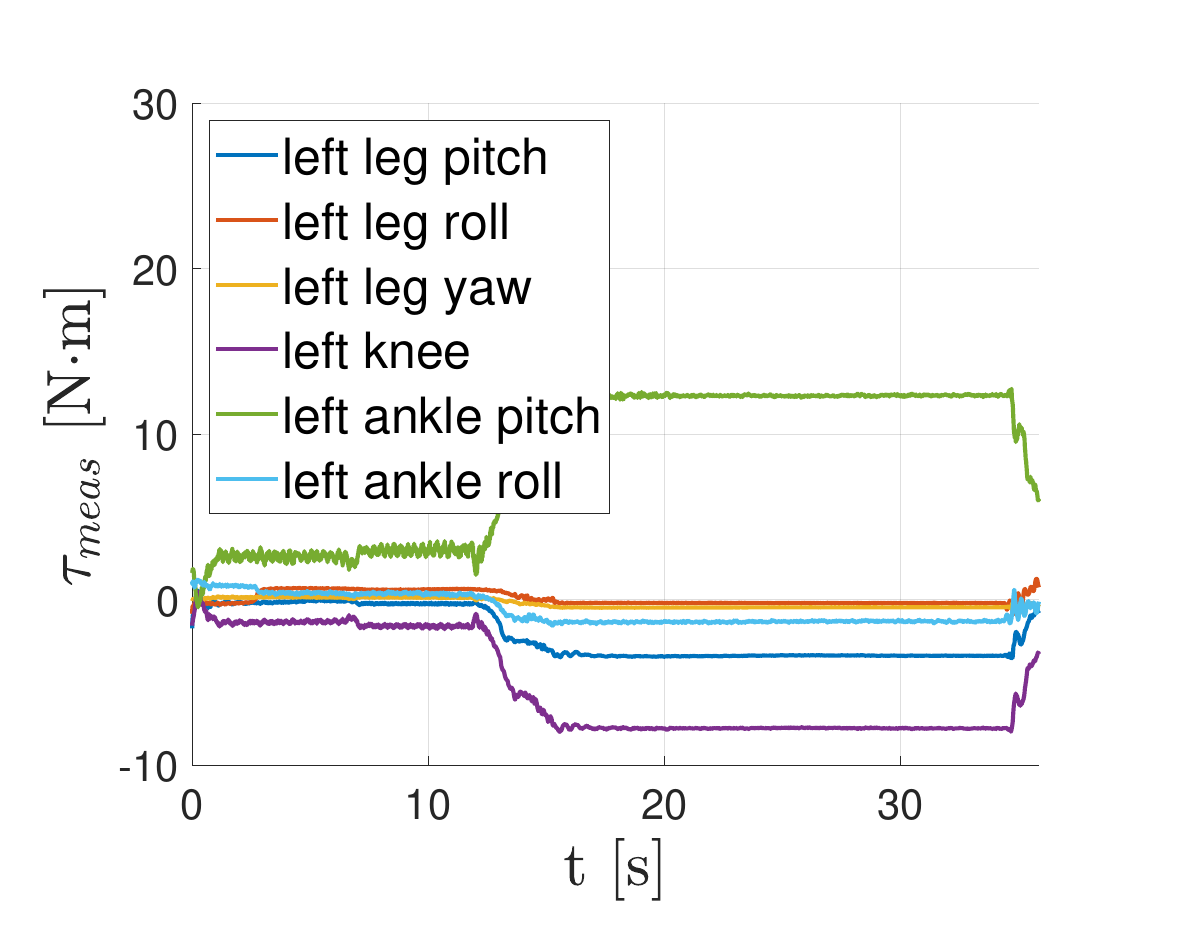}
         \caption{}
         \label{subfig:front-exp-plot-jointTorqueMeas-lLeg}
    \end{subfigure}
    \\
    \begin{subfigure}[]{0.32\textwidth}
         \centering
         \adjincludegraphics[width=\textwidth,trim={{0.00\width} {0.05\height} {0.0\width} {0.05\height}},clip]{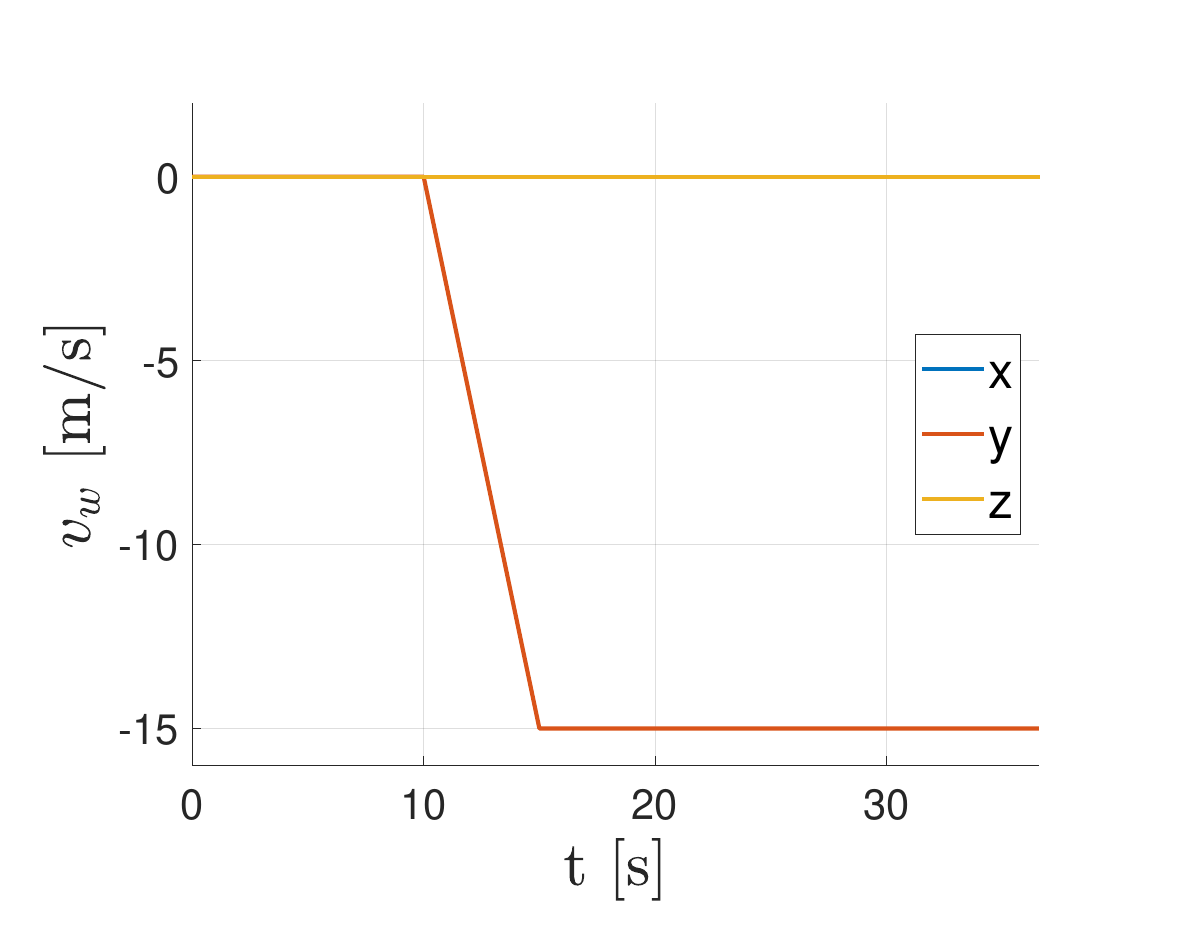}
         \caption{}
         \label{subfig:lateral-exp-aero-wind}
    \end{subfigure}
    \begin{subfigure}[]{0.32\textwidth}
         \centering
         \adjincludegraphics[width=\textwidth,trim={{0.00\width} {0.05\height} {0.0\width} {0.05\height}},clip]{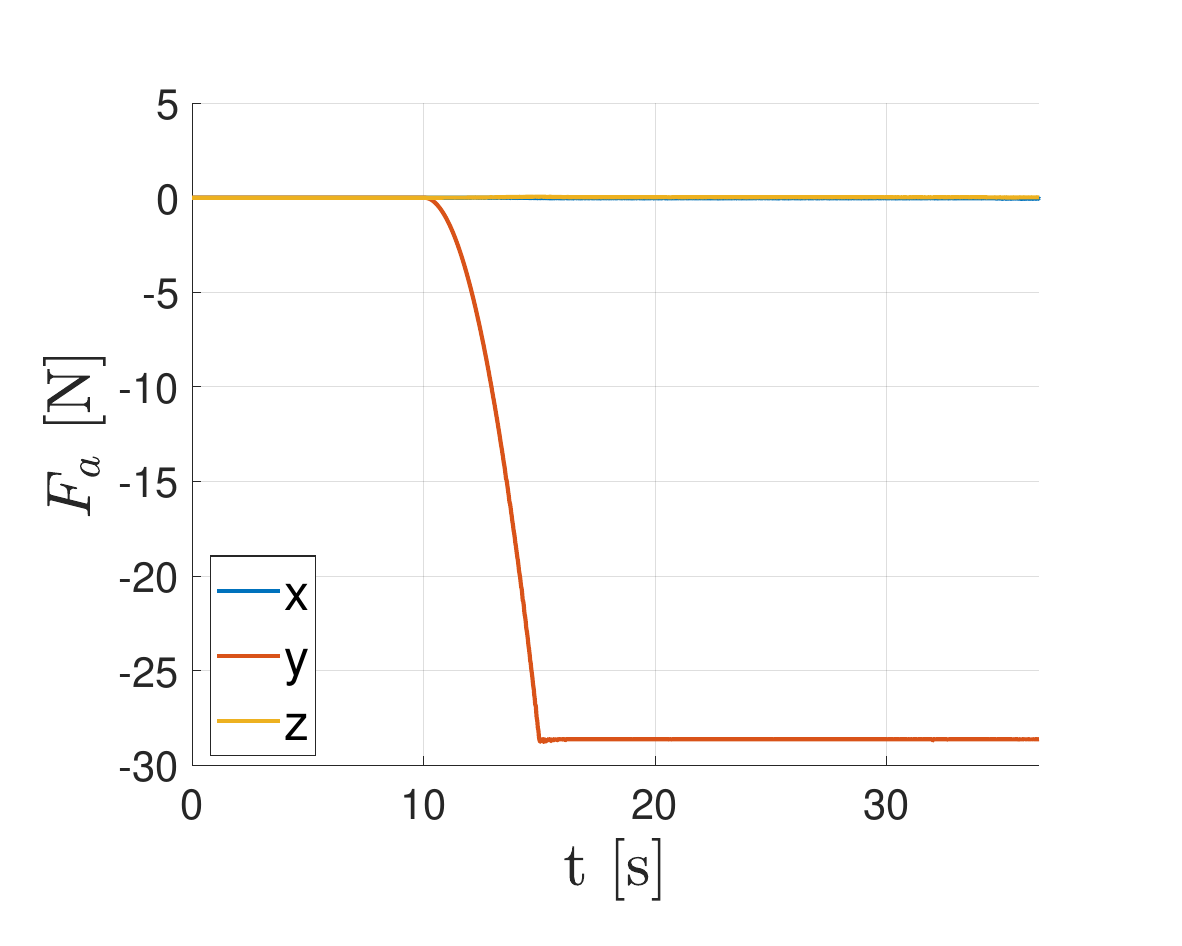}
         \caption{}
         \label{subfig:lateral-exp-aero-force}
    \end{subfigure}
    \hfill
    \begin{subfigure}[]{0.32\textwidth}
         \centering
         \adjincludegraphics[width=\textwidth,trim={{0.00\width} {0.05\height} {0.0\width} {0.05\height}},clip]{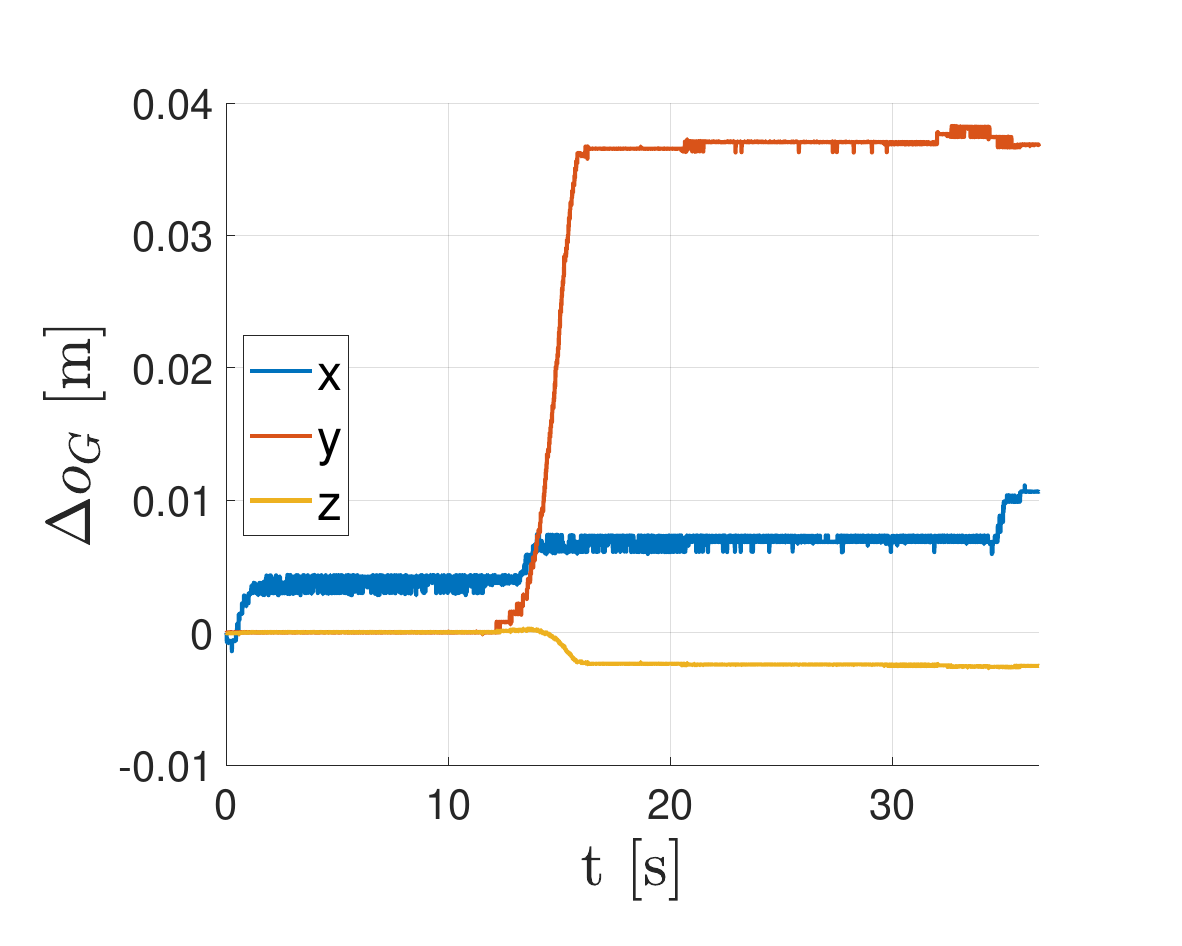}
         \caption{}
         \label{subfig:lateral-exp-plot-CoM}
    \end{subfigure}
    \\
    \begin{subfigure}[b]{0.32\textwidth}
         \centering
         \adjincludegraphics[width=\textwidth,trim={{0.00\width} {0.05\height} {0.0\width} {0.05\height}},clip]{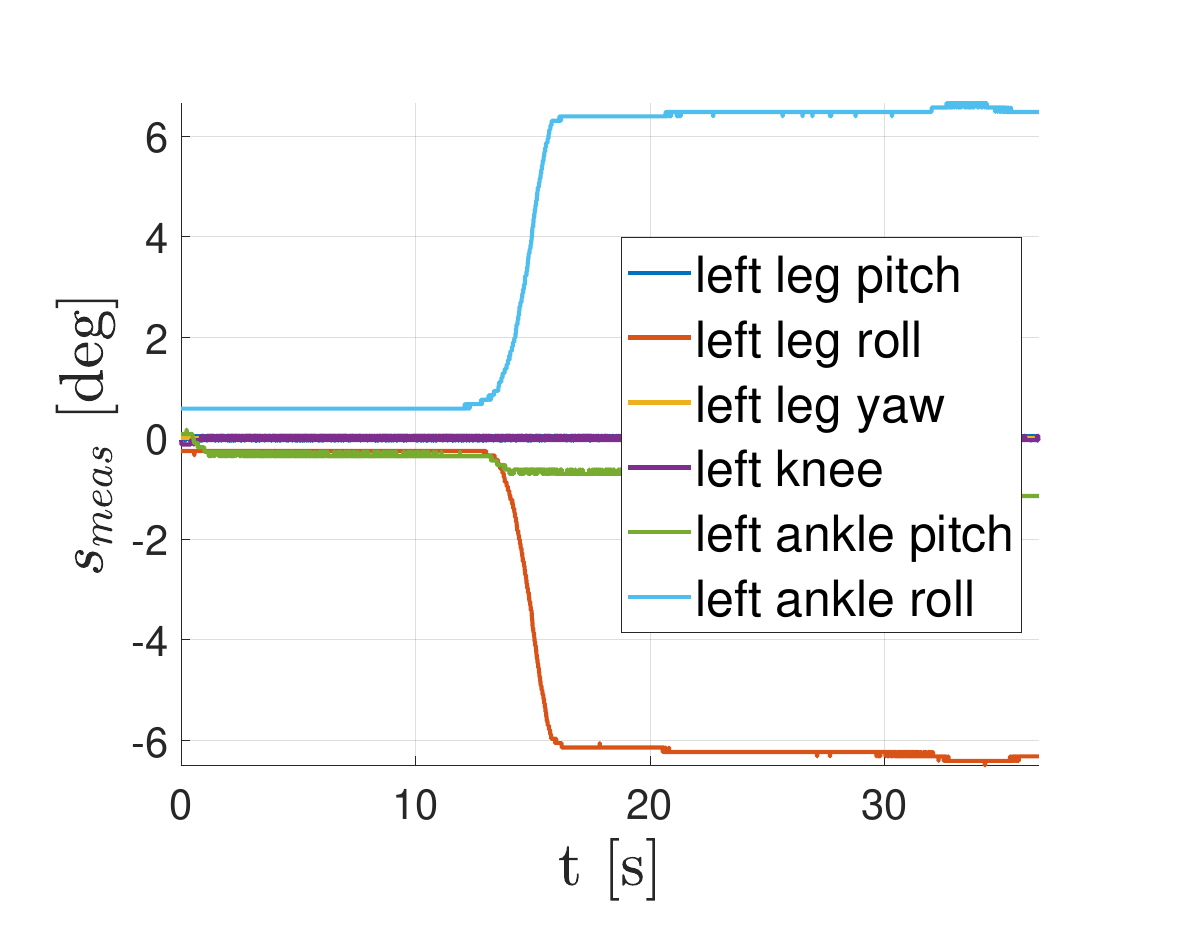}
         \caption{}
         \label{subfig:lateral-exp-plot-jointPosMeas-lLeg}
    \end{subfigure}
    \hfill
    \begin{subfigure}[b]{0.32\textwidth}
         \centering
         \adjincludegraphics[width=\textwidth,trim={{0.00\width} {0.05\height} {0.0\width} {0.05\height}},clip]{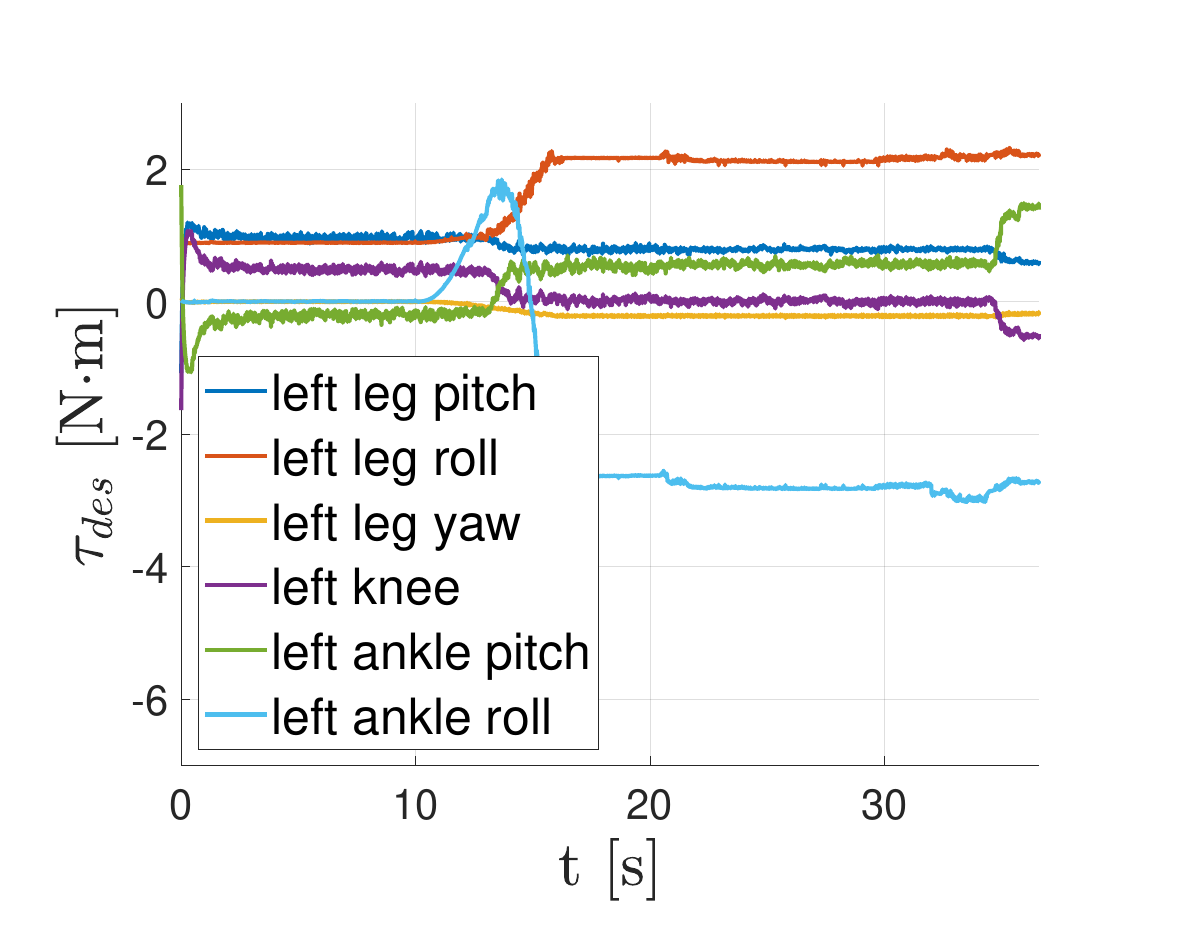}
         \caption{}
         \label{subfig:lateral-exp-plot-jointTorqueDes-lLeg}
    \end{subfigure}
    \hfill
    \begin{subfigure}[b]{0.32\textwidth}
         \centering
         \adjincludegraphics[width=\textwidth,trim={{0.00\width} {0.05\height} {0.0\width} {0.05\height}},clip]{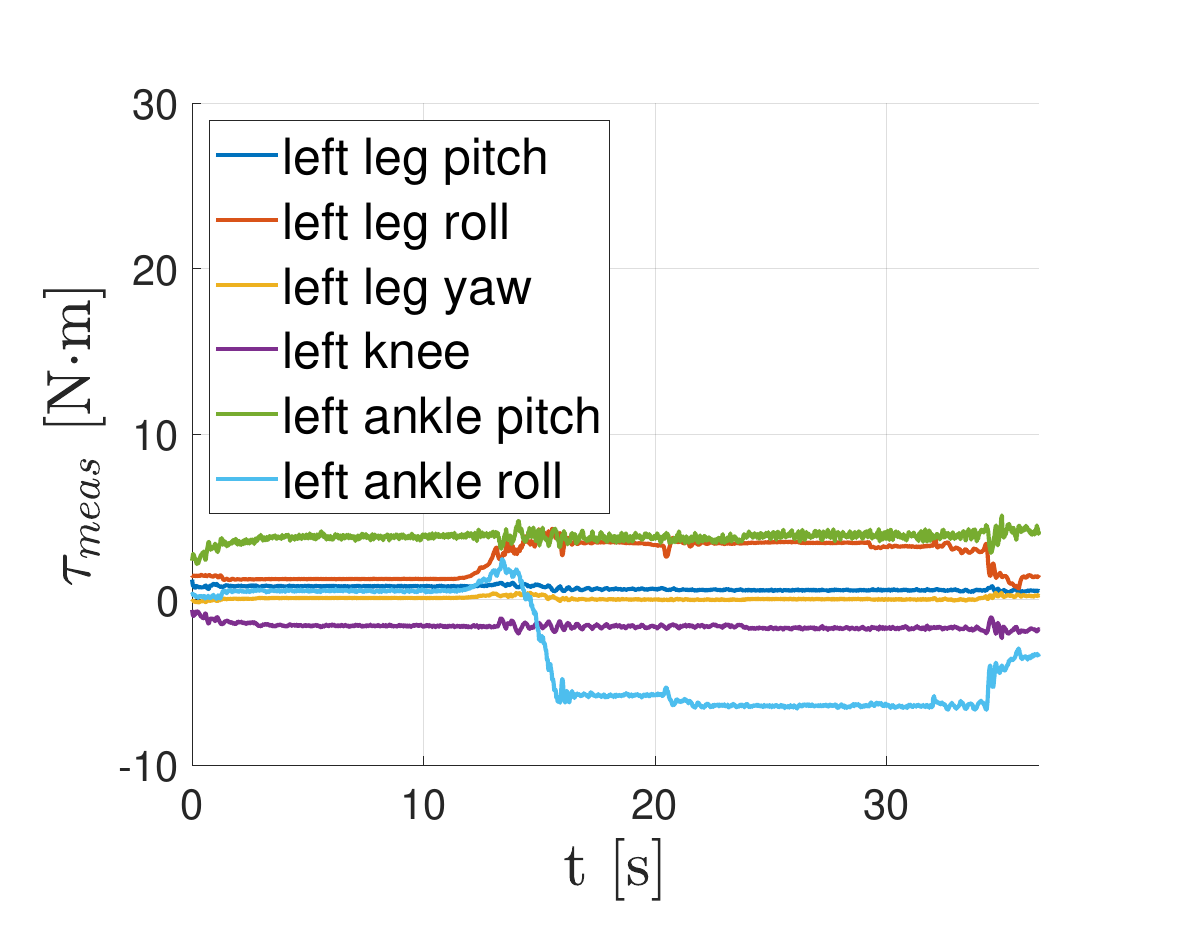}
         \caption{}
         \label{subfig:lateral-exp-plot-jointTorqueMeas-lLeg}
    \end{subfigure}
    \caption{\textbf{iRonCub ground experiments.} Frontal wind ground experiment: (\subref{subfig:front-exp-aero-wind}) wind velocity profile, (\subref{subfig:front-exp-aero-force}) total aerodynamic force on the robot, (\subref{subfig:front-exp-plot-CoM}) robot CoM error, (\subref{subfig:front-exp-plot-jointPosMeas-lLeg}) measured left leg joint positions, (\subref{subfig:front-exp-plot-jointTorqueDes-lLeg},\subref{subfig:front-exp-plot-jointTorqueMeas-lLeg}) desired and measured left leg joint torques. Lateral wind ground experiment: (\subref{subfig:lateral-exp-aero-wind}) wind velocity profile, (\subref{subfig:lateral-exp-aero-force}) total aerodynamic force on the robot, (\subref{subfig:lateral-exp-plot-CoM}) robot CoM error, (\subref{subfig:lateral-exp-plot-jointPosMeas-lLeg}) measured left leg joint positions, (\subref{subfig:lateral-exp-plot-jointTorqueDes-lLeg},\subref{subfig:lateral-exp-plot-jointTorqueMeas-lLeg}) desired and measured left leg joint torques.}
    \label{fig:6}
\end{figure}

Currently, we lack the capability to systematically test the aerodynamic-aware controller through real flight experiments. However, we conducted tests on the real robot, focusing on balancing trials with jet engines turned off, during which we simulated an artificial fictitious wind profile directly injected into the controller as feedback, but not physically acting on the robot. These tests aim to verify the robustness and reliability of the aerodynamic-aware controller on the real robot. Specifically, we evaluate whether the robot can maintain balance, and understand if the addition of aerodynamics enhances control without disrupting the existing implementation.
We conducted two different experiments, both starting from the same initial position of the robot. Each experiment involved applying artificial relative wind profiles to the robot, with different magnitudes and directions of relative wind velocity.
In the first experiment, we imposed a frontal wind velocity profile, as shown in \cref{subfig:front-exp-aero-wind}. This wind velocity pattern resulted in a consistent estimation of the total aerodynamic force using the Axisymmetric Model, depicted in \cref{subfig:front-exp-aero-force}. This estimation was derived by summing all forces acting on each individual link.
In the second experiment, we replicated the procedure using a lateral wind gust, as shown in \cref{subfig:lateral-exp-aero-wind}. This wind profile led to the total aerodynamic force profile presented in \cref{subfig:lateral-exp-aero-force}, employing the same aerodynamic model as in the first experiment.
In both experiments, the observed outcome is that the robot adjusts its motion in the opposite direction of the aerodynamic force to counterbalance it and minimize the estimated linear momentum derivative. This adjustment is shown in \cref{subfig:front-exp-plot-CoM,subfig:lateral-exp-plot-CoM}, which depicts how the CoM position adapts to offset the aerodynamic force.
In the first experiment involving frontal wind, the modification in CoM position primarily occurs through variations in ankle pitch joints, as can be seen in \cref{subfig:front-exp-plot-jointPosMeas-lLeg}. These joints are actuated (\cref{subfig:front-exp-plot-jointTorqueMeas-lLeg}) in accordance with the desired joint torques optimized by the controller (\cref{subfig:front-exp-plot-jointTorqueDes-lLeg}).
Similarly, in the second experiment, adjustments are observed in the leg hip and ankle roll joints (\cref{subfig:lateral-exp-plot-jointPosMeas-lLeg}). Once again, these joints are actuated (\cref{subfig:lateral-exp-plot-jointTorqueMeas-lLeg}) in response to the desired joint torques optimized by the controller (\cref{subfig:lateral-exp-plot-jointTorqueDes-lLeg}).

\section{Discussion}\label{discussion}

We proposed a comprehensive approach to address the challenge of modeling and controlling aerodynamic forces during the flight of a humanoid robot. This methodology encompasses simulating and validating aerodynamic forces, followed by learning aerodynamic models for integration into the control architecture. Our results show the effectiveness of the newly proposed control strategy in enhancing robot performance within a specified flight envelope.

The current limitation is the inability to test the proposed algorithm in real flight scenarios. Currently, the required sensors for accurately estimating wind direction are unavailable on our robot. Additionally, setting up repeatable and reliable experiments in open-air environments poses challenges.
On the other hand, given the high-temperature flow exiting from the jet turbines and the toxic gases produced by the combustion of the Jet A1 fuel, performing indoor experiments with such a robotic platform would require dedicated facilities that we currently do not have access to. As a consequence, we are currently exploring the possibility of equipping ourselves with appropriate indoor facilities for the experiments.
The control of the robot also presents difficulties. iCub's arm actuation is cable-driven, introducing undesired elasticity and delays, resulting in slow and imprecise joint motion.
However, despite these limitations, we are actively conducting experiments with the real robot to enhance both the setup and the hardware, aiming to overcome these obstacles.

Regarding the methodology presented, we assumed the flow to be incompressible despite the nearly unitary Mach number flow exiting the jet engine nozzles. We considered this assumption acceptable for two main reasons: 
i) conducting experiments in a wind tunnel with active jet engines was not feasible, hence the data were effectively collected under incompressible flow conditions;
ii) during flight, it is necessary to keep the jets as far away from the robot surfaces as possible to prevent damage. This precaution allows us to assume that the incompressibility of the flow near the robot's surface is an acceptable assumption since direct interaction effects between the jet and the flow generated around the links are minimized. Furthermore, in agreement with the analysis detailed in Supplementary Note~6, we are investigating design modifications to keep the jets far from the robot, as presented in \cite{vanteddu2024cad}.

To justify the steady flow assumption, we refer to the cited literature works indicating that employing steady turbulence models to solve highly turbulent flows around bluff bodies effectively captures the correct average aerodynamic forces acting on the bodies, even if it neglects physical fluctuations that occur at small temporal and spatial scales resulting from turbulence effects \cite{defraeye2010aerodynamic, blocken2018aerodynamic, blocken2021impact,  defraeye2010computational}. 
The steady CFD simulations utilized for collecting the aerodynamic dataset introduce another significant assumption, which neglects the unsteady effects produced by the robot's movements within the designated flight envelope, essentially implying that joint velocities and robot accelerations remain sufficiently low, allowing variations in wind velocity to be considered quasi-steady.
This approximation extends to the identified aerodynamic models, as the Deep Neural Network (DNN) is trained with only the robot's joint angles and relative wind velocity.

Regarding control design, the proposed strategy is a direct extension of the existing flight controller, employing full compensation of aerodynamic forces by treating them as disturbances. While this approach is effective, alternative control strategies could be implemented to exploit aerodynamic forces for enhancing flight performance.

In our current experiments, we are focusing on developing a new robot design featuring more precise arm actuation. Additionally, we are exploring methods to integrate onboard sensors for accurately estimating wind intensity and velocity. These enhancements aim to improve the overall performance and reliability of the robot.

To address the limitations posed by the use of steady Reynolds-Averaged Navier-Stokes (RANS) equations with turbulence modeling, we plan to further explore turbulence resolving simulation techniques, such as Large Eddy Simulations (LES), offering a more detailed representation of turbulent flow around the robot, potentially leading to more accurate flow characteristics predictions. However, LES comes with a high computational cost, especially for high-$\reynolds$ flows. 
Furthermore, we are exploring the use of GPU-accelerated Detached Eddy Simulations (DES) that also include compressibility effects to address the jet modeling challenges.

Continuing with the advancements in aerodynamic models, we could use DNNs not only to map quasi-steady quantities but also the aerodynamic effects due to the robot motion. This approach could enhance the accuracy and versatility of aerodynamic force estimation by incorporating additional dynamic factors into the modeling process.
Furthermore, the current aerodynamic force estimation lacks direct measurement of flow quantities and relies solely on relative wind velocity direction. Future research directions could explore integrating distributed pressure sensors onto the robot covers. These sensors could feed data into a Physics-Informed Neural Network architecture, that uses both simulated and real-time pressure data to predict the pressure distribution on the robot and consequently estimate aerodynamic forces at future states of the robot, improving the reliability of aerodynamic force predictions.

Advanced flight control strategies could exploit the aerodynamic forces to stabilize the robot during high speed flight. When combined with a dedicated trajectory planner, this approach could facilitate the execution of aggressive flight maneuvers. Moreover, the results obtained from the methodology proposed in this paper will be useful to understand future designs, i.e. the new iRonCub design can prioritize a more aerodynamic shape, fundamental for effectively exploiting aerodynamic forces for control purposes. This \textit{aerodynamic shaping} can be achieved through two main strategies: 
i) designing wings to exploit aerodynamic lift; 
ii) developing morphing covers that adapt to the external aerodynamic flow, thereby reducing aerodynamic drag \cite{bergonti2022morphingcovers}. 
By incorporating these features into the design of the next version of iRonCub, we can enhance its aerodynamic performance and enable more sophisticated flight control capabilities, improving efficiency and fuel consumption, and extending the flight endurance of the robot, which would be a crucial design requirement for effective real-world deployment in search and rescue missions.

\section{Methods}\label{methods}

In this section, we present the methods employed to produce the results achieved in this paper, following the same outline of the Results section: 
\begin{itemize}
    \item[] \textit{iRonCub-Mk1 Mechanical Design}: we show the design methods and strategies adopted to build the hardware components of the iRonCub-Mk1 prototype.
    \item[] \textit{PoliMi Wind Tunnel Facility}: we introduce the main characteristics of GVPM wind tunnel, then we describe the adaptations of wind tunnel setup for experiments on a real humanoid robot, including the design of the hardware interfaces to support \textit{iRonCub-Mk1} robot and measure the aerodynamic forces and pressures acting on it.
    \item[] \textit{Computational Fluid Dynamics Models}: we detail the equations and models employed for the computational aerodynamic study of flying humanoid robots in wind tunnel setup and free airstream flow.
    \item[] \textit{Aerodynamic Models for Learning and Online Control}: we report our mesh sensitivity analysis for large aerodynamic forces dataset generation, subsequently, we present the \textit{Linear Regression} and \textit{Deep Neural Network} algorithms used for whole-body aerodynamic modeling based on CFD simulations.
    \item[] \textit{Aerodynamic Control Design and Simulation Environment}: we showcase the design of the whole-body momentum-based aerodynamic controller and detail the aerodynamic simulator implemented for the results' validation.
\end{itemize}

\subsection{iRonCub-Mk1 Mechanical Design}

\subsubsection*{Actuators Selection}

The iRonCub robot's design process started with the choice of the actuation system.
Most flying robots adopt electric propellers because they offer a faster response and lower engineering complexity, but they need high power and, consequently, large battery capacity. For this reason, after a brief analysis, we decided to employ jet engines \cite{lerario2020modeling}.
In particular, if we consider the following propulsion systems:
\begin{itemize}
\item Dynamax CAT 8 FAN: an electric motor with a maximum thrust of \qty{\sim15}{\kilo\gram} at \qty{\sim13}{\kilo\watt} consumption (\url{https://edfdynamax.com/\%22cat-8\%22-p\%26p-35lbs-thrust});
\item JetCat P220-RXi: a model jet engine with a maximum thrust of \qty{\sim22}{\kilo\gram} at \qty{\sim0.6}{\litre/\minute} consumption.
\end{itemize}
Assuming that energy and fuel consumption are linear versus the generated thrust, the engine weights are negligible \cite{lerario2020modeling}. We further assume that the energy density of the LiPo batteries is \qty{210}{\watt\hour/\kilo\gram} \cite{ulvestad2018brief,abdilla2015power} and the density of jet fuel (i.e. Jet A1) is similar to the one of diesel \qty{0.8}{\kilo\gram/\litre} \cite{marketing2007diesel}.
If we consider a \qty{40}{\kilo\gram} robot performing five minute flight we would need \qty{26.31}{\kilo\gram} of battery pack versus \qty{4.61}{\kilo\gram} of fuel. The results of this preliminary analysis motivated us to choose fuel-powered jet engines as actuators on both the jetpack and the arms of the robot for our use case.

\subsubsection*{Number and Position of Jet Engines}

After selecting the actuation strategy, we need to understand the number of engines and their placements. In particular, considering a \qty{40}{\kilo\gram} robot, we would need at least two JetCat P220-RXi to lift the robot. Due to space limitations, these engines are placed on the robot back, on the left and right of the robot's battery backpack.
Positioning the jet engines only on robot's back, however, produces a moment around the robot CoM that impairs the control of the robot attitude. For attitude stabilization and control purposes, it was necessary to add other engines on the front part of the robot.
We decided to integrate a P100-RX JetCat in each forearm. We choose the forearms so that the jet nozzles remain at an adequate distance from the ground during takeoff maneuvers, while also being able to perform thrust vectoring by moving the robot's arms joints. Additionally, we made the jet interfaces design of the forearms modular to allow switching between the P100-RX turbines and the similar, but more powerful, P160-RXi-B turbines.
See Supplementary Table~1 for the specifications of the selected jet engines.

\subsubsection*{Design of Jet Engines Interfaces}

The baseline robot from which we started our design is the humanoid robot iCub version 2.7. As a design principle, we decided to create interfaces that minimally modify the line of the robot.
For the back engines, we have decided to create a jetpack (\cref{subfig:jetpack_connection_sectional}) as it can be easily assembled and removed from the robot. Integration of the jetpack on the robot required to design a stainless steel spine, which has been assembled permanently on the robot back to enforce the mechanical structure of the robot in view of the external forces applied by the jet engines. Furthermore, this modification provided us with three easily accessible connection points, two at the top and one at the bottom attached to the spine as seen in \cref{subfig:jetpack_connection_sectional_super_zoom}. The design of the steel spine shape was fundamental to equally distribute the jet engine reaction forces on the jetpack rather than directly on the screws that assemble the mechanism. 

In this iRonCub version, we decided to directly substitute the robot hands and forearm with a new interface including jet engines. The choice of removing the robot arms was due to the impossibility of protecting them from the heat generated by the jet engines. Future work is ongoing to understand how to integrate back safely the hands of the robot.
Since the objective of this prototype is to achieve stable flight, the need for hands was not crucial and removing them simplified the design of the jet interfaces. Also in this case, safety constraints required us to enforce the elbow joint with additional components that can withstand the thrust generated by the engine. The material of the link connecting the jet engine and the elbow is titanium, which has good heat insulation properties and structural strength comparable with that of steel.
Jet engines are assembled with the help of custom-designed brackets.
The forearm assembly can be seen in \cref{subfig:forearm_assembled,subfig:forearm_exploded}.

\subsubsection*{FEM Analysis}

Both the jetpack and the forearm assemblies have been tested with Finite Element Method (FEM) analysis in Ansys Mechanical software to verify the structural integrity and the safety factor. The assemblies have been subjected to fixed boundary conditions at the connection points and boundary loads in the direction and magnitude of the jet reaction forces. The result of the analysis is shown in Supplementary Fig~1. The maximum stresses developed in the jetpack and the forearm are respectively \qty{57}{\mega\pascal} and \qty{77}{\mega\pascal}. The stresses developed are always at least $5$ times lower than the Ultimate Tension Strength of the material with least strength (Ergal in this context).

\subsection{PoliMi Wind Tunnel Facility (GVPM)}\label{methods/GVPM}

The experimental activity was performed in the large wind tunnel of Politecnico di Milano (GVPM). The GVPM is a closed-circuit wind tunnel, arranged in a vertical layout with two test sections located on the opposite sides of the loop, as shown in \cref{subfig:gvpm_layout}. The test section in the upper side of the loop is \qty{13.84}{\meter} wide and \qty{3,84}{\meter} high, widely used for wind tunnel campaigns requiring the reproduction of the atmospheric boundary layer and characterized by a maximum wind speed of \qty{16}{\meter/\second}. The tests with the iRonCub humanoid robot were performed in the $\qty{4}{\meter} \times \qty{3.84}{\meter}$ low-turbulence test section \cite{gvpm} positioned in the lower side of the loop, as can be observed from the particular in \cref{subfig:gvpm_test} and characterized by a maximum wind speed of \qty{55}{\meter/\second} and a turbulence intensity lower than \qty{0.1}{\percent}. 

The dimensions of the low-turbulence test section of GVPM permitted planning and performing a full-scale experiment with the real iRonCub-Mk1 robot. Specifically, it was possible to perform real-scale experiments with a low blockage effect of the robot on the wind tunnel flow (\emph{blockage ratio:} $BR \le \qty{3}{\percent}$) comparable to similar experiments \cite{defraeye2010aerodynamic}.

Moreover, it was possible to perform tests varying both the wind speed and attitude of the robot, thus reproducing possible real flight conditions for the robot. For this purpose, a metallic remotely driven supporting pylon and a turntable embedded in the test section floor were employed to automatically change the pitch and yaw angles of the robot.

The GVPM is also equipped with state-of-the-art instrumentation of the main experimental fluid dynamics techniques, from force and pressure measurements to optical systems. The flow and model settings as well as the instrumentation and data acquisition are managed through a Control Room outside the wind tunnel plenum containing the circuit using Labview software.

\subsubsection*{Experiment setup}\label{methods/experiment_setup}

Each test was performed under a constant airspeed and a specific robot joint configuration. The angles of attack and side-slip were manipulated by adjusting the motorized wind tunnel pole: the angle of attack $\alpha$ is the angle between the robot vertical axis and the horizontal plane, and the sideslip angle is the angle between the longitudinal symmetry planes of the robot and the wind tunnel. Tests were categorized into two types: $i)$ \textit{hovering tests}, where the angle of attack remained fixed ($\alpha = \qty{90}{\degree}$) while the side-slip angle varied ($\beta\in [\qty{-90}{\degree},\qty{90}{\degree}]$), and $ii)$ \textit{flight tests}, where the side-slip angle was constant ($\beta = \qty{0}{\degree}$) while the angle of attack varied ($\alpha \in [\qty{25}{\degree},\qty{65}{\degree}]$). The \textit{hovering tests} refer to tests in which the robot is in the optimal hovering configuration but the wind tunnel airspeed is non-null to reproduce the conditions of a constant intensity wind gust. The chosen range for these angles was determined by the setup constraints, including the robot, its support, and the movable pole. Despite the pole's ability to rotate fully on the yaw axis, the robot would be affected by the scale windshield wake for half of the total range. Additionally, the pole's pitching movement was limited to \qty{40}{\degree}, thereby constraining the yaw and pitch angle range for testing.

Both force and pressure measurements were collected during the wind tunnel experimental campaign. A six-component strain gauge RUAG 192-6L installed on the head of the GVPM supporting pylon was used for aerodynamic force measurements. The calibration report delivered by the balance manufacturer declared a maximum error below 0.2\% of the design loads for the load class corresponding to the load conditions measured in the present test campaign.
The force and pressure measurements have been averaged over \qty{10}{\second} time intervals to extract their mean values.
An aluminum short strut connected the metric surface of the balance to the robot. The balance was shielded by means of a nylon fairing manufactured by rapid prototyping technique. Both the aluminum struct and the nylon fairing have been custom-designed to allow this specific experimental activity on humanoid robots. The iRonCub robot set up in the GVPM wind tunnel is shown in \cref{subfig:gvpm_robot}.

Pressures over the robot skin were acquired by means of four low-range 32-port scanners by Pressure System Inc. embedded inside the jet pack of the robot connected by piping to a total number of 124 pressure taps. The positions of the pressure taps over the robot skins have been selected to collect meaningful data for all the different tested wind directions (see Supplementary Fig.~2), the nylon covers designed to accommodate the pressure taps have also been manufactured by rapid prototyping technique.

\subsection{Computational Fluid Dynamics Models}\label{methods/CFD_models}

We started by analyzing the typical airflow properties of iRonCub-Mk1 during a plausible mission envelope to understand the correct equations to be used in CFD simulations. In particular, the negligible Mach number ($\mach$) computed considering the designed flight speed and altitude of the robot led us to consider the flow over the robot surface as incompressible (see Supplementary Note~1). 
To further simplify the calculations, we decided to neglect unsteady turbulence effects and adopt the steady \textit{Reynolds-Averaged Navier-Stokes} equations \cite{Reynolds1895RANS} to address the turbulence solution problem using ensemble averages (see Supplementary Note~3).

After selecting the system of equations to be solved, we defined the numerical processes to be used for CFD simulations, more specifically geometry modeling and the fluid volume modeling processes. Our goal when selecting the modeling methods is to keep high results accuracy within the constraints represented by the computational cost and calculation feasibility.

In the geometry modeling process, we simplified non-regular surfaces and small gaps found in the CAD geometries of both the robot and the wind tunnel support. This operation is necessary to prevent Ansys Fluent Meshing workflow from failing due to sharp edges, small gaps, and cavities between different components. We employed two consecutive simplification procedures using PTC Creo and Ansys SpaceClaim CAD software. The simplified models of both the robot and the wind tunnel support preserved their main geometric features after the simplification processes (see Supplementary Fig.~3).

Concerning the fluid volume modeling, we generated meshes of the discretized fluid volume to enable the usage of the Finite Volume Method to solve the flow field around the robot in the wind tunnel \cite{eymard2000finite}. We defined a square prism control volume of $\qty{4}{\meter} \times \qty{4}{\meter} \times \qty{6}{\meter}$ surrounding the robot model assembled on the wind tunnel support (see Supplementary Fig.~3). Subsequently, we generated a good quality mesh for all the simulated configurations using the Watertight workflow of Ansys Fluent Meshing by setting the mesh sizing parameters reported for \texttt{953C} msh in Supplementary Table~3. We also added a refinement hex-core control volume surrounding the robot and a boundary layer refinement (developing in the normal direction from the robot surface) made of 5 cell layers with an average normal first cell height of \qty{1.74e-4}{\meter} and an average tangential cell length of \qty{2.24e-3}{\meter} to limit the number of boundary elements solution solved using (approximated) wall functions. The final mesh on the robot and support models is displayed in Supplementary Fig.~4.

Once concluded the meshes generation, we solved the RANS equations using Ansys Fluent solver with the following boundary conditions: \textit{no-slip wall}, for having zero velocity values on the surfaces of the robot and the support; \textit{symmetry plane} condition for the wind tunnel lateral walls (to avoid solving the boundary layer also where not needed, i.e. far from the robot surfaces); constant \textit{velocity-inlet} and \textit{pressure-outlet} conditions, specifying the flow velocity at the entrance of the fluid control domain and the pressure at its exit. Given the low turbulence inlet levels of the wind tunnel chamber, we set the \textit{Turbulence kinetic energy} to $k=\qty{2.89e-6}{m^2/s^2}$, the \textit{Turbulence dissipation rate} $\varepsilon=\qty{0.0236e-6}{m^2/s^3}$ and the \textit{Specific turbulence dissipation rate} $\omega=\qty{8.17}{s^{-1}}$ at the domain external boundaries following \cite{constantinescu2003turbulence}. Despite the low turbulence levels in the wind tunnel, we expect the flow to quickly develop turbulence given the complex geometry of the robot and the relatively high reference Reynolds number, therefore the turbulent quantities of the flow are expected to have a low dependence on boundary turbulence values \cite{lopes2017decay}.
This set of simulations comprises 104 runs, all conducted under identical boundary flow conditions defined at the geometry boundaries. However, these simulations varied in terms of geometry configurations and turbulence models employed. Specifically, we tested six different robot joint configurations: four symmetric (\texttt{hovering}, \texttt{flight30}, \texttt{flight50}, \texttt{flight60}) and two non-symmetric (\texttt{config05}, \texttt{config09}). For each configuration, we adjusted the attitude angles by modifying the support geometry, mirroring the conditions of the wind tunnel experiments. Subsequently, all tests were repeated to assess the performance of the aforementioned turbulence models. The average number of mesh elements for this simulation set was \qty{9.53e6}{}. Each test necessitated a unique mesh due to the variations in tested geometries. We built the meshes and computed the simulation results on an Intel(R) Xeon(R) Silver 4214 @ 2.20GHz CPU (48 threads).

For the free airstream simulations performed to acquire the aerodynamic forces dataset, we used only the robot model (without the wind tunnel support pilon) within a spherical fluid domain, assigning  \textit{velocity-inlet} conditions on the hemispherical surface facing the wind speed and \textit{pressure-outlet} on the other one.
This simulation set comprises approximately 8500 simulations, utilizing solely the \textit{SST $k-\omega$} turbulence model, which proved to be more accurate when compared to wind tunnel data. These simulations were conducted using a coarser mesh consisting of approximately \qty{0.33e6}{} elements (see Supplementary Table~3). Within this set, there were 30 different robot joint configurations, with three being symmetric (\texttt{hovering}, \texttt{flight30}, and \texttt{flight60}), and the remaining 27 being non-symmetric. Being the free air-stream fluid volume spherical, for each robot joint configuration, only one mesh was generated, and variations in robot attitude angles (comprising 19 angles of attack and 18 yaw angles, spanning all possible relative wind velocity directions) were simulated by adjusting the boundary conditions applied on the sphere surface. We built the meshes and computed the simulation results on an Nvidia RTX A4500 GPU (16GB VRAM).

\subsection{Aerodynamic Models for Learning and Online Control}\label{methods/aero_models}

The process we followed for the generation of the aerodynamic forces models could be summarized in three main parts: \textit{i)} generation of a model for CFD with simplified meshes to run fast simulations and build a large dataset of aerodynamic forces acting on each robot link, \textit{ii)} design of a Deep Neural Network algorithm to map the robot state (attitude and joint positions) into the aerodynamic forces of each robot link, leveraging on the aerodynamic dataset, \textit{iii)} design of an axisymmetric model of links' aerodynamic forces, for comparison with DNN model and robustness analysis.

\subsubsection*{Mesh sensitivity analysis for dataset generation}\label{methods/aero_models/mesh_reduction_dataset}

The significant computational cost required for precise CFD simulations poses a challenge in generating extensive datasets. Therefore, we opted to reduce the computational burden per simulation by reducing the number of mesh elements. Despite this approach could potentially compromise the accuracy in predicting flow boundary layer properties, we investigated the impact of mesh element count on overall aerodynamic force components. By comparing these results with experimental data, we validated the effectiveness of this methodology.

The mesh reduction study has been performed taking into account the relevant surface and volume mesh sizing parameters. We started from the highly refined \texttt{953C} mesh, made of \qty{9.53e6}{} elements, and gradually increased the sizing parameters to generate coarser meshes at each step up to the least refined \texttt{011C} mesh, made of \qty{0.11e6}{} elements, as reported also in Supplementary Table~3.

In Supplementary Fig.~5 there are the results of the mesh sensitivity analysis for simulations conducted on iRonCub in the \texttt{flight30} joint configuration, alongside with data obtained from wind tunnel experiments. These results demonstrate that the aerodynamic force from simulations with reduced mesh does not differ more than 15\% of the experimental data. Furthermore, the error remains constrained within the same bounds as the most accurate simulations.

In Supplementary Fig.~5 we compare velocity contours on the longitudinal plane between the most and less accurate CFD simulations to examine whether global aerodynamic similarity correlates with similar local behavior. The results show a considerable degradation in resolution in the wake of the robot for the less accurate simulation, while both solutions exhibit comparable values near the robot's surface. Hence, we opted for a less refined mesh to conduct the automated simulations. The final mesh contains $0.33\times 10^6$ elements (\texttt{033C} mesh, see Supplementary Table~3), representing the minimum number of elements feasible for generating meshes for various configurations, reducing the total number of cells of about 97\% with respect to the \texttt{953C} mesh.

The process for generating automatic CFD simulations can be divided into a modeling and a simulation phase.
The modeling phase requires three consecutive actions implemented with the software Ansys Workbench: \textit{i)} the joint configuration is modified in Ansys SpaceClaim according to randomly generated joint configurations, covering the robot joints state space during possible flight envelopes, subsequently, the spherical fluid volume enclosing the robot is generated; \textit{ii)} the \texttt{033C} poly-hexcore mesh is generated through Ansys Fluent Meshing software; \textit{iii)} The simulation is set up in Ansys Fluent using the same conditions validated with the wind tunnel experiments data.

The simulation phase is performed using Ansys pyFluent open-source API, which allows to perform simulations by accessing Fluent software via Python code, enabling automatic rotation of the mesh for each simulation to change the robot pitch and yaw angles (coherently updating the boundary conditions), heading to the dataset generation by spanning all the possible robot attitude angles for each identified robot joint configuration.

The final dataset is generated for 24 joint configurations (3 symmetric and 21 non-symmetric), 19 pitch angles (from \qty{0}{\degree} to \qty{180}{\degree} with a \qty{10}{\degree} interval), and for 18 yaw angles (from \qty{-160}{\degree} to \qty{180}{\degree} with a \qty{20}{\degree} interval), comprising a total of around 8500 simulations. The generated data have been augmented to enforce symmetry in the aerodynamic models, therefore the non-symmetric configurations results have been mirrored on the longitudinal symmetry plane, resulting in a final doubled homogeneous dataset.

\subsubsection*{Aerodynamic modelling for learning and online control}\label{methods/aero_modelling/sim_control_models}

We used the generated aerodynamic CFD dataset to fit aerodynamic models. More specifically, we designed a Deep Neural Network to map the multidimensional state of the robot (joint positions and wind direction) into the aerodynamic forces acting on the robot links. Then, we also build a less complex Axisymmetric Model that can be easily included into the flight controller for real-time prediction.

The Deep Neural Network (DNN) model has been trained and validated over the complete dataset, taking as input the robot (steady) state for each simulation performed: the relative wind velocity direction and the 19 robot internal DoFs. The outputs of the network are the three aerodynamic force coefficients for each link.
The network architecture is composed of 9 fully connected internal linear layers of 1048 neurons with Rectified Linear Unit (ReLU) activation function \cite{fukushima1969visual}, and by input and output layers of the same dimensions of the input $x\in \mathbb{R}^{22}$ and output $y\in \mathbb{R}^{39}$ vectors. The forward propagation algorithm for each of the internal linear layers can be expressed as
\begin{equation}
    a_{i} = \varphi(W^\top a_{i-1} + b) \, ,
\end{equation}
where $a_{i} \in \mathbb{R}^{n_i}$, $a_{i-1} \in \mathbb{R}^{n_{i-1}}$ are respectively the $i-th$ layer input and output parameters of dimensions $n_i$ and $n_{i-1}$ (based on the number of neurons of the layers), $W \in \mathbb{R}^{n_i \times n_{i-1}}$ is the layer weight matrix, $b \in \mathbb{R}^{n_i}$ is the layer bias vector, and $\varphi(\cdot)$ represents the ReLU activation function applied to each element of the function input vector as
\begin{equation}
    \varphi(u) = 
    \begin{cases}
        u, & u\ge 0 \\
        0, & u<0 \, .
    \end{cases}
\end{equation}
The network layers parameters (weight matrices and bias vectors) have been trained on a random subset (80\%) of the aerodynamic dataset generated through a random partitioning algorithm, the rest of the data (20\%) have been used for the validation of the trained neural network, to address possible overfitting.
The training has been conduced for 60000 epochs using Mean Squared Error (MSE) as loss function \cite{qi2020mean}, expressed as 
\begin{equation}
    \mathcal{L}_{MSE}(\theta) = \frac{1}{N}\sum_{i=1}^{N} || \hat{y}_i(x_i,\theta) - y_i(x_i) ||_2^2
\end{equation}
where $\theta$ represents the whole set of network internal parameters, $N$ is the number of samples used to compute the loss, $\hat{y}_i(x_i,\theta)$ are the forward propagation output computed for input $x_i$ and $\theta$ parameters, and $y_i(x_i)$ are the data from CFD simulations. 

We used the Adaptive Moment Estimation (Adam) optimization algorithm \cite{kingma2014adam} to train the network and we added a 10\% dropout regularization \cite{srivastava2014dropout} for each internal layer as a regularization for the training, and as an additional measure to prevent overfitting on the training dataset. We preferred the dropout method over $\ell_1$ and $\ell_2$ regularizations \cite{badola2020analysis} because we noted an improved training and validation losses convergence for this specific problem.

The Python code for the training of the proposed network architecture has been implemented using the deep learning library PyTorch \cite{paszke2019pytorch} using Nvidia RTX A4500 GPU as hardware, while the selected network hyperparameters have been chosen using the Optuna \cite{optuna_2019} optimization framework considering as evaluation metrics the training and validation MSEs.

Concerning the Axisymmetric Model design, we introduced a further assumption that considers each link isolated from the others and axisymmetric \cite{hui2022centroidal}. In this way, each link is supposed to be subject to aerodynamic drag and normal force components only, and any form of mutual aerodynamic interference is neglected. For such axisymmetric objects (in particular, spheres and cylinders) we can design axisymmetric models that describe the behavior of the drag force ($C_D$) and normal force ($C_N$) coefficients as sinusoidal functions of the link angle of attack \cite{wieselsberger1922further}, \cite{kritzinger2004drag}, \cite{hoerner1965fluid}. The approximated robot geometry according to such models appears as represented in Supplementary Fig.~6.

We decided to further simplify the model for the flying humanoid robot case by neglecting the $\reynolds$ number effect and the geometry effect (included in the computed force areas $C_D A$ and $C_N A$). These assumptions led us to write the following model for axisymmetric links:
\begin{equation}\label{eq:aero-force-model}
    F_{a_{i}} (\alpha_{l_{i}}, v_{a_i})  = - k_a |v_{a_i}| C_D A(\alpha_{l_{i}}) v_{a_i}
    + k_a C_N A(\alpha_{l_{i}})\ \frac{v_{a_i} \times \hat{k}_i \times v_{a_i}}{\sin\alpha_{l_{i}}} \, ,
\end{equation}
where $k_{a}=\frac{1}{2}\rho$ is the aerodynamic factor scaling the coefficients by the air density $\rho$, $\alpha_{l_{i}}$ is the i-th link angle of attack defined as the angle between the link relative wind velocity vector $v_{a_{i}}$ and symmetry axis $\hat{k}_i$. $C_D A$ and $C_N A$ are the aerodynamic drag and normal force areas as functions of $\alpha_{l_{i}}$.
The \textit{Axisymmetric Model} is derived from the model reported in \eqref{eq:aero-force-model}. The models of link drag force ($C_D A$) and normal force ($C_N A$) areas are obtained using a combination of \textit{Lasso} and \textit{Least Squares} linear regressions to identify the coefficients of a set of sinusoidal functions up to the $3^{rd}$ order, following similar literature models, and considering only the independent function terms (e.g. removing $\cos^2(\alpha_{l})$ which linearly depends on $\sin^2(\alpha_{l})$).
Being the drag coefficient continuous and symmetric over the span of the link angle of attack $\alpha_{link}$ for physical and geometric reasons, it should follow the constraints:
\begin{equation}\label{aeroModelLassoLSQconditions1}
    \frac{d C_D A}{d \alpha_{l}} \left(\qty{0}{\degree}\right) = 0 , \qquad 
    \frac{d C_D A}{d \alpha_{l}} \left(\qty{180}{\degree}\right) = 0.
\end{equation}
For modeling the normal force coefficient, we selected the cylindrical body model detailed in Supplementary Note~4, which includes the symmetry conditions:
\begin{equation}\label{aeroModelLassoLSQconditions2}
    C_N A \left(\qty{0}{\degree}\right) = 0 \, , \quad 
    C_N A \left(\qty{180}{\degree}\right) = 0 \, , \quad
    \frac{d C_N A}{d \alpha_{l}} \left(\qty{0}{\degree}\right) = 0 \, , \quad 
    \frac{d C_N A}{d \alpha_{l}} \left(\qty{180}{\degree}\right) = 0 \, ,
\end{equation}
leading to the following models for the aerodynamic force areas of each link:
\begin{subequations}
    \begin{equation}
        C_D A (\alpha_l) = w_0 + w_1 \cos(\alpha_l) + w_2 \sin^2(\alpha_l) + w_3 \sin^3(\alpha_l) + w_4 \cos^3(\alpha_l) \, , 
    \end{equation}
    \begin{equation}
        C_N A (\alpha_l) = w_5 \sin^2(\alpha_l) \cos(\alpha_l) \, .
    \end{equation}
\end{subequations}
The data for obtaining model coefficients for symmetric links are enhanced by mirroring the dataset to enforce symmetry. The Lasso regression algorithm helped to reduce the number of non-zero coefficients, decreasing model redundancy. The model coefficients are reported in Supplementary Table~2.

While the DNN model provides a more accurate estimation of the aerodynamic forces, the Axisymmetric Model is less complex to implement and to include in a real-time flight controller. Also, results are easier to interpret than with the DNN model. On the negative side, the approximations introduced by this model, in particular the hypothesis of no aerodynamic interference between links, negatively affects the model prediction MSE, which shows worse performances than the DNN.

\subsection{Aerodynamic Control Design and Simulation Environment}\label{methods/control_sim_design}

We designed a whole-body controller for flying humanoid robots accounting for aerodynamic effects. To test and measure the effectiveness of the proposed control strategy, we created a simulation environment based on \emph{Whole-Body Simulator} \cite{guedelha2022flexible}, a software suite designed to perform dynamic simulations of rigid body systems, based on MATLAB/Simulink and on the \texttt{iDynTree} physics engine \cite{nori2015icub}. We included both in the simulator and the flight controller the formulation of aerodynamic forces derived from the models presented in the previous section. Supplementary Note~5 reports details on the robot modeling and the complete derivation of flight control design.

\paragraph*{Notation}\label{methods/control_sim_design/control/notation}

\begin{itemize}
    \item $\mathcal{I} = \left\{O; \mathbf{X}, \mathbf{Y}, \mathbf{Z} \right\}$ is the inertial reference frame;
    \item $\mathcal{G}[\mathcal{I}]$ is a frame centered in the robot CoM, with the same orientation as $\mathcal{I}$;
    \item $\mathcal{B}[\mathcal{I}]$ is a frame centered in the robot \emph{base link}, with the same orientation as $\mathcal{I}$;
    \item ${}^{\mathcal{I}} v_{a} \in \mathbb{R}^3$ is the relative wind velocity of the robot CoM computed as ${}^{\mathcal{I}} v_{a} = {}^{\mathcal{I}} v_{\mathcal{G}} - {}^{\mathcal{I}} v_{w}$, where ${}^{\mathcal{I}} v_{w}$ indicates the absolute wind velocity in $\mathcal{I}$;
    \item $e_3 = ( 0, 0, 1, 0, 0, 0)^\top \in \mathbb{R}^6$ indicates the gravity force direction in $\mathcal{I}$;
    \item the operator $\hspace{0.1 cm}\dot{}\hspace{0.1 cm}$ indicates the time derivative of the vector or matrix on which it is applied.
\end{itemize}

\begin{assumption} \label{ass:aero-torque}
    The effects of the aerodynamic moments acting on each robot link are negligible compared to the effect of the aerodynamic forces.
\end{assumption}

\subsubsection*{Integration of Aerodynamic Forces for Simulation}\label{methods/control_sim_design/simulator}

Our models provide aerodynamic forces acting at the link's CoM. These forces are transformed into aerodynamic wrenches applied to the link local frame of reference, then they are multiplied by the transposed Jacobian matrix of the link, and summed as per
\begin{equation}
    {}^{\mathcal{B}[\mathcal{I}]}f_a = \sum_{i=1}^{n_a} J_{a_{i}}^\top F_{a_{i}} \, ,
\end{equation}
where ${}^{\mathcal{B}[\mathcal{I}]}f_a$ is the generalized total aerodynamic wrench acting on the robot, $F_{a_{i}}$ is the $i-th$ aerodynamic wrench at the $i-th$ link, and $J_{a_{i}}$ is the map between the system velocity $\nu$ and the link velocity ${}^{\mathcal{I}} \dot{o}_i$.
The generalized total aerodynamic wrench is integrated in the simulator to derive the robot dynamics evolution. Supplementary Fig.~7 shows the simulated aerodynamic forces, obtained by extracting a frame from the \texttt{iDynTree} internal visualizer.

\subsubsection*{Flight Control Design} \label{methods/control_sim_design/control}

The flight control is designed to stabilize the robot \emph{centroidal momentum}, whose time derivative equals the sum of all external wrenches applied to the robot, including aerodynamic forces. In what follows, we review the control strategy proposed in previous work \cite{pucci2017momentum, nava2018position} to which we added the contribution of aerodynamics. The centroidal momentum rate of change with aerodynamic forces writes:
\begin{equation}\label{eq:centroidal-momentum-aerodynamics}
    {}^{\mathcal{G}[\mathcal{I}]} \dot{h} = A_T(q) T + A_a(q) f_a - mge_3 \quad ,
\end{equation}
where $A_T \in \mathbb{R}^{6 \times m}$ a proper matrix that sums the effect of jet forces in the momentum equation, $T \in \mathbb{R}^{m}$ the jet intensities, $m$ the total robot mass and $g$ the gravity acceleration.

Being the centroidal momentum strictly related to jet forces, a possibility to stabilize its dynamics is to perform \emph{thrusts vectoring}, i.e., control thrusts intensity and direction to stabilize the momentum over a user-defined trajectory. However, the thrust direction is nonlinearly related to the robot joints through matrix $A_T(q)$, and this complicates the control design.

To address the problem, we decided to perform \emph{relative degree augmentation}. Furthermore, we assume that, for the purpose of this paper and the flight envelope we developed, the impact of aerodynamic forces variations is negligible. After rearranging some terms (see Supplementary Note~5), the second derivative of momentum rewrites:
\begin{equation}
    {}^{\mathcal{G}[\mathcal{I}]}\ddot{h} = (\Lambda_a(q, f_a) + \Lambda_T(q, T)) \nu + A_T \dot{T} \, ,
\end{equation}
which is linear w.r.t. joints velocity (contained in $\nu$) and thrust rate of change $\dot{T}$. Hence, we decided to use these quantities to control the momentum, and define $u := (\dot{T}, \dot{s})$.

In previous work we demonstrated that it is possible to write a smooth control input $u^*$ that renders the closed-loop equilibrium point $(\dot{\tilde{h}}, \tilde{h}, I) = (0,0,0)$ globally asymptotically stable, where $\tilde{h} = h-h^d$ is the momentum error, $h^d$ is the reference momentum trajectory and $I := \int_0^t \tilde{h}$. This is achieved by applying \emph{feedback linearization}: the input $u^*$ is first used to cancel out nonlinearities in the momentum dynamics, and then to impose a desired closed-loop dynamics with proven stability properties \cite{pucci2017momentum}. To implement the aerodynamics control proposed in this paper, we included the additional term $\Lambda_a(q, f_a) \nu $ in the control law design.

The control algorithm is implemented as a Quadratic Programming (QP) problem:
\begin{align}\label{eq:optimization-QP}
    u^{**} & = \underset{u}{\mathrm{argmin}} (w_1|u-u^*|^2 + w_2|\dot{s}-\dot{s}^*|^2) \\
    & s.t. \qquad u_{min} \leq u \leq u_{max}, \notag
\end{align}
where $\dot{s}^*$ defines a \textit{postural task} to resolve actuation redundancy and keep the robot position close to a desired shape, and the constraints $u_{min}$ and $u_{max}$ are not simply bounds of the input variables: they are designed to bound both the input $u$ and its time integral thanks to the properties of hyperbolic tangent \cite{rapetti2020}.

The solution of the QP optimization $u^* = (\dot{T}^*,\dot{s}^*)$ needs to be actuated by the jet engines and the robot joint torques. Concerning the jet engines, the optimal thrust derivative $\dot{T}^*$ is integrated and the obtained thrust force is plugged into the simulator directly. Concerning instead the optimal joints velocity $\dot{s}^*$, it is connected to joint torques via an inverse dynamics controller. From Euler-Poincar\'e formalism (see Robot Modeling in Supplementary Note~5), we considered the $n$ rows or elements associated with the joints dynamics as
\begin{equation}
    \bar{M}_s \ddot{s} + \bar{b} = \tau \quad .
\end{equation}
Note that the effect of the aerodynamic wrenches is also included in the joints dynamics. To stabilize the reference joints velocity provided by the flight controller, we apply \emph{high gain control} on the joints acceleration and define the following closed loop dynamics:
\begin{equation}
    \ddot{s}^{**} = -K_{Ds} (\dot{s}-\dot{s}^*) -K_{Ps}(s-s^*) \, ,
\end{equation}
with $K_{Ds}, K_{Ps}$ positive gains matrices, and $s^*$ is obtained by integrating $\dot{s}^*$.  Although $\dot{s}^*$ and $s^*$ depend on the robot state, we assume that with proper gain tuning the joints dynamics can be stabilized fast enough to avoid coupling effects with the flight controller. Finally, we choose the joint torques as:
\begin{equation}
    \tau = \bar{b} - \bar{M}_s^{-1} \ddot{s}^{**} \, .
\end{equation}
With the control framework described above we were able to control the robot in presence of simulated wind, while the baseline controller, without the addition of aerodynamic forces, failed to stabilize the robot in this use case, as presented in the Results section.


\newpage

\subsection*{Data Availability}\label{data}

The data that support the findings of this work are available from the corresponding authors upon reasonable request.

\subsection*{Code Availability}\label{code}

The code developed to derive the results of this work is available at https://github.com/ami-iit/paper\_paolino\_2024\_nature\_learning-aerodynamics-control.

\subsection*{Acknowledgments}
This research article was made possible by the effort of many people who contributed at different stages of the project. In particular, we would like to acknowledge the precious help of Paolo Maria Viceconte, Giuseppe L'Erario, Silvio Traversaro, and Francesca Bruzzone, in making this paper a reality.

\subsection*{Authors' contributions}
\begin{itemize}
    \item Conceptualization: A.P., G.N., F.B., P.R.V., D.P.
    \item Methodology: A.P., G.N., F.B., P.R.V., D.P.
    \item Software: A.P., G.N., F.D.N., F.B., P.R.V.
    \item Validation: A.P., G.N., F.D.N., F.B., P.R.V., D.G., L.R.
    \item Formal analysis: A.P., G.N., D.P.
    \item Investigation: A.P., G.N., F.D.N., F.B., P.R.V.
    \item Resources: G.N., A.Z., D.P.
    \item Data Curation: A.P., G.N., F.D.N.
    \item Writing: A.P., G.N., F.D.N., F.B., P.R.V., D.G., L.R., A.Z., R.T., G.I., D.P.
    \item Visualization: A.P., F.D.N., F.B., P.R.V.
    \item Supervision: G.N., A.Z., R.T., G.I., D.P.
    \item Project administration: G.N., D.P.
\end{itemize}

\subsection*{Competing interests}
The authors declare no competing interests.

\end{document}